\documentclass{article}
\usepackage[a4paper,top=3cm,bottom=2cm,left=3cm,right=3cm,marginparwidth=1.5cm]{geometry}

\usepackage[english]{babel}
\usepackage[utf8]{inputenc} 
\usepackage[T1]{fontenc}    
\usepackage{hyperref}       
\usepackage{url}            
\usepackage{booktabs}       
\usepackage{amsfonts}       
\usepackage{nicefrac}       
\usepackage{microtype}      

\usepackage{amsmath}
\usepackage{amsthm}
\usepackage{multicol}
\usepackage{multirow}
\usepackage{subcaption}
\usepackage{adjustbox}
\usepackage{enumitem}
\usepackage{lipsum}
\usepackage{algorithm}
\usepackage{xcolor}
\usepackage{wrapfig}
\usepackage[title]{appendix}
\usepackage[noend]{algpseudocode}
\usepackage{paralist}

\title{Evaluation metrics for behaviour modeling}

\author{%
  Daniel Jiwoong Im, Iljung Kwak, Kristin Branson \\
  Janelia Research Campus, HHMI, Virginia\\
  \texttt{jiwoong.im90@gmail.com},\\
  \texttt{\{kwaki, bransonk\}@janelia.hhmi.org} \\
}

\begin{document}

\maketitle
\begin{abstract}
A primary difficulty with unsupervised discovery of structure in large data sets is a lack of quantitative evaluation criteria. In this work, we propose and investigate several metrics for evaluating and comparing generative models of behavior learned using imitation learning. Compared to the commonly-used model log-likelihood, these criteria look at longer temporal relationships in behavior, are relevant if behavior has some properties that are inherently unpredictable, and highlight biases in the overall distribution of behaviors produced by the model. Pointwise metrics compare real to model-predicted trajectories given true past information. Distribution metrics compare statistics of the model simulating behavior in open loop, and are inspired by how experimental biologists evaluate the effects of manipulations on animal behavior. We show that the proposed metrics correspond with biologists' intuitions about behavior, and allow us to evaluate models, understand their biases, and enable us to propose new research directions. 
\end{abstract}

\section{Introduction}

A fundamental question in neuroscience is to understand the algorithms animals use to navigate the world, find resources and mates, and avoid predation. We want to understand the neural and biomechanic implementations of these algorithms, and the reasons these algorithms evolved~\cite{brain2.0}. One approach is to hand-craft an analytical model (usually of neural activity in a small circuit or portion of the brain), show correspondences between the model and experimental data (e.g.~neural recordings), then posit that interpretable properties of the model also hold true for the real organism~\cite{abbott2008theoretical}. Recently, complex models (e.g. high-dimensional linear or deep networks) have been fit to experimental data using machine learning methods~\cite{yamins2016using,kriegeskorte2018cognitive}. Again, assuming properties of the model are true of the organism, the machine learning model is then investigated and interpreted to gain insight. An advantage of this approach are that we can better fit the complexity of the organism, which will be critical in understanding larger circuits, the brain, and the organism as a whole. It also has the potential to allow us to go beyond our own, sometimes limited, imaginations, and discover new ideas from big neuroscience data sets. The obvious disadvantage is that these learned models are difficult to understand. In the best case, we have reduced the problem of understanding animals to that of understanding deep neural networks. 

To be successful in this approach, we must solve the following subproblems. First, we must develop a machine learning method that can learn a model of the animal that is so accurate it uses the same algorithm as the animal. Second, we must glean biological insight from this model. Third, we can develop experiments to specifically test these insights. In this paper, we focus on this first problem, how do we learn the model, and, in particular, how do we know if the model we've learned is sufficiently accurate to be of interest? Can we develop quantitative criteria of accuracy so that we can develop new models and compare them? 

We focus on the problem of modeling behavior of an organism as a whole, in particular locomotion and social behavior of {\em Drosophila melanogaster}, fruit flies. Our goals are to understand the low-dimensional structure of behavior, what properties of their environment animals encode, and how this influences their behavior. Modeling behavior in its naturalistic complexity is a recent trend in neuroscience research enabled by new technology for monitoring animal behavior and whole-brain neural activity. 

Our models are learned from pose trajectories of animals automatically extracted from video of animals behaving~\cite{Eyrun2014}. We train models that input sensory information about the animal's current environment and information about the animal's past behavior, and predict the movement of the animal at the next time point~\cite{Eyrun2016,johnson2020probabilistic}. Note that this is a form of unsupervised learning, as no annotations from humans are necessary~\cite{mathieu2015deep}. 

We propose several new criteria for evaluating the accuracy of a learned behavior model. Compared to previous work, these criteria look at longer temporal relationships in behavior, are relevant if behavior has some properties that are inherently unpredictable, and highlight biases in the overall distribution of behaviors produced by the model. We show the performance of state-of-the-art machine-learning-based models with these criteria, and argue that all of them are useful for understanding different properties of model accuracy.

\section{Related Work}
`\label{sec:related}

Learned models that predict future behavior from past have recently been proposed in neuroethology research~\cite{linderman2019hierarchical,heras2019deep,NIPS2018_8289,wiltschko2015mapping}. We contribute to this body of research by closely considering how to evaluate the accuracy of these models. Our work builds off of \cite{Eyrun2016}, in which deep networks are trained to predict future fly behavior given past. Their only quantitative evaluation was performed using the same criterion used for training, the negative log-likelihood of the prediction one frame into the future. In this paper, we propose a variety of new criteria to evaluate the models they proposed and other baselines. 

Our work is closely related to activity forecasting~\cite{kitani2012activity,becker2018evaluation,alahi2016social,sadeghian2019sophie,ridel2020scene,hasan2018mx}, in which the goal is to predict an object or person's future trajectory. The majority of this work focuses on predicting human or car trajectories, with the goal of utilizing these predictions e.g.~to avoid collisions. In contrast, the goal in our work is unsupervised discovery of behavior structure and biological insight. Thus, criteria for success may be different. Activity forecasting has been used in sports analytics~\cite{felsen2017will,yue2014learning} to understand strategies players are using. Our research into evaluation criteria is relevant to this work as well. 

In imitation learning, the goal is to learn to control agents or robots by imitating experts, for example, a humanoid agent learning to walk from motion capture data of humans walking~\cite{ross2010efficient, wang2017robust, merel2017learning,ho2016generative}. As in activity forecasting, in the majority of this research, the model can be treated as a black box, again differing from our goals. 

The large majority of these applications evaluate model accuracy using only the prediction in the next time step. \cite{deo2020trajectory} evaluates the quality of multiple predicted trajectories. We contribute to all of these approaches by proposing and evaluating several new evaluation criteria.

\section{Problem Setup}
\label{sec:representation}
\begin{figure}[t]
    \centering
    \begin{minipage}{0.49\textwidth}
        \includegraphics[width=\linewidth]{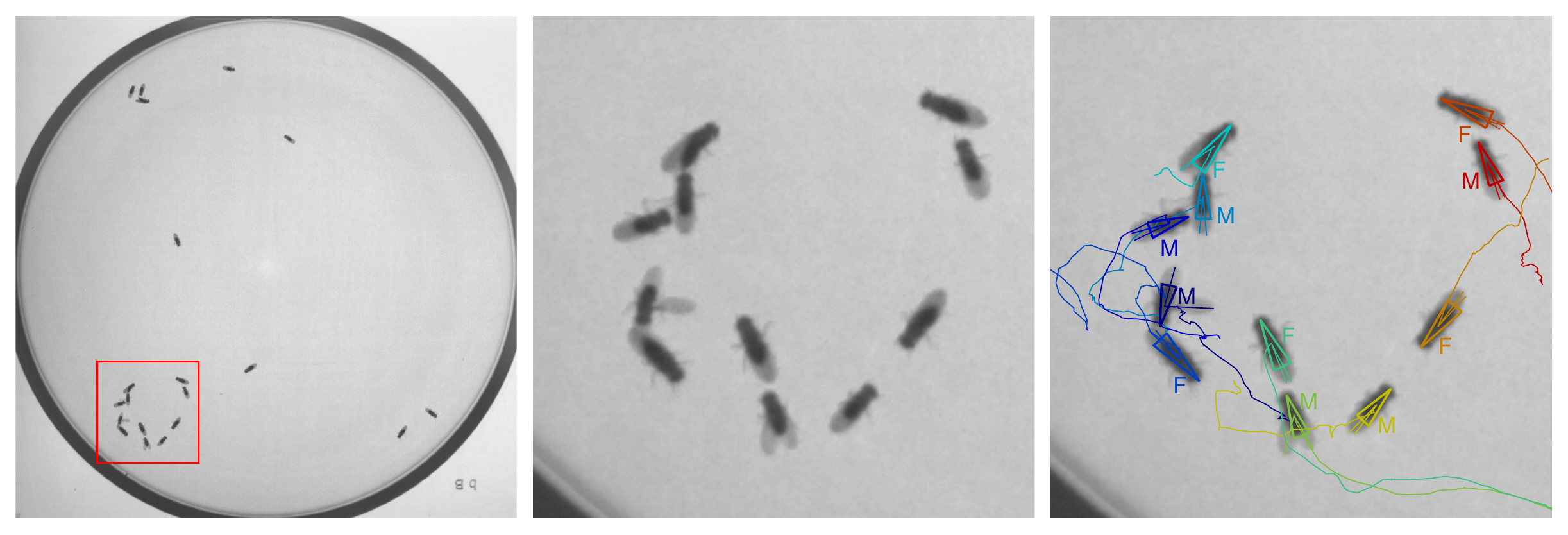}
        \vspace{-0.35cm}
    \end{minipage}
    \begin{minipage}{0.49\textwidth}
        \includegraphics[width=\linewidth]{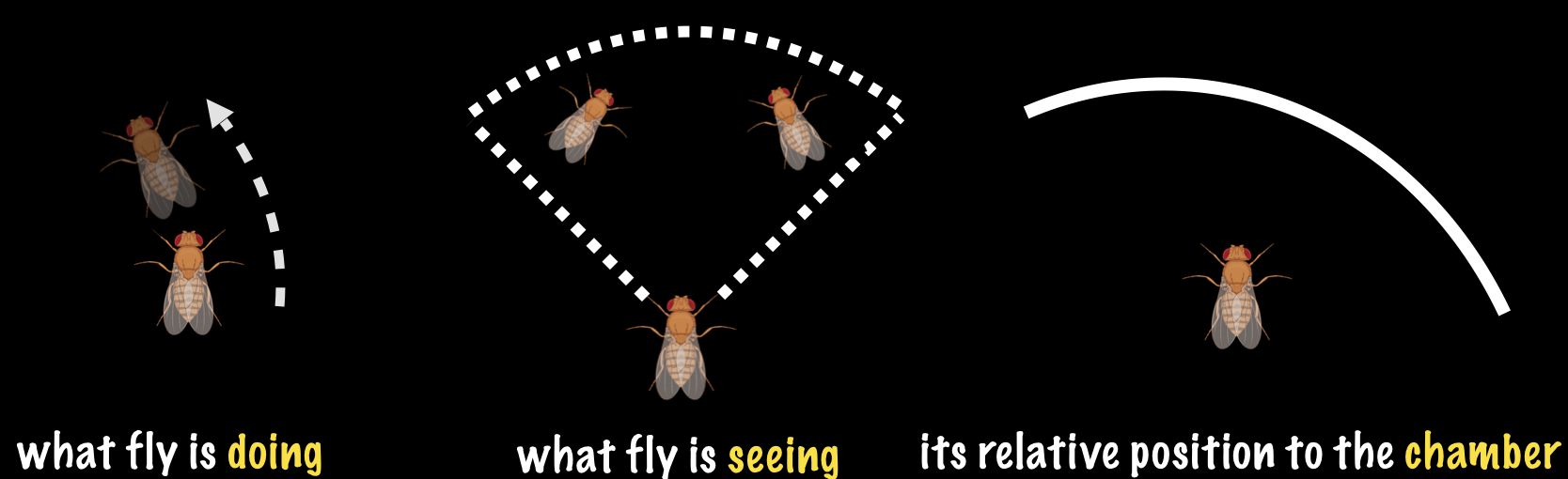}
        \vspace{-0.5cm}
    \end{minipage}
    \caption{Feature representation. (a) Groups of interacting animals are tracked. (b) Animals' movement and sensory input are computed from trajectories. }
    \label{fig:features}
    \vspace{-0.35cm}
\end{figure}

Before introducing the metrics for assessing behaviour models,
we describe the general framework used for learning models. 
Following~\cite{Eyrun2016}, our models are trained to predict movement between the current and next frame, given past movement
sequences and current sensory data. To train these models, we begin with videos of animals behaving, then use tracking software~\cite{Eyrun2014} to estimate their position and pose in every frame (Fig.~\ref{fig:features}(a)). We convert these trajectories into feature representations that are in the animal's coordinate system, and roughly approximate what it senses and degrees of freedom of motion. We estimate the visual and tactile sensory information available to each fly (Fig.~\ref{fig:features}(b)). These are nonlinear functions of the distance to the nearest other animal and the distance to the arena wall, respectively, in each of 72 circular sectors, resulting in a 144-dimensional representation of sensory information. At each time step $t$, our models are trained to predict the 8-d inter-frame movement of the animal: its forward $f_t$ and sideways $s_t$ velocities, change in body orientation $\theta_t$, major axis length $m_t$, and left and right wing angles and length $\phi^{l,r}_t$ and $l^{l,r}_t$. To obtain a multimodal, probabilistic model, we discretize each motion feature independently into 51 bins, with the output in each bin being the probability of selecting that movement. This was shown to be important for performance in \cite{Eyrun2016}. 

\begin{figure}[t]
    \centering
    \begin{minipage}{0.16\textwidth}
        \includegraphics[width=0.99\linewidth]{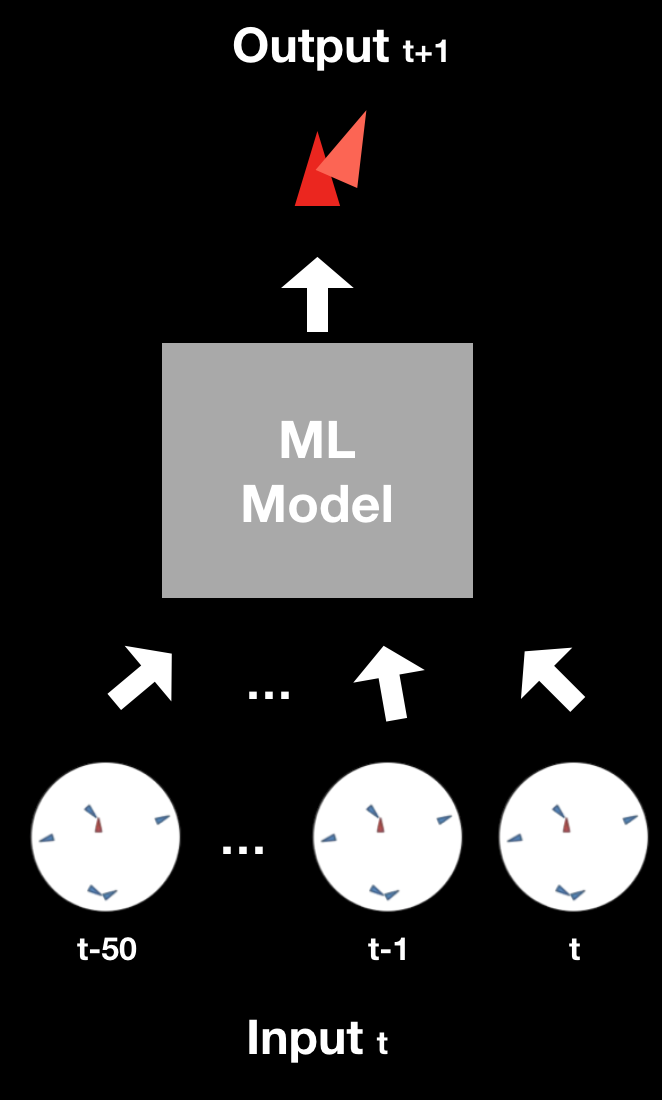}
        \vspace{-0.5cm}
    \end{minipage}
    \begin{minipage}{0.16\textwidth}
        \includegraphics[width=0.99\linewidth]{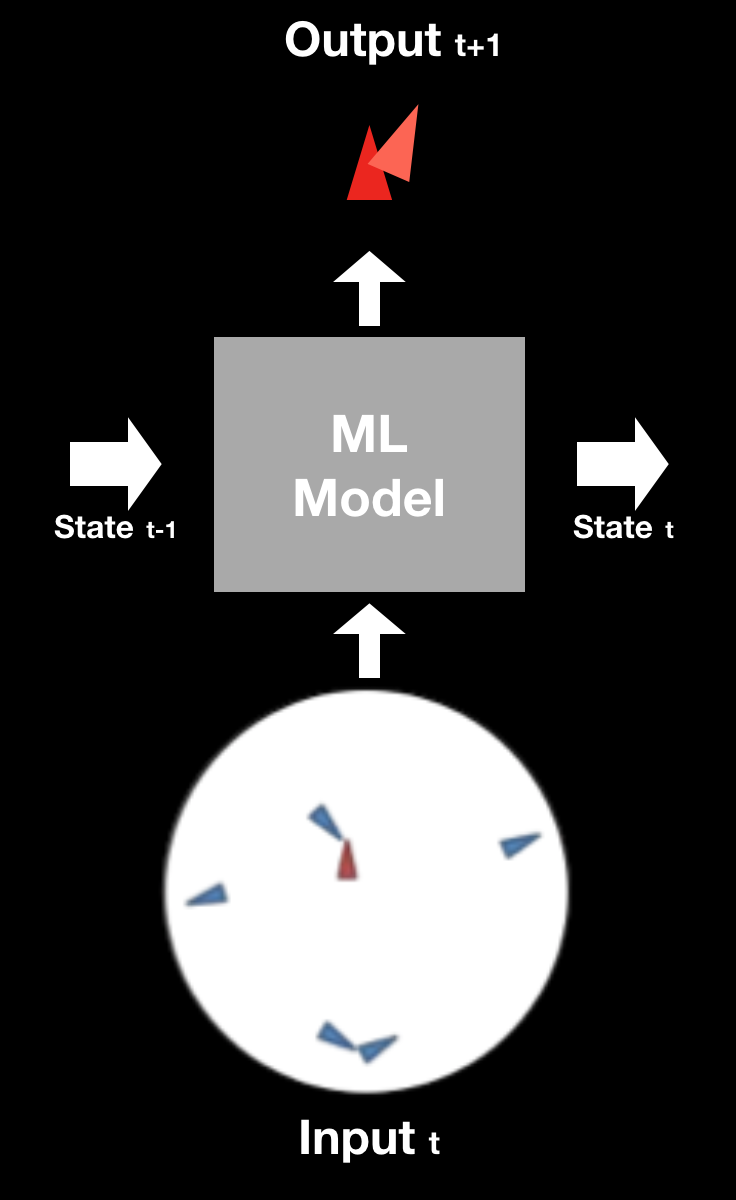}
        \vspace{-0.5cm}
    \end{minipage}
    \begin{minipage}{0.37\textwidth}
        \includegraphics[width=0.99\linewidth]{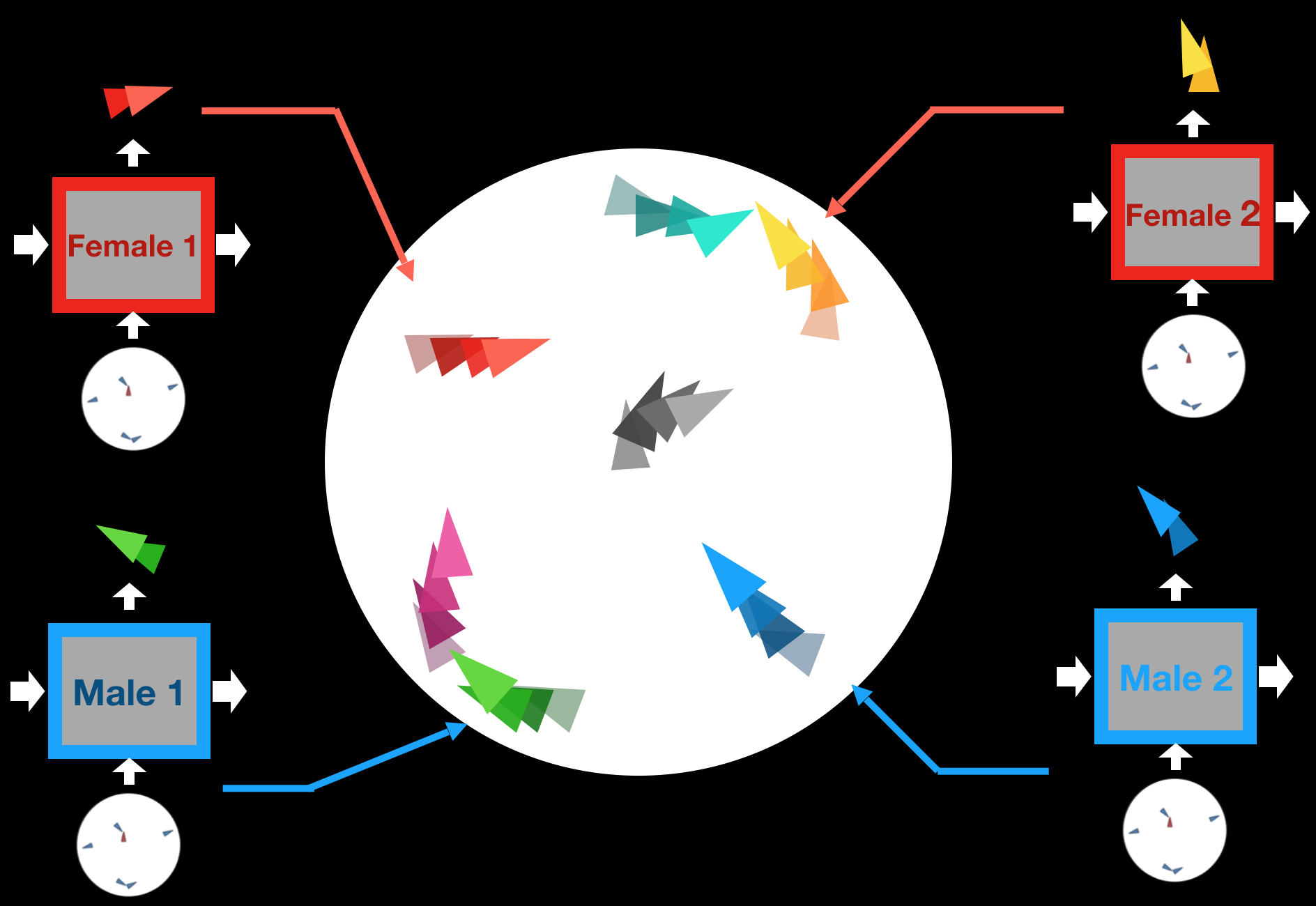}
        \vspace{-0.5cm}
    \end{minipage}
    \begin{minipage}{.27\textwidth}
        \includegraphics[width=0.99\linewidth]{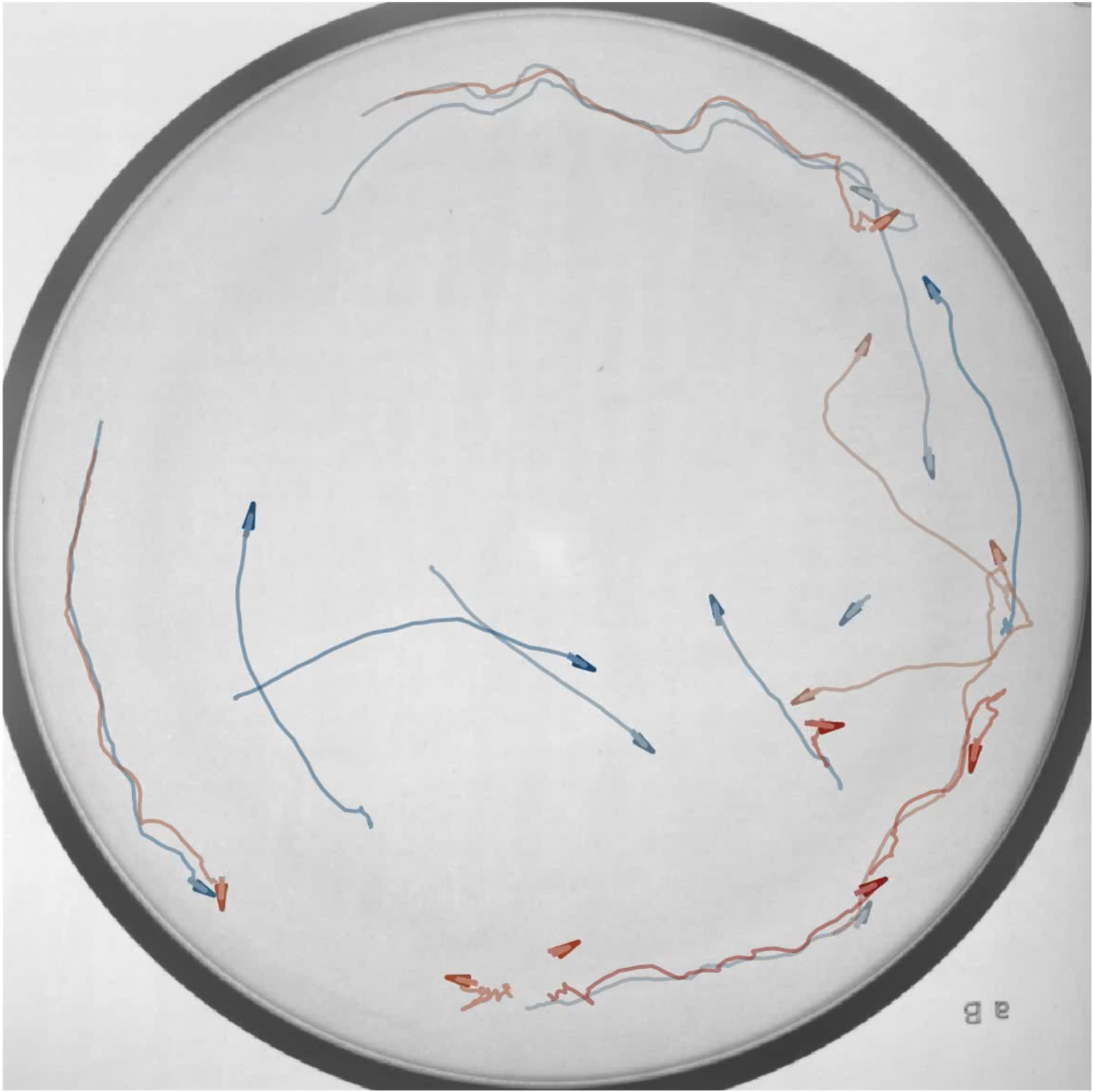}
        \vspace{-0.5cm}
    \end{minipage}
    \caption{(a) Autoregressive and (b) sequential model types. (c) Schematic of multi-agent, multi-frame simulations. (d) Example 20-fly simulation. Male and female flies are indicated by blue and red triangles, resp. The trail indicates centroid positions in the previous 100 frames.}
    \label{fig:fly_model}
    \vspace{-0.5cm}
\end{figure}

We train four types of models: linear regression (LINEAR), convolutional neural network (CNN) \cite{Krizhevsky2009}, recurrent neural network with gated 
recurrent units (RNN) \cite{Cho2014}, and Hierarchical RNN (HRNN) \cite{Eyrun2016}. All these models take as input the 144-d sensory information. Linear regression and CNN are autoregressive models, and in addition input the past movement sequence from $t-\tau$ to $t$ (Figure~\ref{fig:fly_model}(a)). RNN and 
HRNN are sequential models, where instead input the previous internal state ${\bf h}_{t-1}$ (see Figure~\ref{fig:fly_model}(b)). Deep networks are trained using sum of cross entropy loss over each motion feature. Linear regression is trained to minimize the mean-squared error for each motion output independently. As the behavior of male and female flies in these contexts differed~\cite{Robie2017}, we trained separate models for each sex. 



\label{sec:simulations}

While the model was trained to predict single-frame movement, we can predict multiple times in a row to simulate behavior of many frames. We can create simulations of multi-agent interactions by simulating separate instances of the male and female models as shown in Figure~\ref{fig:fly_model}(c). Figure~\ref{fig:fly_model}(d) shows trajectories that result from such a multi-agent simulation. 


There are several simulation configurations that we consider: 
\begin{compactitem}
\item $\mathcal{S}_{\text{SMSF}}$ : All agents are learned models.
\item $\mathcal{S}_{\text{RMSF}}$ : All male agents are real and all female agents are learned models.
\item $\mathcal{S}_{\text{SMRF}}$ : All male agents are learned models and all female agents are real flies.
\item $\mathcal{S}_{\text{LOO},i}$: All agents are real flies except the $i^{th}$ fly.
\end{compactitem}


\section{Evaluation Metrics}
\label{sec:evaluation metrics}

In previous work \cite{Eyrun2016}, models are evaluated based on error in predicting behavior one frame into the future (one-step error). This is sensible, as this is closely related to the criterion used to train the models. However, this criterion alone does not capture everything important in modeling behavior. Because of the temporal smoothness of behavior, at any given time point it is very likely that the animal will continue to do what it was doing previously, thus the constant velocity model is often quite accurate. 

\subsection{$n$-step prediction error}
We propose examining the error of predicting $n$ steps into the future ($n$-step error) by using the model to simulate the behavior of the current fly $i$ for $n$ steps in closed-loop with the real behavior of other flies ($\mathcal{S}_{\text{LOO},i}$). Examining $n$-step error allows us to discriminate between noise in predictions that ultimately is averaged out and irrelevant over many frames and small per-frame errors that accumulate and result in large errors over time. 

The $n$-step error can be computed for any subset of the trajectory properties (e.g. centroid position and body orientation), and we can either examine each separately or weight and sum them. In our analyses, we focus on analyzing centroid position.

There are several reasons that animal behavior will not be perfectly predictable. Behavior depends on properties of individuals, internal state, or the environment that cannot be estimated from our input data; for example, we are not measuring chemical or odor the flies may respond to. In addition, it may be that behavior is truly stochastic \cite{Roberts2016}. If future trajectories are unpredictable and not unimodal in distribution, then Euclidean distance will not capture errors that matter. 

Instead of looking at the $n$-step error for one prediction, we propose using our models to create $m$ sample predictions, and compute the minimum error to any one sample. To do this, our models require probabilistic interpretations:
\begin{align}
    err_{xt}^{nm} := \min_{m'=1,...,m} \| \hat{x}_{t+n}^{(m')} - x_{t+n} \|_2
\end{align}
where $x_{t+n}$ is the true position at time $t+n$ and $\hat{x}_{t+n}^{(m')}$ is the $m'$th sample predicted using a our stochastic model $n$ frames into the future.

\subsection{Real vs Fake Discriminators}

Parallels can be drawn between our goal of modeling behavior and conditional Generative Adversarial Networks (GANs). Previously, \cite{Im2018} measured how realistic are samples of generative models are by using the divergence between real and fake distributions. We train a discriminative network $D$ to distinguish between real and fake trajectories. These networks input the past trajectory of the animal, its current environmental state, and the predicted trajectory. We evaluate the trained discriminator on a separate test set to evaluate our model. If the discriminator has poor accuracy, then the model is producing simulated samples indistinguishable from real, and the model is performing well. Following \cite{Im2018}, we train the discriminator to minimize the least-squares divergence between the distributions of real and simulated trajectories:
\begin{align}
        \mathcal{L}_{\text{LS}} &= \min_\varphi \mathbb{E}_{\bf{r}\sim p_{\text{data}(\bf{r})}} [(D_\varphi(\bf{r})-1)^2] + \mathbb{E}_{\bf{r}\sim p_{\text{model}(\bf{r})}}[(D_\varphi(\bf{r}))^2]
\end{align}
Note that we found similar results using the Jenson-Shannon divergence. 

Compared to the $n$-step error, the Real vs Fake (RvF) criterion allows us to use all frames and all pose features in the predicted trajectory. The discriminator learns how to combine these features to determine if a simulated animal moves in a realistic manner. If our discriminator is perfect, then we can think of this as measuring something related to the likelihood of our predicted trajectory given the past and environment. This is the opposite of our $n$-step error, which is more related to the likelihood of the true trajectory given our model distribution. 

We also asked biologist experts to distinguish real from fake trajectories in videos displayed to them.

\subsection{Distribution Errors}

The previous evaluation criteria conditioned on the model input -- the animal's current environment and past trajectories, and depends on the relationship between the model inputs and the true future trajectory. We refer to these as {\em point-wise} metrics because they compare past to future trajectories on a point-by-point basis. In this section, we consider {\em distribution} metrics, which compare the pooled distribution of statistics of the simulated trajectories to that of real trajectories. For these metrics, we simulate {\em all} animals in the video ($\mathcal{S}_{\text{SMSF}}$). These metrics are inspired by how biologists quantitatively compare animal behavior to, for example, determine the effects of an experimental manipulation. Instead of comparing one type of real fly to another, we compare real flies to simulated flies. 

{\bf Behavior Feature Histogram Distance}

In biology experiments, to understand how two samples differ, it is common to reduce behavior to a scalar, interpretable statistic, then compare empirical histograms of these statistics. Example statistics include the instantaneous speed of the fly, change in body orientation, the distance to the closest other animal, and distance to the arena wall. In particular, speed and inter-animal distance were related to the two principal components of variation in behavior of flies in a large neural activation screen~\cite{Robie2017}. 

We propose both visualizing and comparing the histograms to understand biases in the simulation distributions, as well as computing divergences (e.g. L1 distance) between the histograms to quantitatively compare algorithms.



{\bf Behavior Classifier Error}

Instead of choosing continuous features to describe behavior, we can discretize behavior into semantically meaningful classes~\cite{Robie2017} based on combinations of these statistics. For example, chasing is a component of fly courtship. We can compare real to fake trajectories by comparing the fraction of time that they spend performing these behaviors. This metric is also inspired by how biologists have performed interpretable comparisons of animal behavior based on their domain knowledge. 

We applied the chase classifier trained on many genotypes of real flies~\cite{Robie2017} to real and simulated trajectories, and compared the fraction of time the flies spent chasing. We can compare the raw fractions directly for intuition about how the behaviors differ, or quantify error:
\begin{equation}
    err_{\text{chase}} := \log \frac{p}{\hat{p}} p + \log \left( \frac{1-p}{1-\hat{p}} \right) (1-p)
\end{equation}
where $p$, $\hat{p}$ are the fraction of time real, simulated flies chase. This is proportional to the KL-divergence between binomial distributions with parameters $p$ and $\hat{p}$.



\section{Experiments}

To understand the proposed criteria, we considered a data set of 32 videos of $\approx 120$ male and female fruit flies from 3 genotypes (Section~\ref{sec:datasets}). We evaluated four learned models (Section~\ref{sec:models}). We first present the results of applying the proposed metrics to evaluate and the performance of several learned models (Section~\ref{sec:evaluation}). As we propose that it is necessary to consider many metrics when understanding how well models are performing, we combine these results to attempt to answer fundamental questions about our models (Section~\ref{sec:interpretation}). The model and training details can be found in the supplementary materials (S. M).

\subsection{Datasets}
\label{sec:datasets}

We evaluate models and proposed metrics in modeling the behavior of {\it Drosophila melanogaster} (fruit flies). We train models of the behavior of six different types of flies: male and female flies from three genotypes. These genotypes are a genetic control (CONTROL) and two GAL4 lines expressing dTrpA in different sparse subsets of neurons. Activation of R71G01 neurons induces male flies to court~\cite{zhou2015central,Robie2017}. Activation of R91B01 neurons has a very different phenotype -- flies avoid one another~\cite{Robie2017}. These GAL4 lines were chosen specifically because they result in very different, diametrically opposed behaviors. Each of these different but related data sets gives us a different dataset to compare our models and proposed metrics. In addition, comparing statistics of one type of fly to another provides a baseline that we should expect accurate models to beat.

For each genotype, we analyzed 9-12 15-minute, 30-fps videos, each containing 10 male and 10 female flies, collected by Robie et al.~\cite{Robie2017}. Thus, in total these 3 datasets comprised about 20 million data points. One of the videos for R71G01 was previously analyzed in \cite{Eyrun2016}. We partitioned the datasets into training, validation and test sets (Control: 6 train, 2 validation, 4 test; R71G01: 5 train, 2 validation, 4 test; R91B01: 5 train, 2 validation, 2 test). Following~\cite{Eyrun2016}, we tracked the body position, orientation, size, and wing angles of flies~\cite{Eyrun2014}. Models are trained using the resulting trajectories. 



\subsection{Training Real-vs-Fake Discriminators}

We trained RvF discriminators to predict whether 60-frame (2-second) trajectories were real (1) or fake (0), following how GAN discriminators are trained. Our discriminators were 1-d over time convolutional networks, with 3 convolutional layers of 32, 64, and 128 $1 \times 5$ filters, followed by two 64-d fully-connected layers. All layers used ReLU activation functions. We experimented with training the discriminator using cross-entropy and least-square loss \cite{Im2018} for 100 epochs, with $0.01$ learning rate, $0.0001$ L2-weight-decay coefficient, and 100 batch size. Validation accuracy during training is shown in S.M.~Fig.~\ref{fig:train_disc_supp}.



\subsection{Expert Biologist Real-vs-Fake Study}

We asked expert biologists to annotate whether trajectories were real or fake. In the first study, the biologist was shown one test fly and 19 real flies for 60 frames, equivalent to the task of the RvF discriminator. This was done for 80 real trajectory videos and 20 simulated trajectory videos from each of the four learned models. In the second study, all flies were either simulated or real, and videos were of variable length: 60, 120, 240, or 480 frames. 

\subsection{Additional Baselines}

We consider two fixed, simple baselines: that the fly stays in the exact same position in the next time point (HALT) and that the fly moves with a constant velocity (CONST). Our intuition is that our learned models perform better than these fixed baselines, thus it is important that metrics reflect this as well. 

When sensible, we compare trajectories from one genotype to another genotype. As the behavior of these genotypes is strikingly different, this provides a lower bound on how well we expect our models to perform. These results are indicated by the genotype name. 

As an upper baseline, we compare flies from one video of a genotype to flies of another video of the same genotype. These flies are genetically identical and raised in conditions controlled to be as identical as possible, thus we expect their behavior to be quite similar. These results are indicated by TRAIN DATA, as we use the model training data in this case. 


\subsection{Evaluation of Models and Metrics}
\label{sec:evaluation}

\subsubsection{ $n$-Step Error.} 
\begin{figure}[htb]
    \centering
    \begin{minipage}{0.24\textwidth}
        \includegraphics[width=\linewidth]{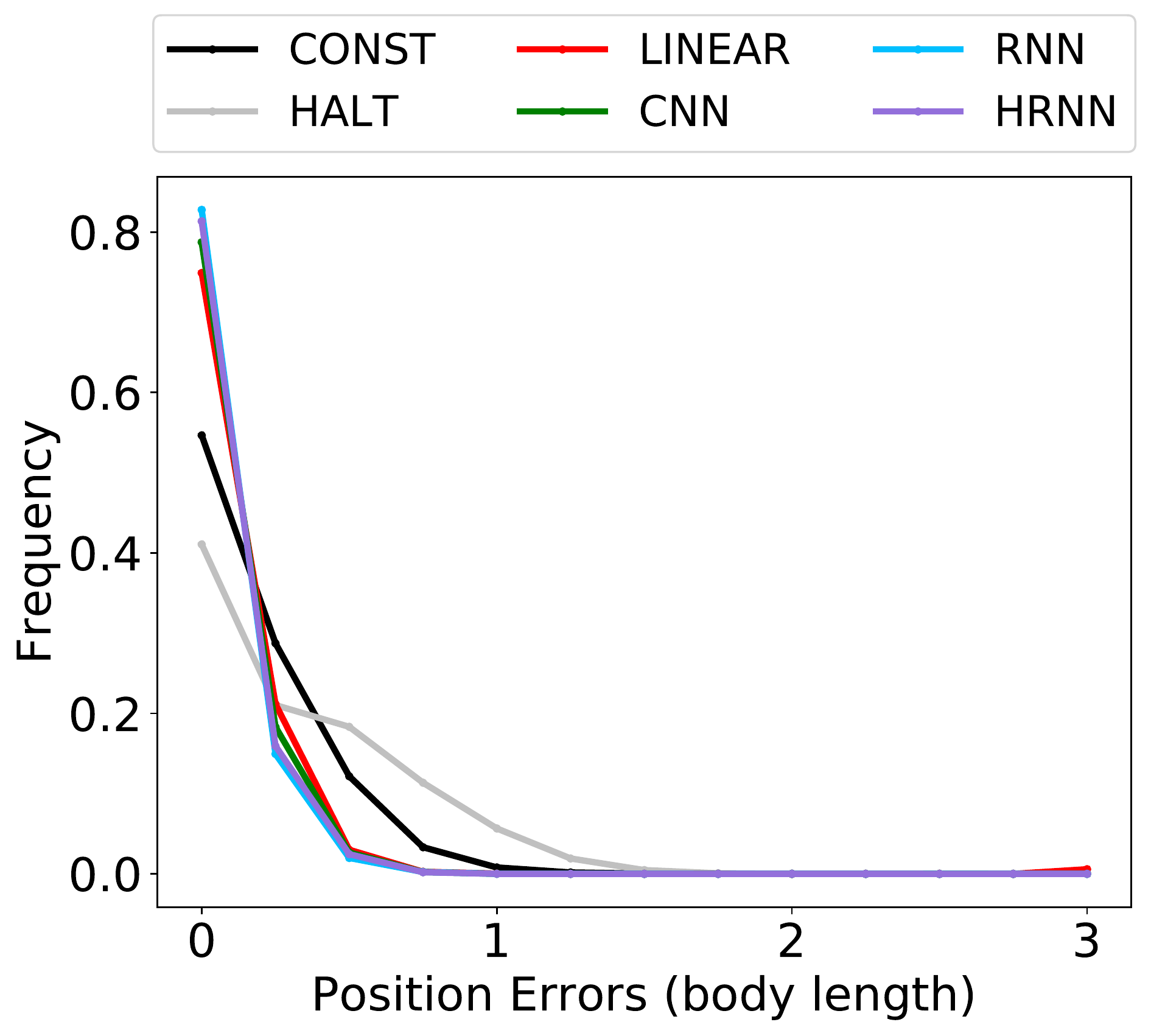}
        \vspace{-0.5cm}
        \subcaption{\begin{small}1-step, 1-sample\end{small}}
    \end{minipage}
    \begin{minipage}{0.24\textwidth}
        \includegraphics[width=\linewidth]{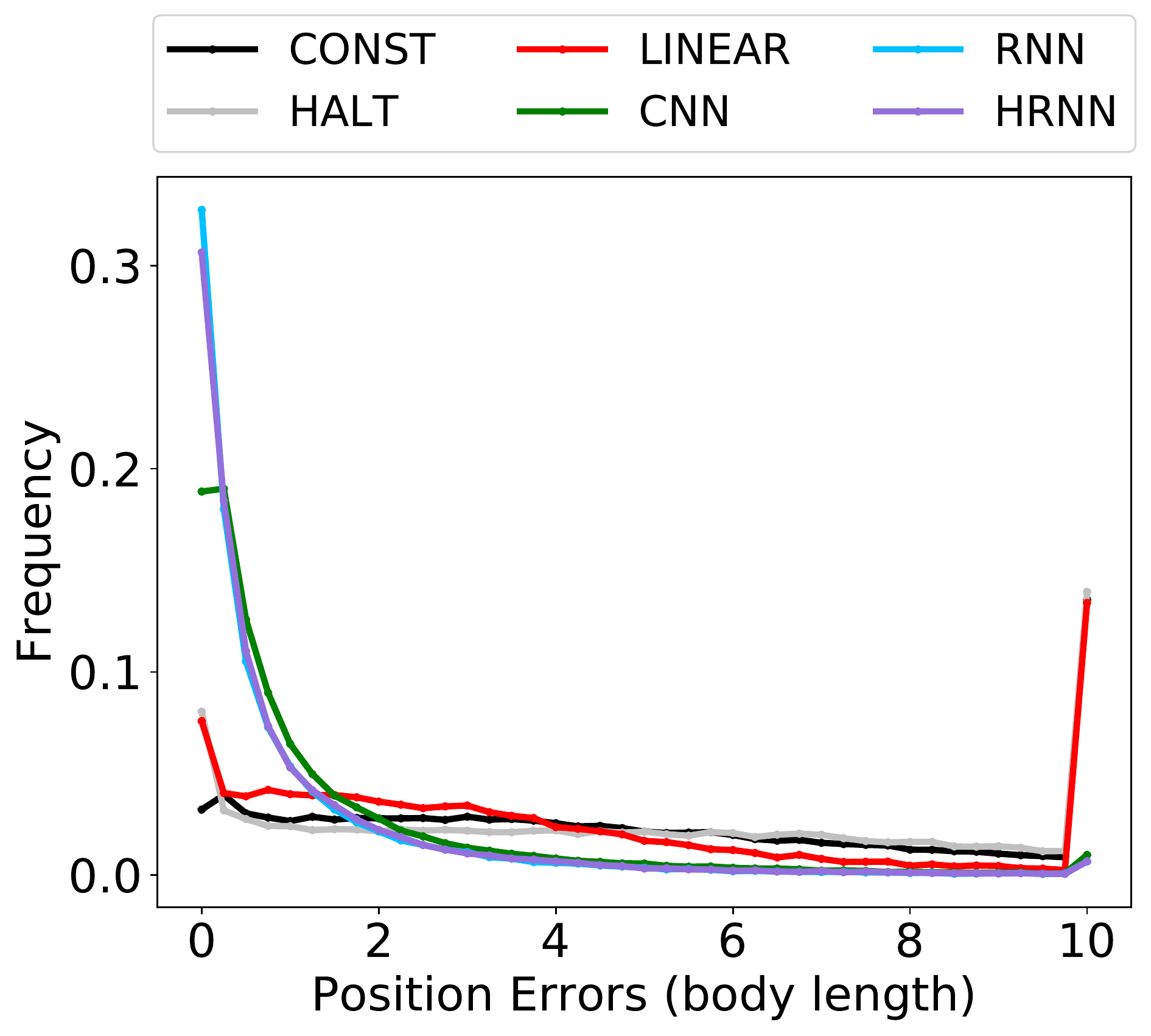}
        \vspace{-0.5cm}
        \subcaption{\begin{small}30-step, 10-sample\end{small}}
    \end{minipage}
    \begin{minipage}{0.24\textwidth}
        \includegraphics[width=\linewidth]{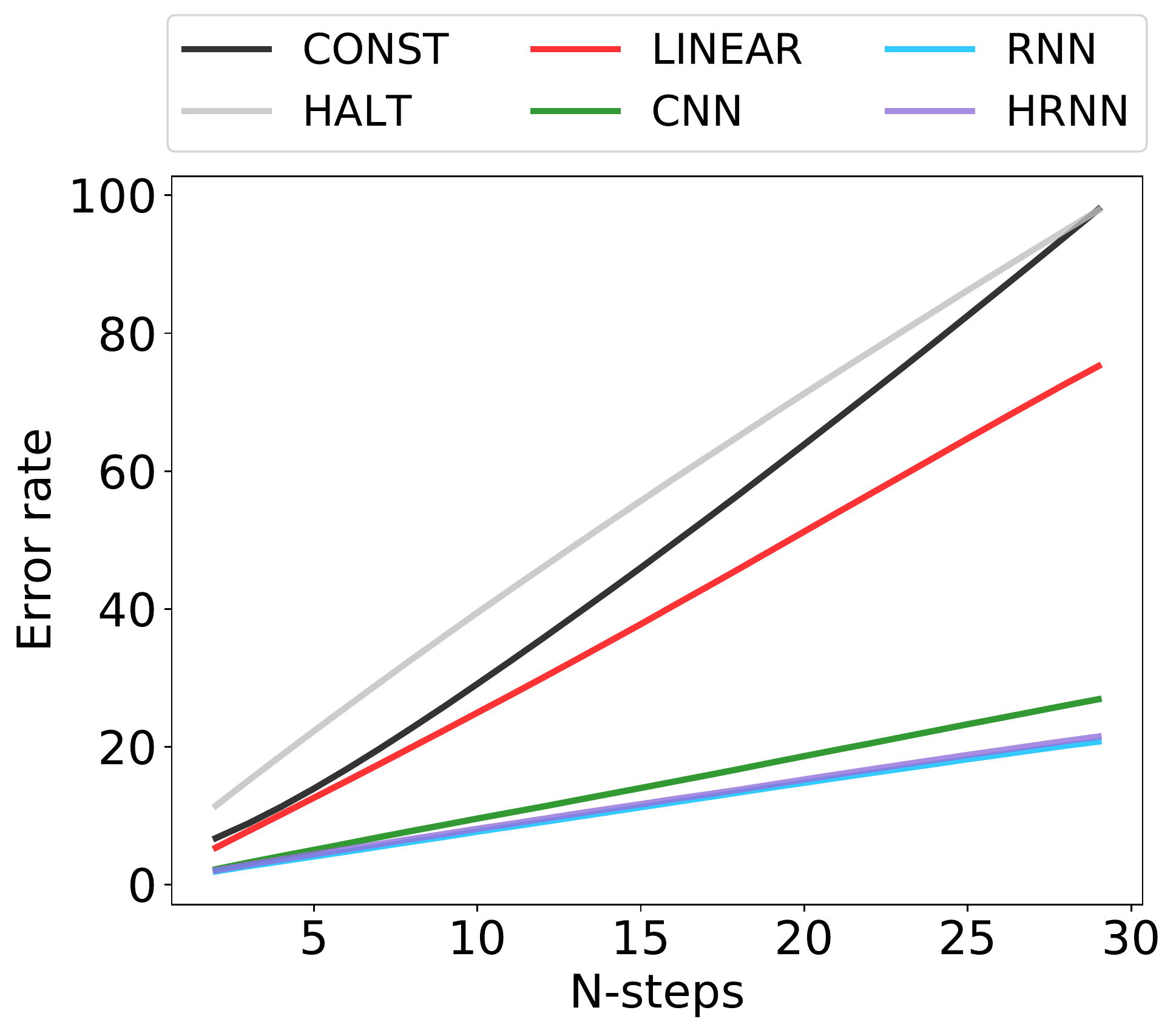}
        \vspace{-0.5cm}
        \subcaption{\begin{small}$n$-step, $10$-sample\end{small}}
    \end{minipage}
    \begin{minipage}{0.24\textwidth}
        \vspace{.5cm}
        \includegraphics[width=\linewidth]{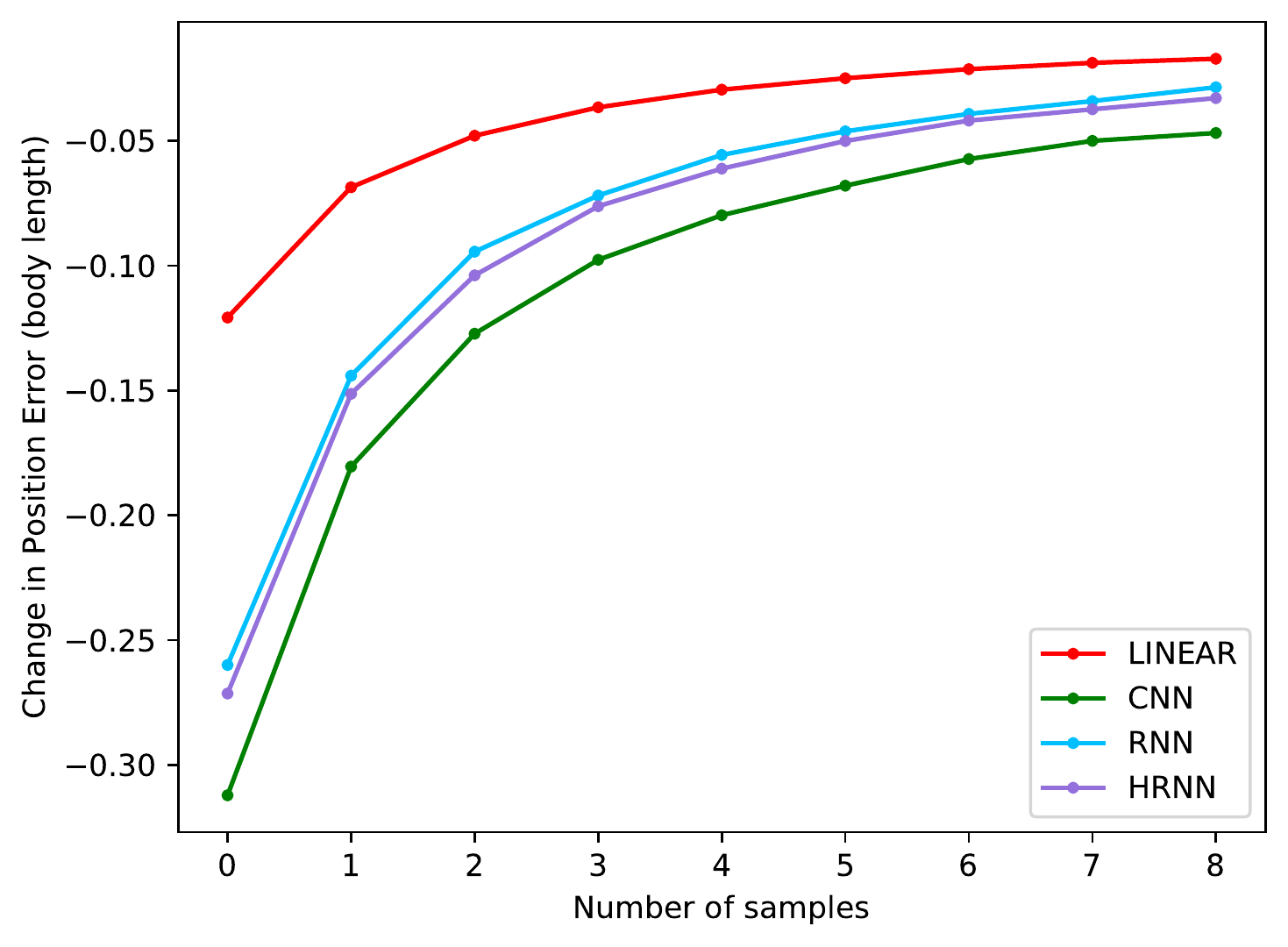}
        \vspace{-0.5cm}
        \subcaption{\begin{small}30-step, $m$-sample $\Delta$\end{small}}
    \end{minipage}
\caption{$n$-step prediction error for male flies, R71G01 (see S.M.~Fig.~\ref{fig:nstep_supp} for other datasets) \label{fig:nstep}}
\end{figure}

\begin{figure}[t] 
    \centering
    \begin{minipage}{0.3\textwidth}
        \centering
        \includegraphics[width=\linewidth]{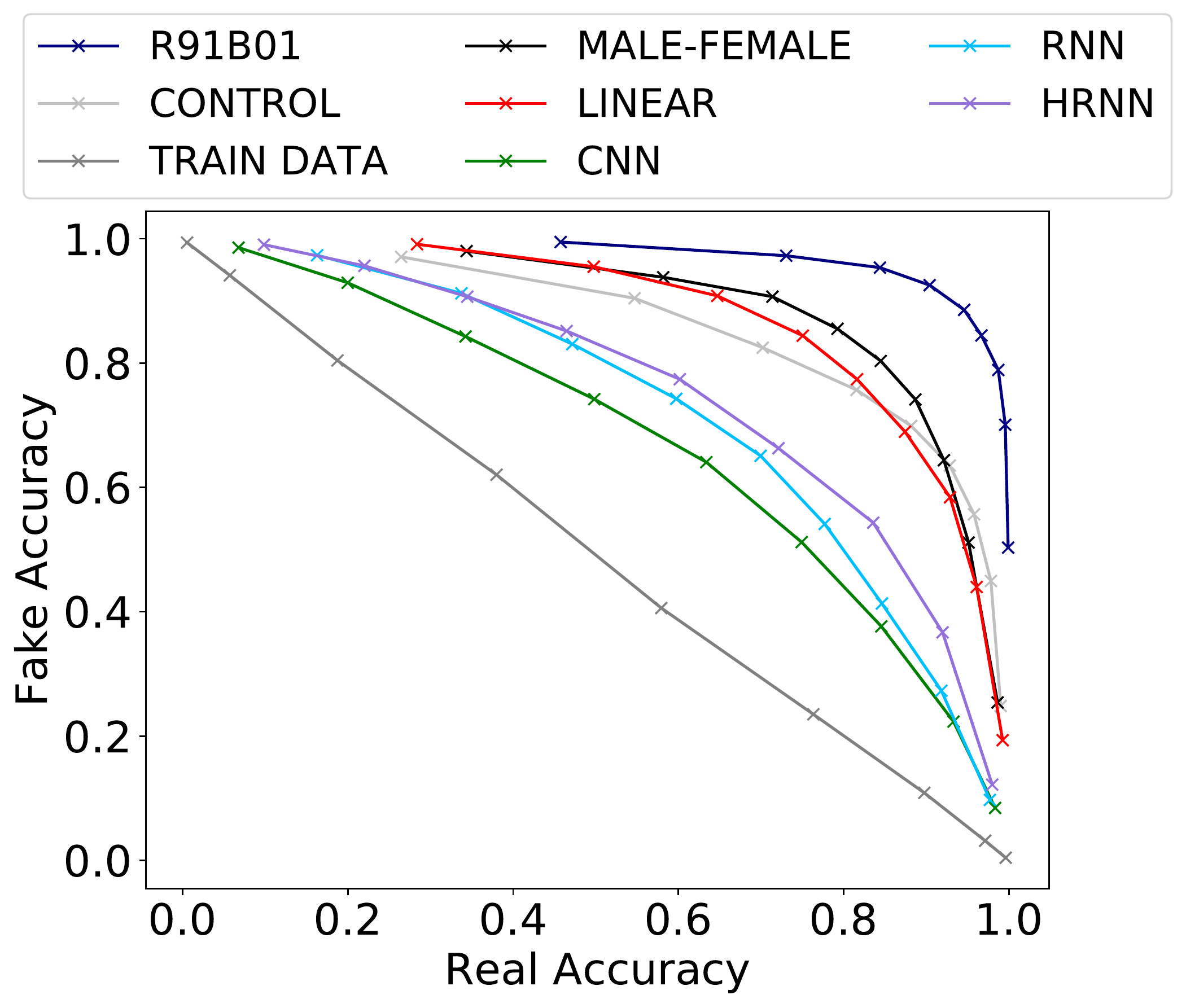}
        \vspace{-0.5cm}
        \subcaption{RvF discriminator accuracy}
        \label{fig:disc_eval}
    \end{minipage}
    \hspace{.02\textwidth}
    \begin{minipage}{0.3\textwidth}
        \centering
        \includegraphics[width=\linewidth]{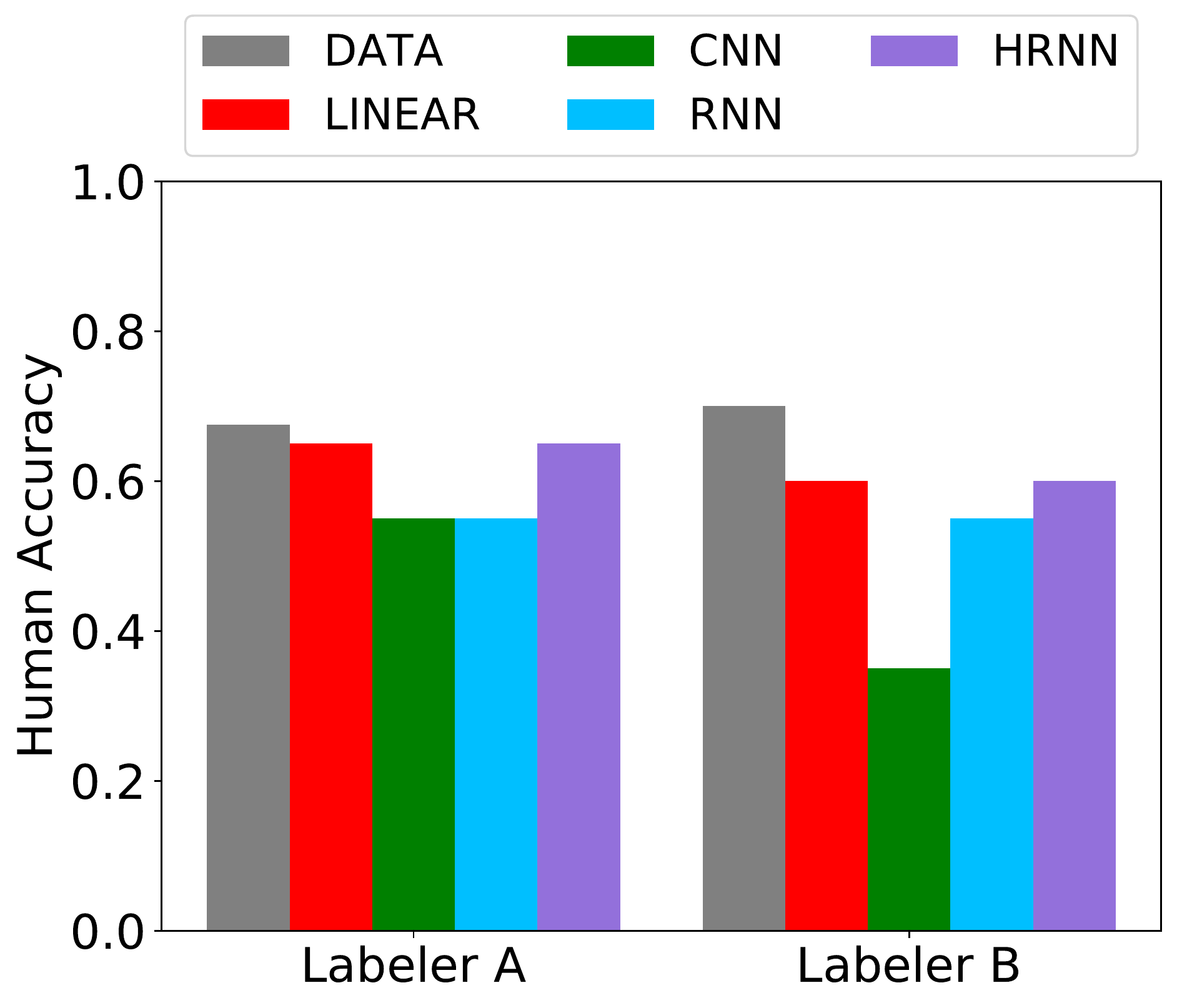}
        \vspace{-0.5cm}
        \subcaption{Biologist accuracy, one test fly}
        \label{fig:human_short}
    \end{minipage}
    \hspace{.02\textwidth}
    \begin{minipage}{0.3\textwidth}
        \centering
        \includegraphics[width=\linewidth]{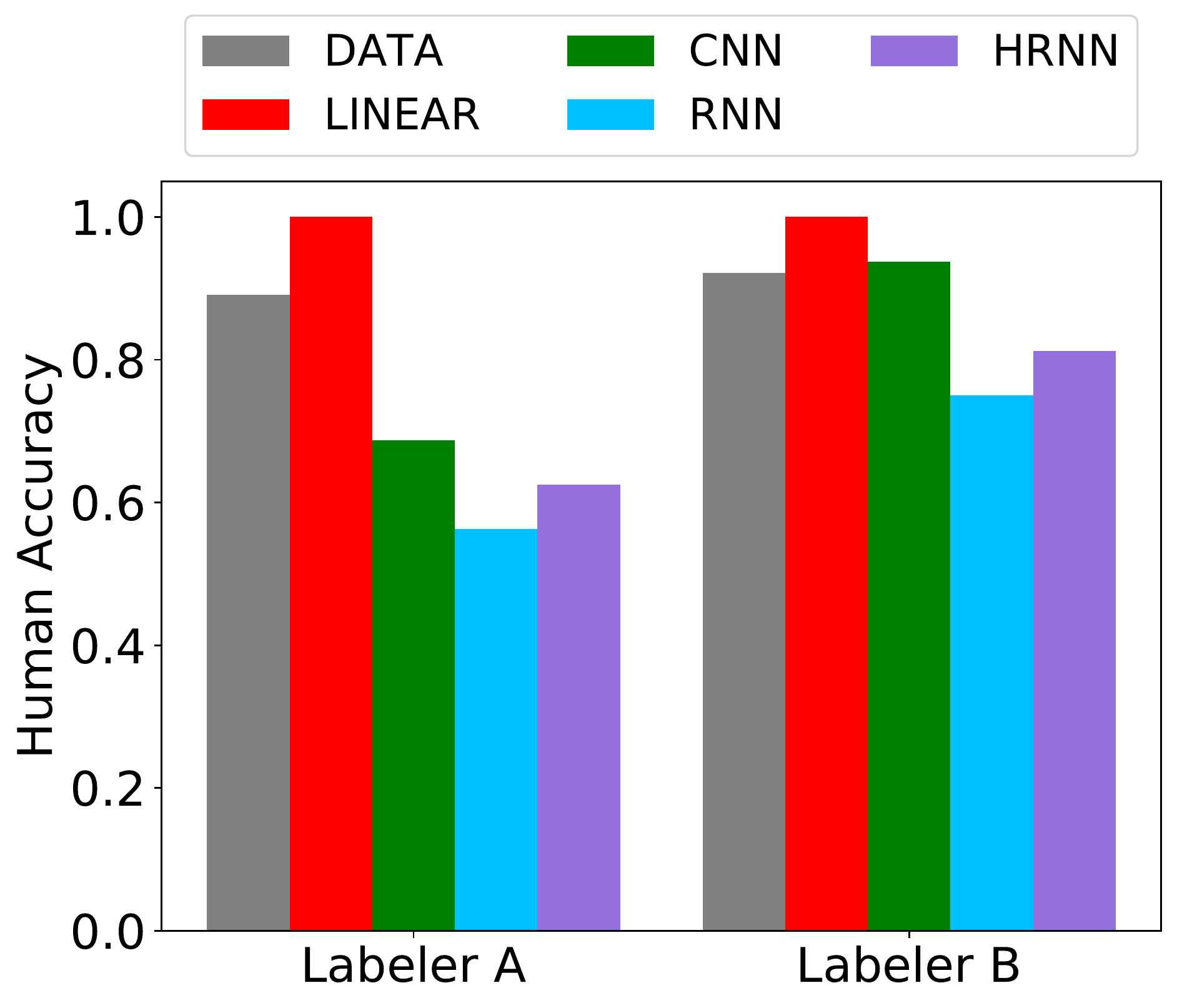}
        \vspace{-0.5cm}
        \subcaption{Biologist accuracy, 20 test flies}
        \label{fig:human_long}
    \end{minipage}
    
    \caption{Accuracy at discriminating real vs fake trajectories for R71G01 males. Low discriminator accuracy corresponds to better models. (a-b) show results for 60-frame trajectories with one test fly interacting with 19 real flies ($\mathcal{S}_{LOO}$). (c) shows results for longer, variable length trajectories with all 20 flies either simulated ($\mathcal{S}_{\text{SMSF}}$) or real. (a) shows results for a trained discriminator and (b-c) show results for two biologist experts.}
\end{figure}

Fig.~\ref{fig:nstep}(a) shows the 1-step, 1-sample centroid position prediction error, similar to what has been used in past work (Sec.~\ref{sec:related}). Even the trivial baseline models CONST and HALT perform well on 1-step predictions, and it is difficult to differentiate models. We propose measuring error farther into the future and considering multiple samples from the model ($n$-step, $m$-sample error, Fig.~\ref{fig:nstep}(b-c)). Using this metric, the deep learning models which look more realistic to biologists (Fig.~\ref{fig:human_long}), all learned models clearly outperform the trivial baselines, and the 3 deep networks clearly outperform the linear model. The ranking of RNN, HRNN, and CNN depends on the data set. 

\subsubsection{RvF Discriminators.} 
Figure~\ref{fig:disc_eval} shows the accuracy of the RvF discriminator. 0.5 is the optimal accuracy, meaning the discriminator can't tell real from fake. Our upper limit baseline is TRAIN DATA, where we train the discriminator on real data from different videos of the same genotype. We see that this is very close to the optimal accuracy. Our low baselines are data from other genotypes or sexes (R91B01, CONTROL, MALE-FEMALE). LINEAR performs close to these low baselines. the three deep network approaches all perform in between our high and low baselines. Here, CNNs slightly outperform RNN and HRNN, which are quite close. 

Biologists (Fig.~\ref{fig:human_short},~\ref{fig:human_long}) roughly match the rankings produced by the trained discriminator. Biologist accuracy is $\approx .675$. At this real accuracy, the discriminator's fake accuracy is slightly higher, indicating that the trained classifiers are slightly better at the discrimination task than expert humans. Biologists performed much better given all simulated or real flies (\ref{fig:human_long}). Their accuracy improved slightly for longer videos.

\begin{figure}[htp]
    \centering
    \begin{minipage}{0.19\textwidth}
        \includegraphics[width=\linewidth]{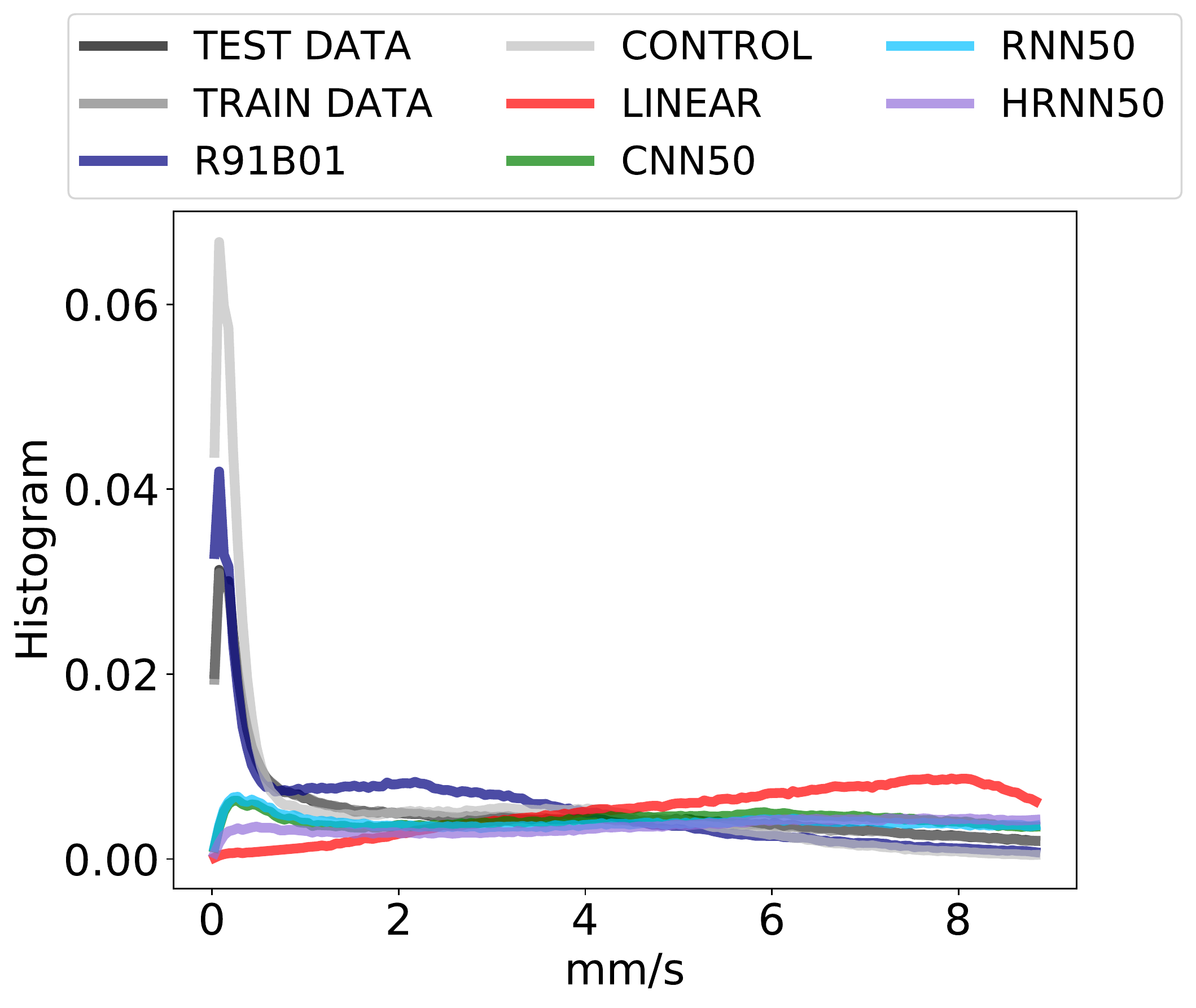}
        \vspace{-0.5cm}
        \subcaption{R71G01 Male \newline  Velocity}
    \end{minipage}
    \begin{minipage}{0.19\textwidth}
        \includegraphics[width=\linewidth]{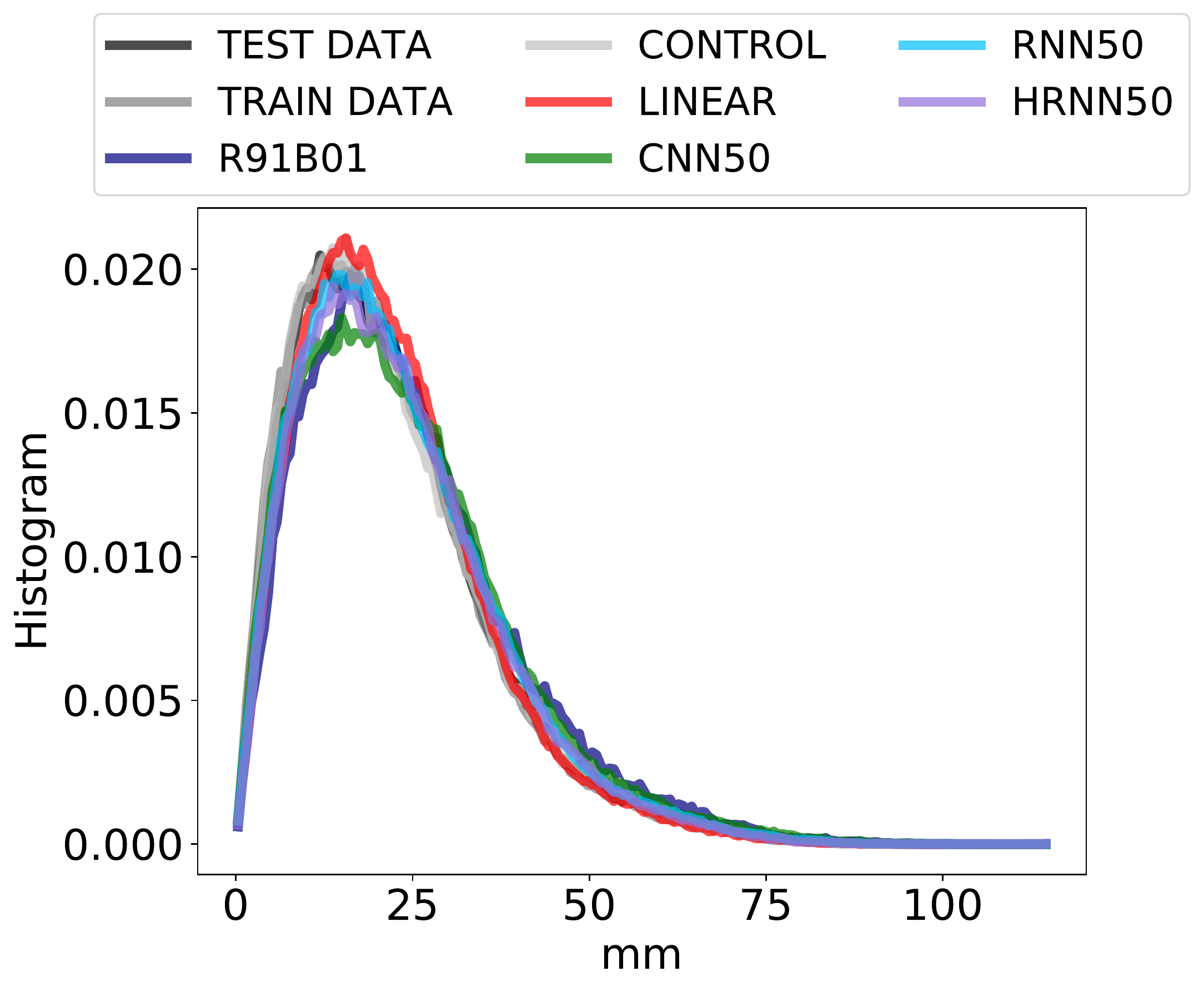}
        \vspace{-0.5cm}
        \subcaption{R71G01 Female\newline Inter Distance}
    \end{minipage}
    \begin{minipage}{0.19\textwidth}
        \includegraphics[width=\linewidth]{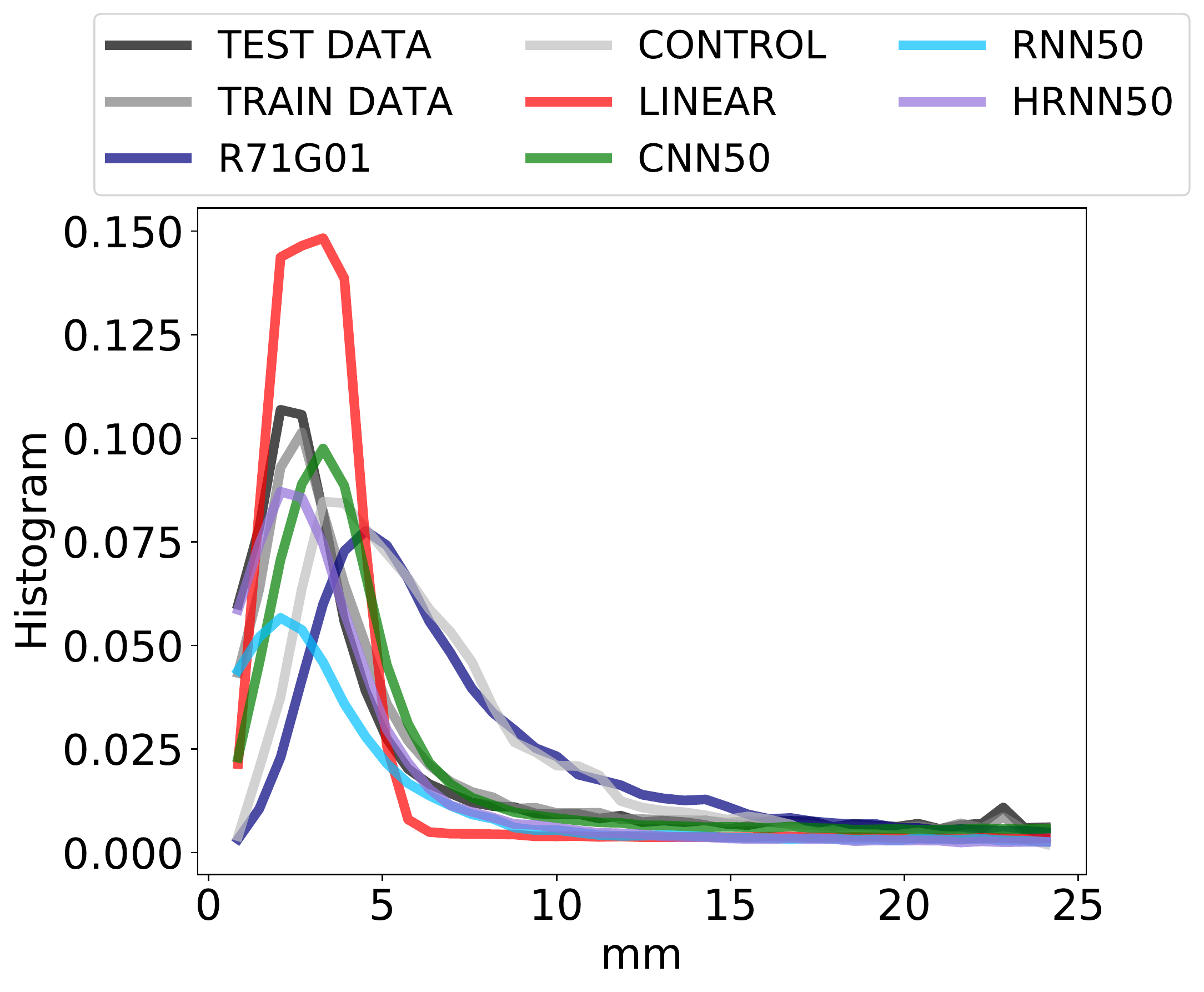}
        \vspace{-0.5cm}
        \subcaption{R91B01 Male\newline Wall Distance}
    \end{minipage}
    \begin{minipage}{0.19\textwidth}
        \includegraphics[width=\linewidth]{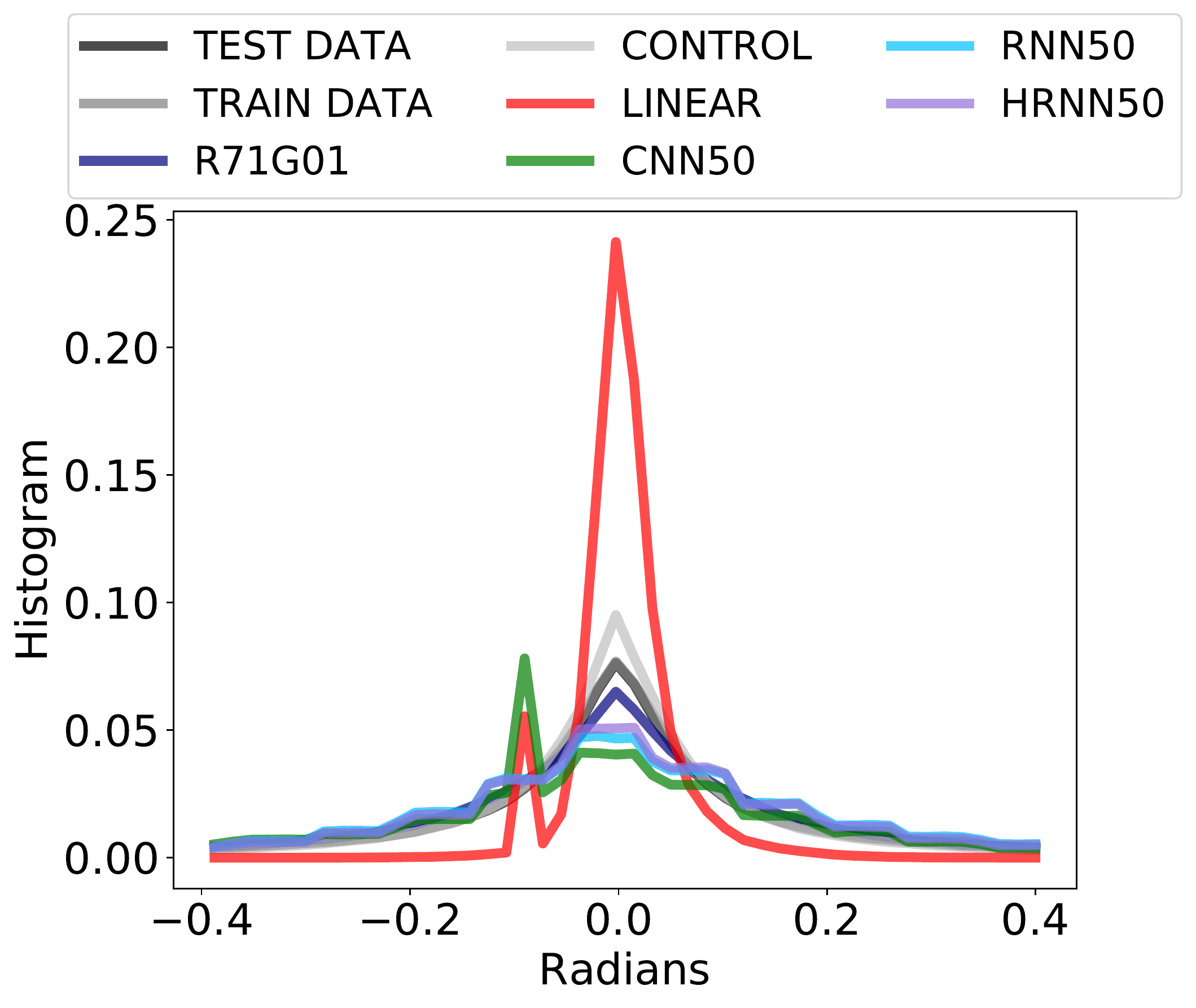}
        \vspace{-0.5cm}
        \subcaption{R91B01 Female\newline Angular Motion}
    \end{minipage}
    \begin{minipage}{0.19\textwidth}
        \includegraphics[width=\linewidth]{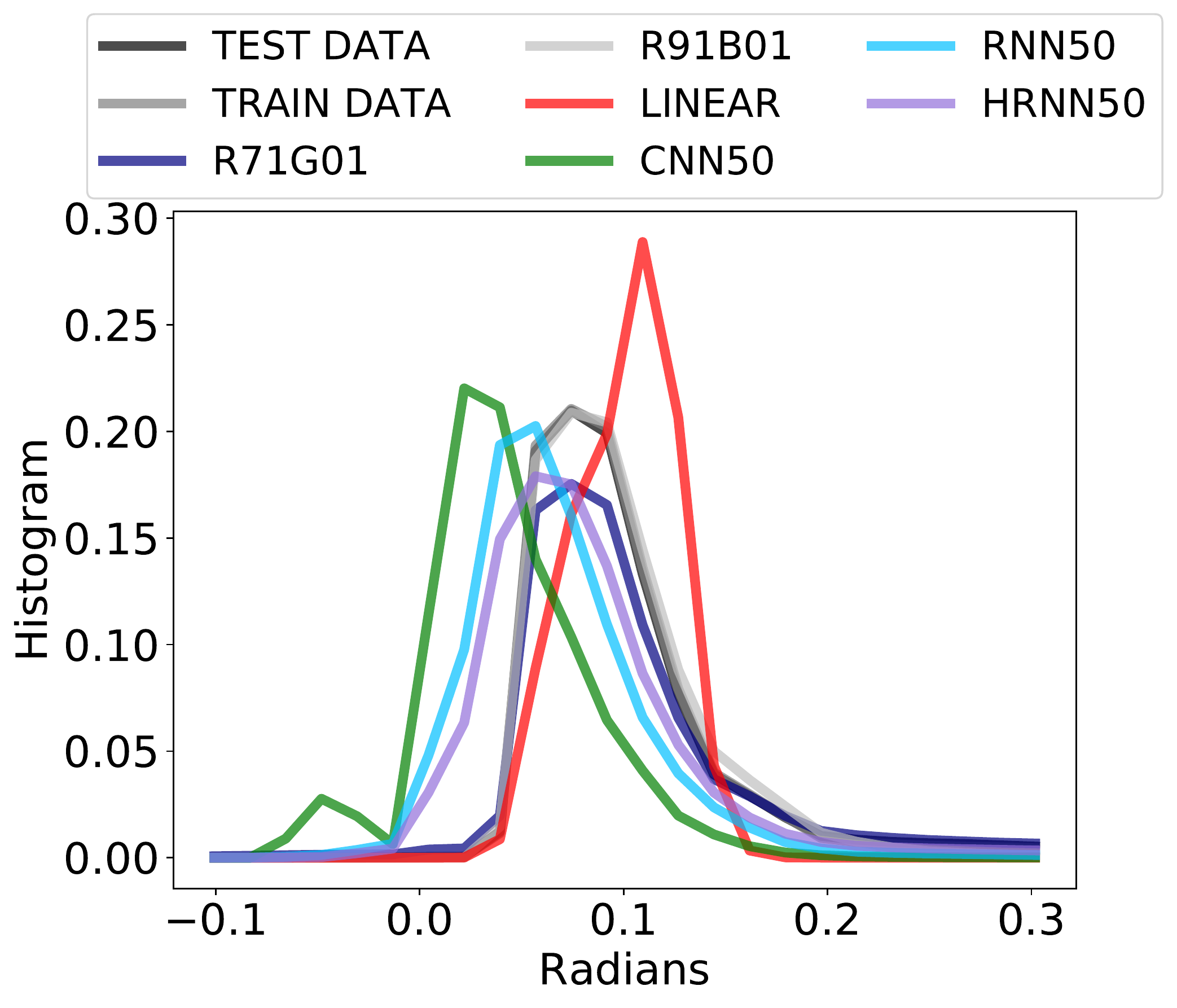}
        \vspace{-0.5cm}
        \subcaption{CONTROL\newline Wing Angle}
    \end{minipage}
    \caption{Distributions of selected behavior features. Bin widths are: (a) $0.05$mm/s, (b-c) $0.61$ mm, (d) $0.0175$ rad/fr, (e) $0.0175$ rad.  (see S.M.~Fig.~\ref{fig:histogram_feat_supp} for full datasets)}
    \label{fig:histogram_feat}
\end{figure}


\subsubsection{Histogram Distances.} 
Figure~\ref{fig:histogram_feat} presents the distributions for five behavioral features. For reference, we added the distributions for different videos of the same genotype and sex (TRAIN DATA), and confirm that training and test distributions are very close. It is clear that LINEAR fails to capture important components of the distribution of behavior. L1 distances between the histograms are shown in Table~\ref{tab:histo_dist_gmr71}, and ranking between deep networks depends on which feature histogram we examine. All networks show similar qualitative biases, particularly for the speed. Here, we see that none of our models stop enough (velocity $\approx 0$). 

These results were for simulations in which all flies were simulated, both male and female ($\mathcal{S}_{\text{SMSF}}$). For comparison, S.M. Figure~\ref{fig:simulation} histogram distances when only one fly was simulated ($\mathcal{S}_{LOO}$), when males were real and females simulated ($\mathcal{S}_{\text{RMSF}}$), when males were simulated and females real ($\mathcal{S}_{\text{SMRF}}$). Interestingly, we see big improvements for velocity and angular motion only for $\mathcal{S}_{LOO}$. This indicates that some of the deficiencies in replicating the full distribution of behaviors comes from interactions of the multi-agent simulations.


\begin{table}
    \vspace{-0.5cm}
  \centering
  \begin{minipage}{0.49\textwidth}
  \caption{Chasing frequency for male flies.}
  \label{tab:chasing_eval}
  \centering
  \begin{tiny}
  \begin{tabular}{|l|c|c|c|c|c|}\hline
      & DATA & LINEAR & CNN & RNN & HRNN\\\hline
      R71G01 & 0.23      & {\bf 0.165}   & 0.051         & 0.153     & 0.15 \\
      R91B01 & 0.00027   & 0.00064       & 0.0004        & 0.00038   & {\bf 0.00037} \\
      CONTROL & 0.00043   & 0.03614      & 0.0022  & {\bf 0.00404}   & 0.00256 \\\hline
  \end{tabular}
  \end{tiny}
  \end{minipage}
  \begin{minipage}{0.49\textwidth}
  \centering
  \caption{Chase classifier error.}
  \label{tab:chasing_eval2}
  \begin{tiny}
  \begin{tabular}{|l|c|c|c|c|}\hline
       & LINEAR & CNN & RNN & HRNN\\\hline
      R71G01 & {\bf 0.014}      & 0.185   & 0.0203 & 0.0222  \\
      R91B01 & {\bf 1.37e-05}   & 2.388e-05       & 1.773e-05        & 1.493e-05 \\
      CONTROL & 0.0351   &  {\bf 0.00133}       &  0.00301 & 0.001623 \\\hline
  \end{tabular}
  \end{tiny}
  \end{minipage}
  \vspace{-0.25cm}
\end{table}
\subsubsection{Behavior Classifier Error.}
Table~\ref{tab:chasing_eval} shows the fraction of time real and simulated male flies from each genotype chase. R71G01 male flies chase a large fraction of the time, while control and R91B01 almost never do. The RNN and HRNN are the only two models to capture this well. Surprisingly, the LINEAR R71G01 model produces chasing flies a large fraction of the time, but also produces CONTROL flies that chase a significant fraction of the time.

{\em Summary.} Table~\ref{tab:histo_dist_gmr71_male} shows the performance of each of the models for each of the proposed metrics. We see that all of our proposed metrics place the deep networks between low baselines (data from other genotypes and sexes, trivial, unlearned models) and our high baseline (data from videos of the same type of fly). Furthermore, they prefer the deep networks to the LINEAR model, matching biologists' performance in RvF discrimination. 

\begin{table*}[htp]
    \caption{Performance under various metrics for R71G01 male (see S.M.~Table.~\ref{tab:histo_dist_gmr91} for other datasets). Lower values indicate better models, and we indicate the best model in bold. Red indicates baselines that beat all models.}
    \label{tab:histo_dist_gmr71}
  \centering
  \label{tab:histo_dist_gmr71_male}
    {\tiny
  \begin{tabular}{|l|l|c|ccc|cc|cccc|}\hline
                &               & Upper baseline & \multicolumn{3}{|c|}{Lower baseline} & \multicolumn{2}{|c|}{Fixed models} & \multicolumn{4}{|c|}{Learned models}\\
                &               & TRAIN & R91B01 & CONTROL & FEMALE & CONST & HALT  & LINEAR & CNN  & RNN  & HRNN \\\hline\hline
      \multirow{2}{*}{Biologist} & LOO & -     & -      & -       &-            & -     & -     & 65.6 & {\bf 56.8} & 61.8 & 65.6 \\
                & SMSF & -     & -      & -       &-            & -     & -     & 95.3    & 85.9 & {\bf 78.1} & 81.2 \\ \hline
      -loglik      &       & -     & -      & -       &-            & -     & -     & -      & 47559& {\bf 460101} & 46116 \\ \hline
      1-step   &       & -     & -      & -       &-            & 5.04  & 7.73  & 2.90   & 3.04 & {\bf 2.78} & 2.89 \\
      30-step   &       & -     & -      & -       &-            & 97.7 & 97.6  & 54.6   & 26.9 & {\bf 20.7} & 21.5\\\hline
      RvF      &      & 51.1  & 91.5   & 78.2    & 83.3        & -     & -     & 78.9   & {\bf 62.2}  & 65.1  & 71.1   \\\hline
      Chase                     &  &  -  & -      & -       & -           & -     & -    & {\bf 0.014} & 0.185 & 0.020 & 0.022 \\\hline
      \multirow{4}{*}{HD}       & Vel.          & 0.028 & {\color{red} 0.353}  & 0.421   & -           & -     & -     & 0.994  & {\bf 0.354} & 0.476 & 0.729\\
                                & Inter. Dist.  & 0.047 & 0.136  & 0.071   & -           & -     & -     & 0.083  & 0.078 & 0.071 & {\bf 0.050}\\
                                & Wall Dist.    & 0.044 & 0.753  & {\color{red} 0.209}   & -           & -     & -     & 0.440  & 0.477 & 0.455 & {\bf 0.320}\\
                                & Ang. Mot.     & 0.015 & {\color{red} 0.237}  & 0.271   & -           & -     & -     & 0.917  & 0.652 & 0.455 & {\bf 0.310}\\\hline
  \end{tabular}
    }
    \vspace{-0.5cm}
\end{table*}

\subsection{Interpretation of Results}
\label{sec:interpretation}

We developed the evaluation metrics described above to try to understand whether a given model is sufficiently accurate such that an understanding of it will lead to insight into animal behavior. Do any of our models achieve this accuracy? Which ones? If not, how do we improve them? 

We observe that all the models have high error in predicting position 1 second (30 frames) into the future, and error accumulates linearly the farther into the future we predict (Figure~\ref{fig:nstep}(c)). However, trained discriminators and expert biologists are fairly poor (though above chance) at discriminating real from simulated trajectories (Figure~\ref{fig:disc_eval},~\ref{fig:human_short}). Our results suggest that poor $n$-step prediction error is due to the inherent unpredictability of behavior. Evidence for this is that increasing the number of samples in the $n$-step error is much more important for the multi-modal deep network models than the Gaussian LINEAR model (Figure~\ref{fig:nstep}(d)). We thus conclude that low $n$-step error prediction is not a necessity for a model to be sufficiently accurate. 

Salient differences between the distributions of behavior features for real and simulated trajectories (Figure~\ref{fig:histogram_feat}) suggest that there is still room for improvement with learning accurate models. We believe performance for these metrics can best be improved by modifying the loss function used to train the network. In particular, for the velocity histogram, we believe that too much emphasis is placed on precisely predicting the velocity when a fly is moving quickly, instead of differentiating between still and slow movements. 

We consistently see that deep networks outperform the linear model. However, ranking of the CNN, RNN, and HRNN are inconsistent. We thus believe that there is not a huge difference between these models in accuracy, and these differences in network architecture are not important. Thus contradicts results in \cite{Eyrun2016}, which argued that HRNN outperformed RNN. We believe that better avenues for improving accuracy are different loss functions and hard-example mining. 

\section{Conclusions}

In this paper, we presented several new criteria for evaluating models of animal behavior trained using unsupervised learning. We show that different metrics evaluate different properties of the model, and results from all can be interpreted together to understand how the model differs from real animals, and how we can improve the models. Our results indicate that, between deep networks, architectural changes have minor effects, and instead point us in the direction of developing different loss functions that change how we weigh different types of errors or types of examples. While the current sota models mimic animal behavior closely enough to fool expert biologists and trained discriminator networks a large fraction of the time, there is room for improvement, and qualitative differences in the histograms indicate that more work on training models is necessary. In addition to guiding improvements to model accuracy, we believe these metrics will be essential in measuring effects on accuracy of restricting models to be more interpretable or biologically feasible. 


\bibliography{main}
\bibliographystyle{plain}
\clearpage
\begin{appendices}
\section{Supplementary Materials}
\label{app}

\subsection{Model Details}
\label{sec:models}

\subsubsection{Architecture} 

We compared four different learned models: linear regression (LINEAR), 1-dimensional over time convolutional networks (CNN), GRU recurrent neural networks (RNN), and the hierarchical GRU recurrent neural network structure proposed in \cite{Eyrun2016} (HRNN). 

\begin{figure}[htp] 
    \centering
    \begin{minipage}{0.23\textwidth}
        \includegraphics[width=\linewidth]{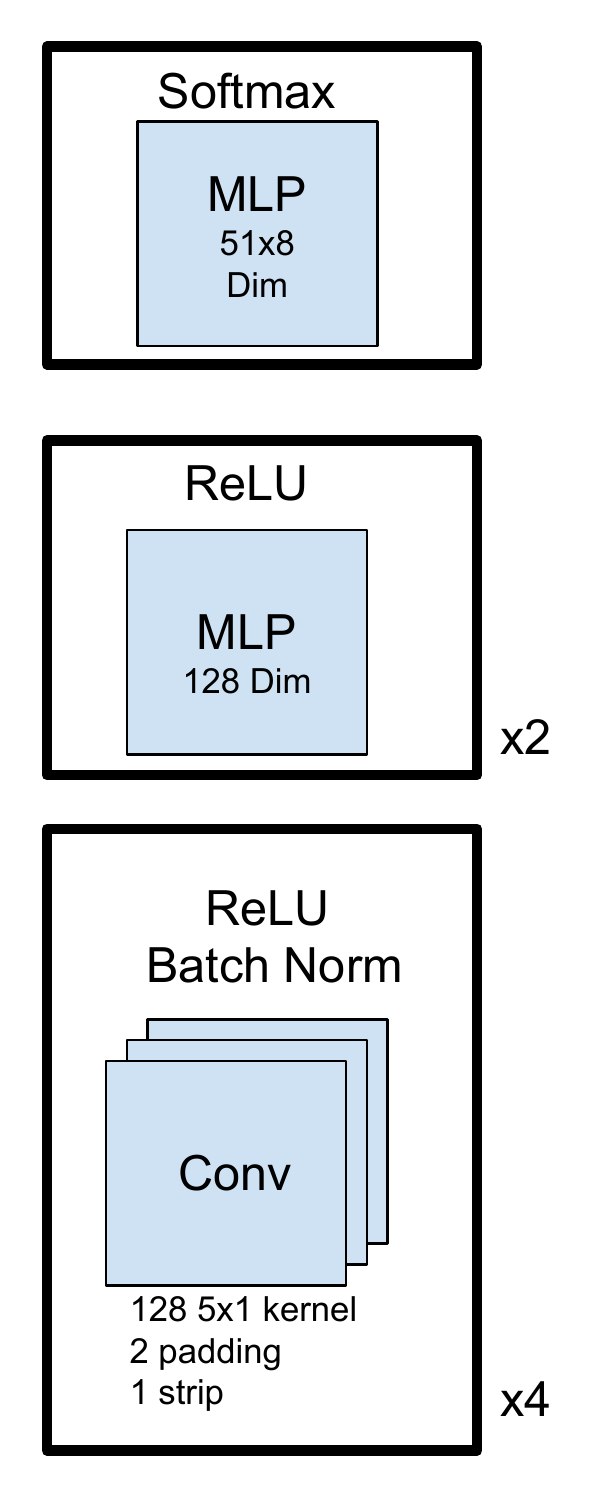}
        \subcaption{CNN}
    \end{minipage}
    \begin{minipage}{0.3\textwidth}
        \includegraphics[width=\linewidth]{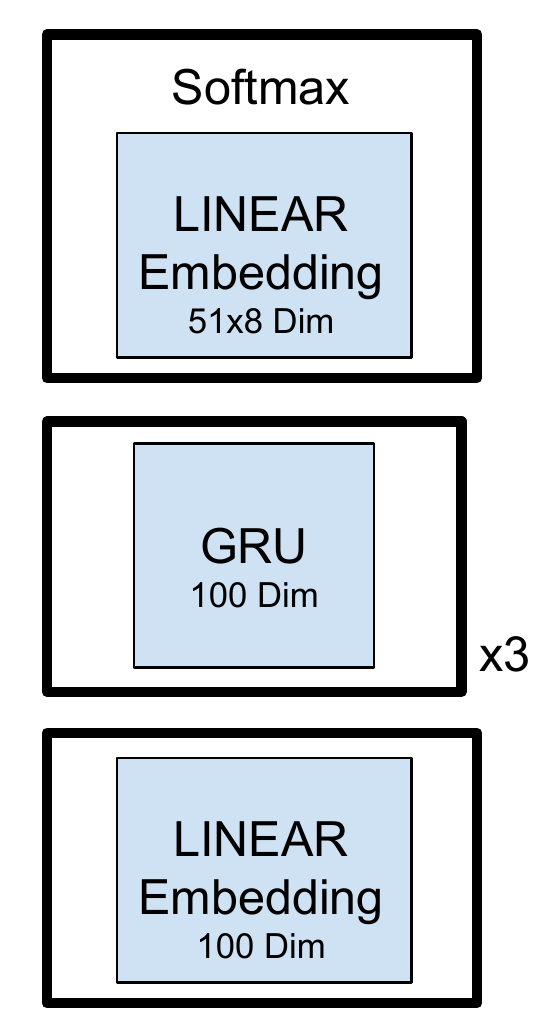}
        \subcaption{RNN}
    \end{minipage}
    \begin{minipage}{0.45\textwidth}
        \includegraphics[width=\linewidth]{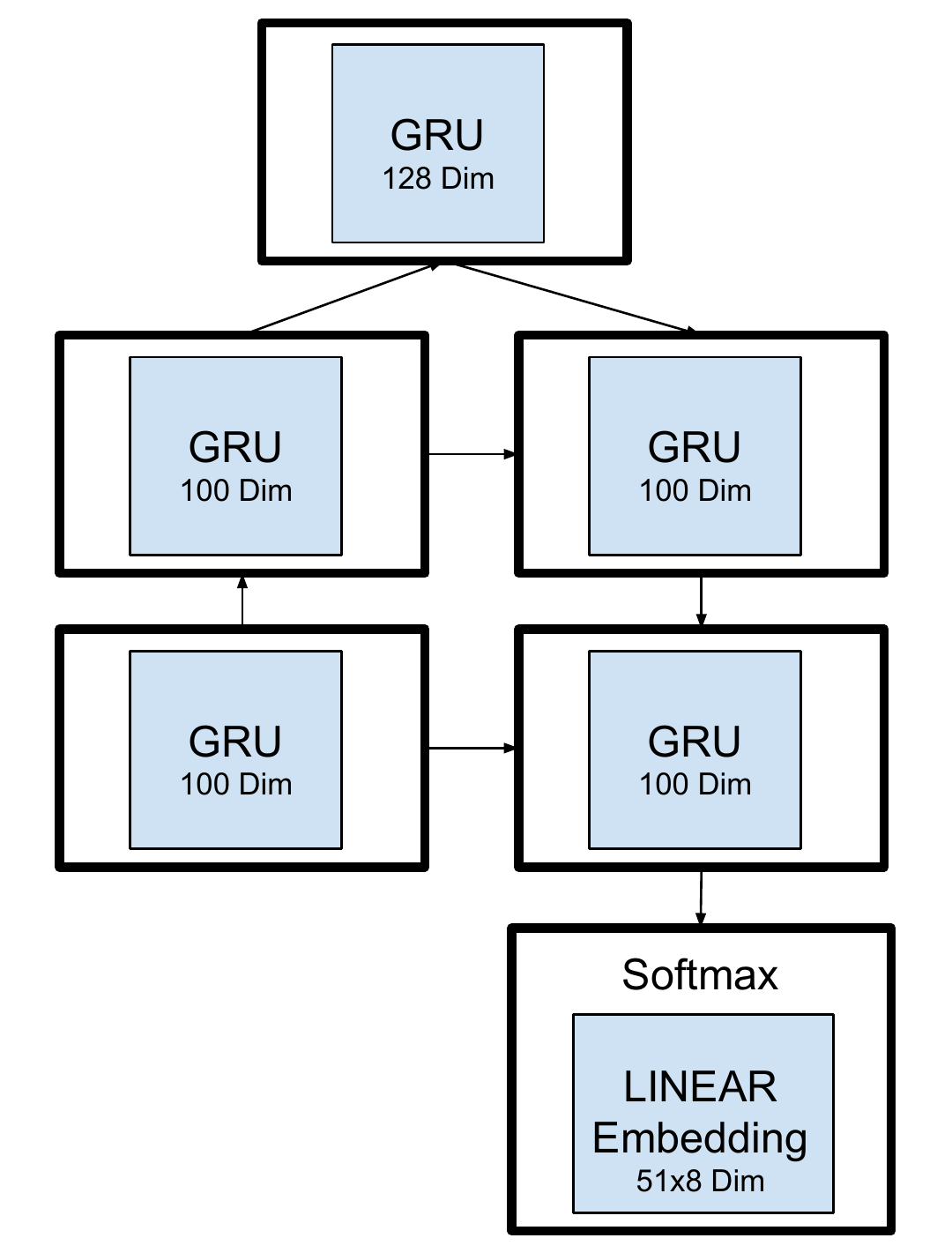}
        \subcaption{HRNN}
    \end{minipage}
    \caption{Architecture Detail}
    \label{fig:architecture}
\end{figure}

The CNN had four convolutional layers with $128$ $1 \times 5$ filters, one stride, and
padding size of two, followed by two $128$-dimensional fully connected layers. 
The RNN had a $100$-dimensional embedding layer, followed by three $100$-dimensional GRU 
layers, and a final softmax layer (See S.M. Figure~\ref{fig:architecture}). 
Following~\cite{Eyrun2016}, the HRNN consists of a $100$-d linear embedding layer, three $100$-d and two $200$-d GRU layers, and a final softmax layer, with skip connections between certain layers (S.M.~Fig.~\ref{fig:architecture}).

All deep networks were trained to maximize log-likelihood of the training data, using the 51-bin representation described in Section~\ref{sec:representation}. Linear regression was trained to predict the continuous-valued trajectory features. To obtain a probabilistic model, we fit a Gaussian using the model error to set the standard deviation. 

Linear regression and CNN input motion features for the past $\tau = 50$ frames, ${\bf x}_{t-\tau:t}$. To match, RNN and HRNN were trained on intervals of length $\tau = 50$. 

\subsubsection{Training}

For each iteration of training, we sample a new batch of 32 $\tau=50$ frame intervals. When sampling, we 
first uniformly sample a training video, then sample a fly and frame uniformly (thus each video, and each fly-frame within a video, have equal representation in the training data. This detail was necessary because of missing data due to  tracking errors.
We used simple stochastic gradient descent with $0.01$ learning rate and L2 weight decay with $0.0001$ coefficient. 
The learning rate and L2 weight decay coefficient are chosen using 
grid search of $[0.2,0.1,0.07,0.01,0.007]$ and $[0, 0.00001, 0.0001]$ respectively.
We considered $\lbrace 50000, 75000, 100000, 150000, 200000\rbrace$ epochs for RNN and HRNN,
and considered $\lbrace 15000, 20000, 25000, 30000\rbrace$ epochs for CNN.
We selected number of epochs based on histogram distance on validation data.

\subsubsection{Vision-Chamber Data}
\begin{figure}[htp] 
    \centering
    \begin{minipage}{0.49\textwidth}
        \includegraphics[width=\linewidth]{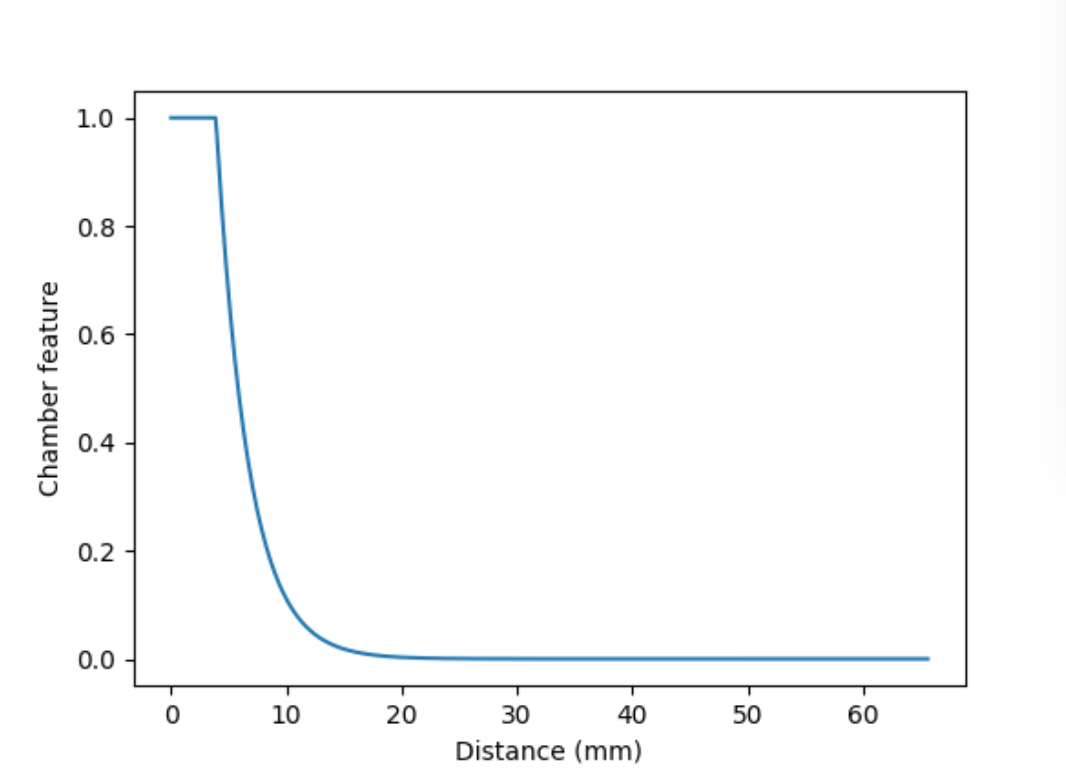}
        \subcaption*{Vision}
    \end{minipage}
    \begin{minipage}{0.49\textwidth}
        \includegraphics[width=\linewidth]{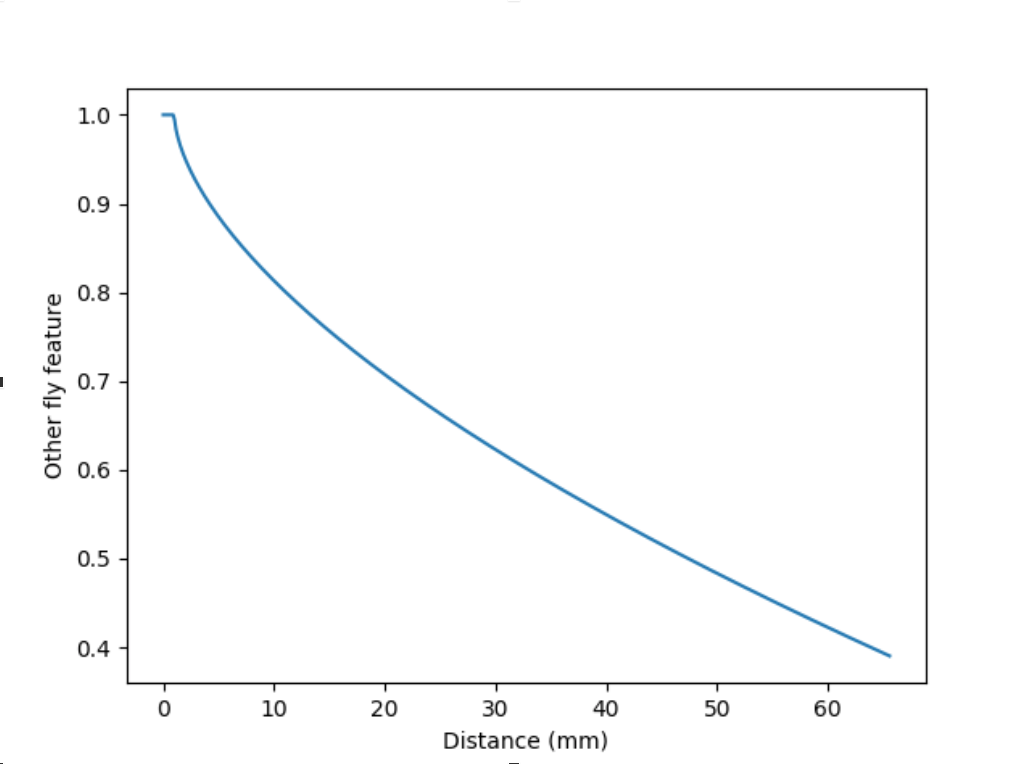}
        \subcaption*{Chamber}
    \end{minipage}
    \caption{Non-linear function applied to vision and chamber}
    \label{fig:vision_chamber_nonlinear}
\end{figure}

The non-linear function applied to vision and chamber distance $d$ are
    \begin{align}
        f_{\text{vision}}(d) &= 1-\text{min} (1, 0.05 \text{max}(0, d-1)^{0.5})\\
        f_{\text{chamber}}(d) &= \text{min} (1, 1.3(0.4 (d-const)) /2 )
    \end{align}
where $const = 3.885505$ \cite{Eyrun2016}.
Figure~\ref{fig:vision_chamber_nonlinear} displays vision and chamber non-linear functions.

Here are the model architectures that we used for CNN, RNN, and HRNN.
Each black box refer to as one block and $\times k$ indicates number of 
layers of blocks.

\clearpage
\subsection{S.M. for Short Term Performance Evaluation}

\begin{figure}[htp] 
    \begin{minipage}{\textwidth}
    \centering
        \begin{minipage}{0.24\textwidth}
            \includegraphics[width=\linewidth]{./figs/eval_30steps_pos_gender0_gmr.pdf}
            \vspace{-0.5cm}
            \subcaption*{Position, Male}
        \end{minipage}
        \begin{minipage}{0.24\textwidth}
            \includegraphics[width=\linewidth]{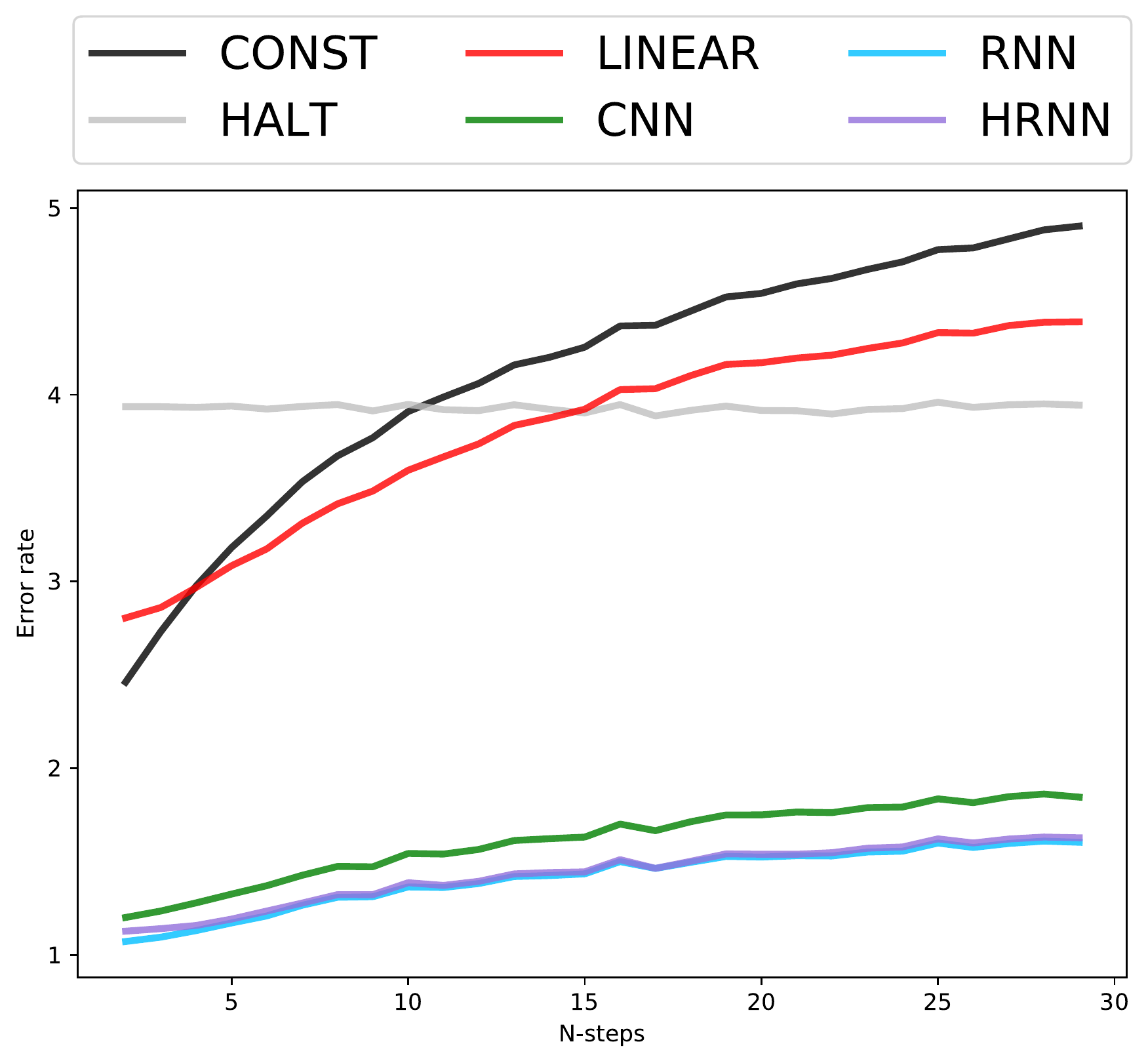}
            \vspace{-0.5cm}
            \subcaption*{Velocity, Male}
        \end{minipage}
        \begin{minipage}{0.24\textwidth}
            \includegraphics[width=\linewidth]{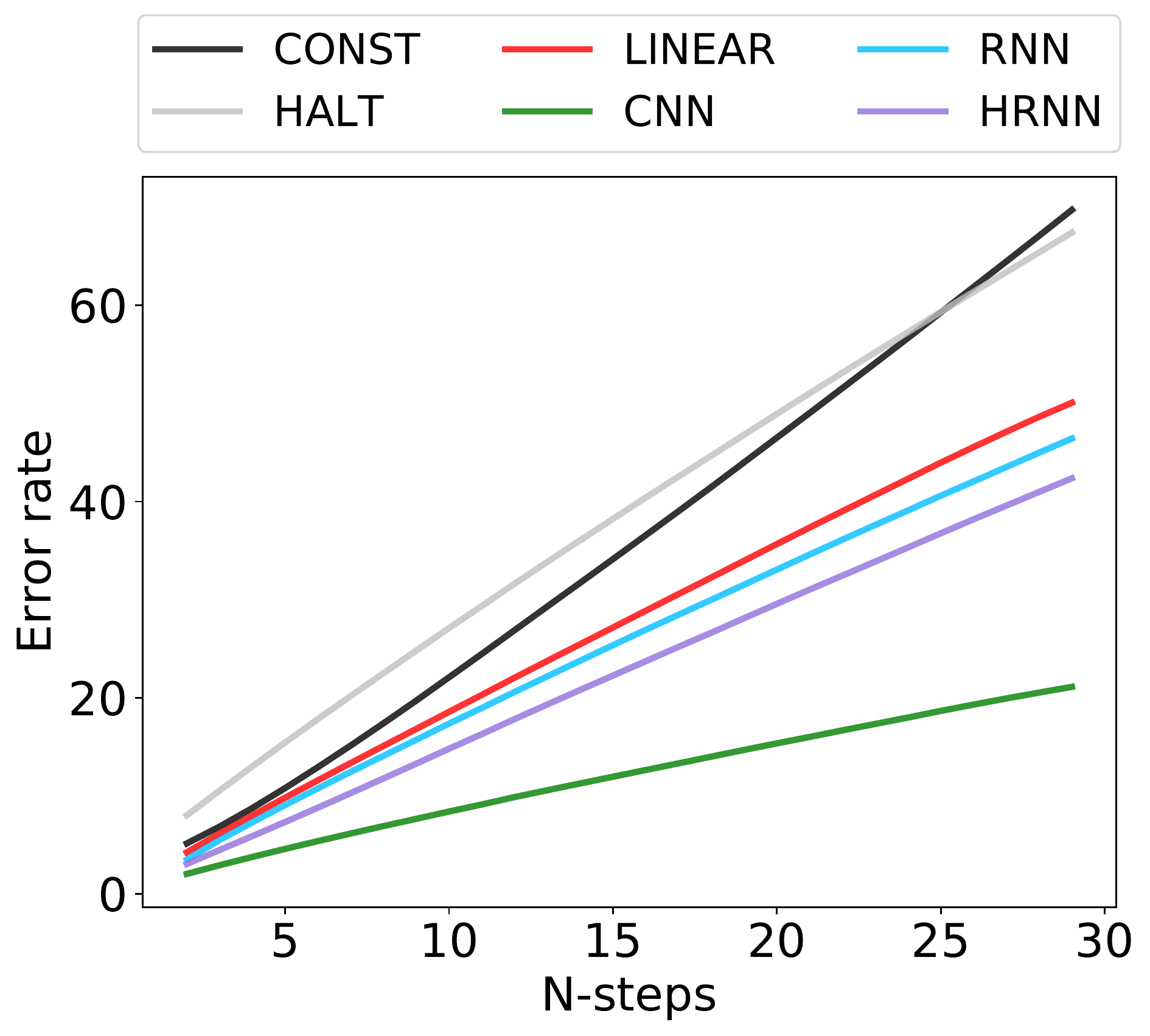}
            \vspace{-0.5cm}
            \subcaption*{Position, Female}
        \end{minipage}
        \begin{minipage}{0.24\textwidth}
            \includegraphics[width=\linewidth]{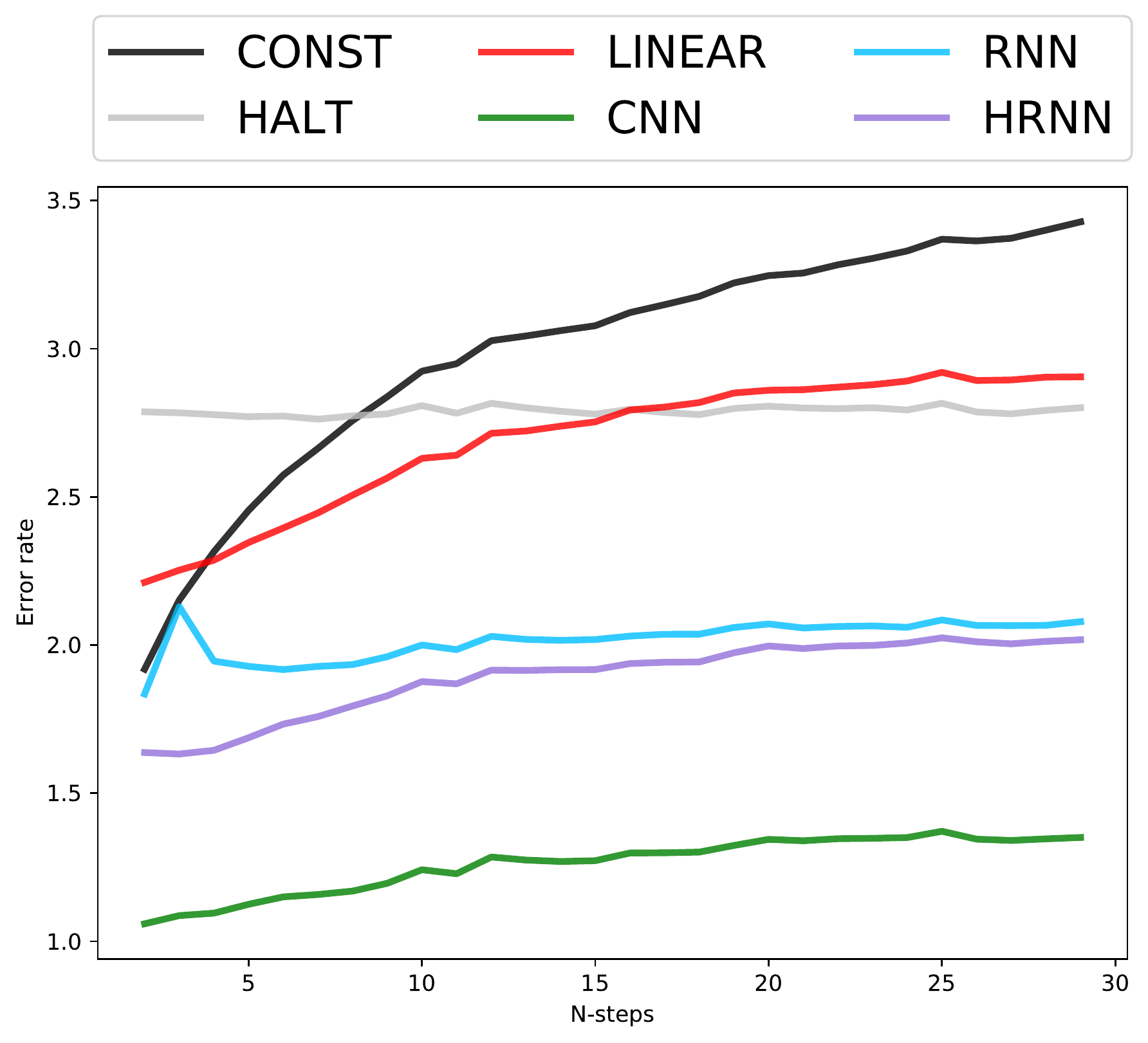}
            \vspace{-0.5cm}
            \subcaption*{Velocity, Female}
        \end{minipage}
        \subcaption{R71G01}
    \end{minipage}\\
    \begin{minipage}{\textwidth}
        \centering
        \begin{minipage}{0.24\textwidth}
            \includegraphics[width=\linewidth]{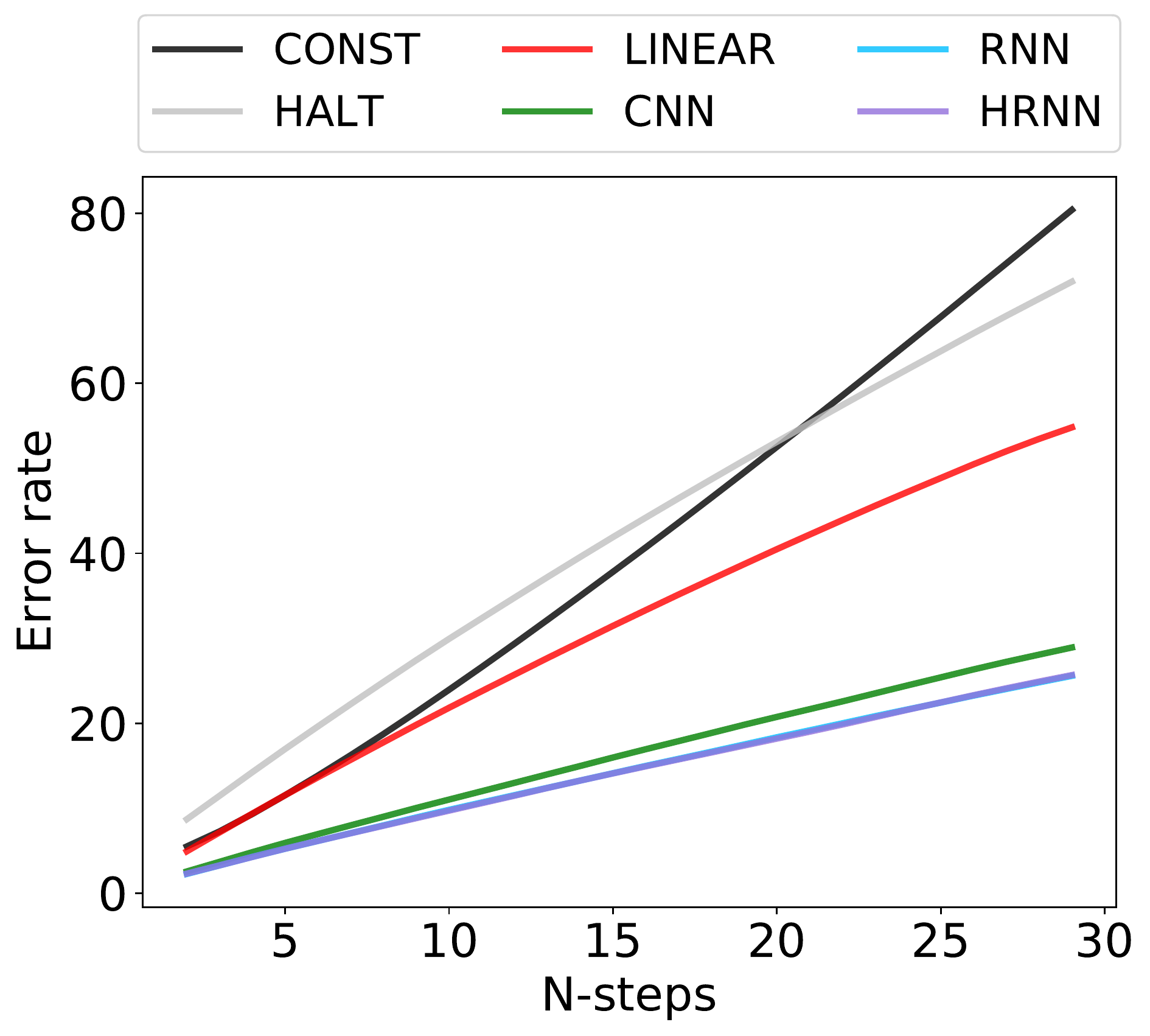}
            \vspace{-0.5cm}
            \subcaption*{Position, Male}
        \end{minipage}
        \begin{minipage}{0.24\textwidth}
            \includegraphics[width=\linewidth]{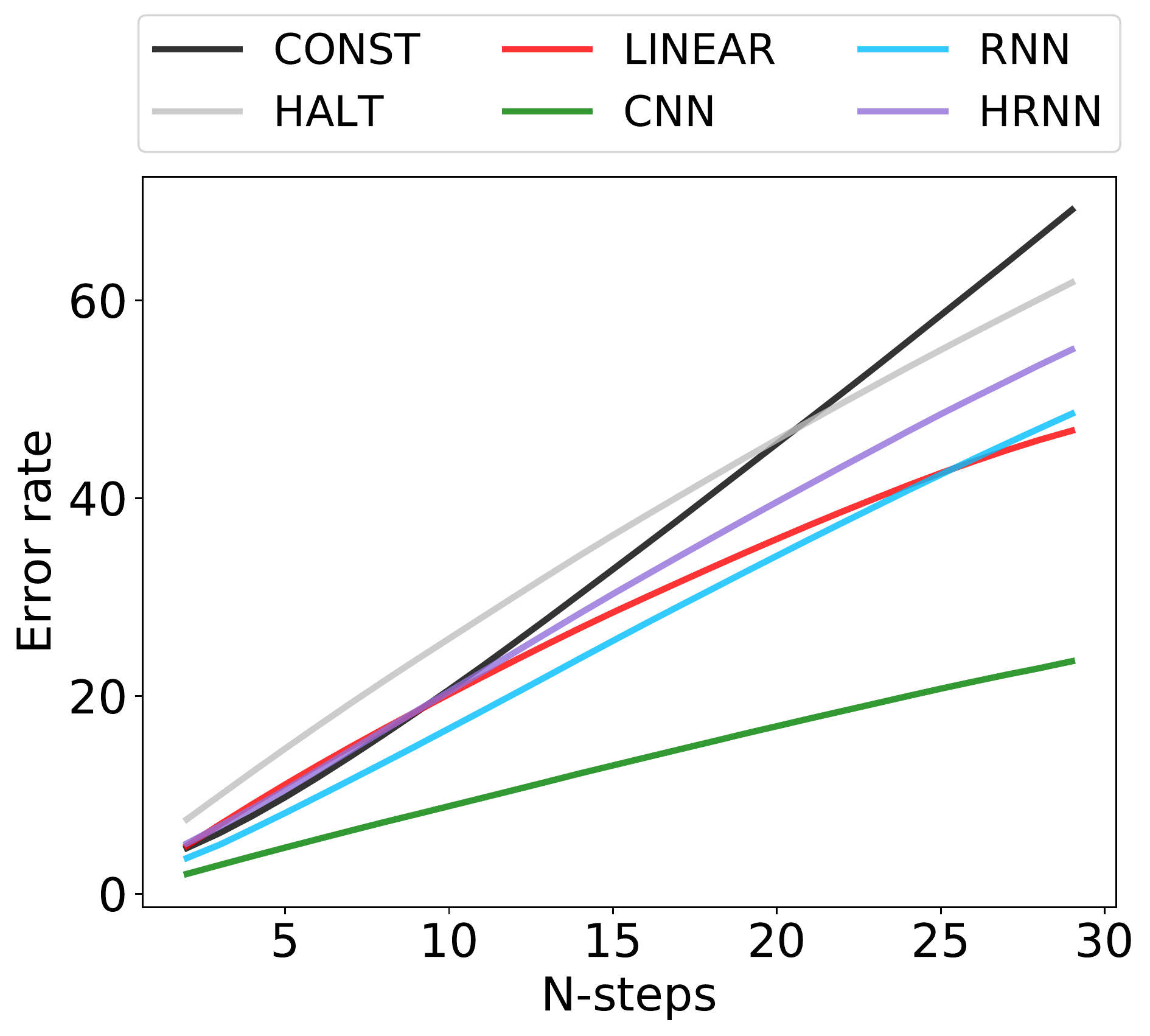}
            \vspace{-0.5cm}
            \subcaption*{Position, Female}
        \end{minipage}
        \begin{minipage}{0.24\textwidth}
            \includegraphics[width=\linewidth]{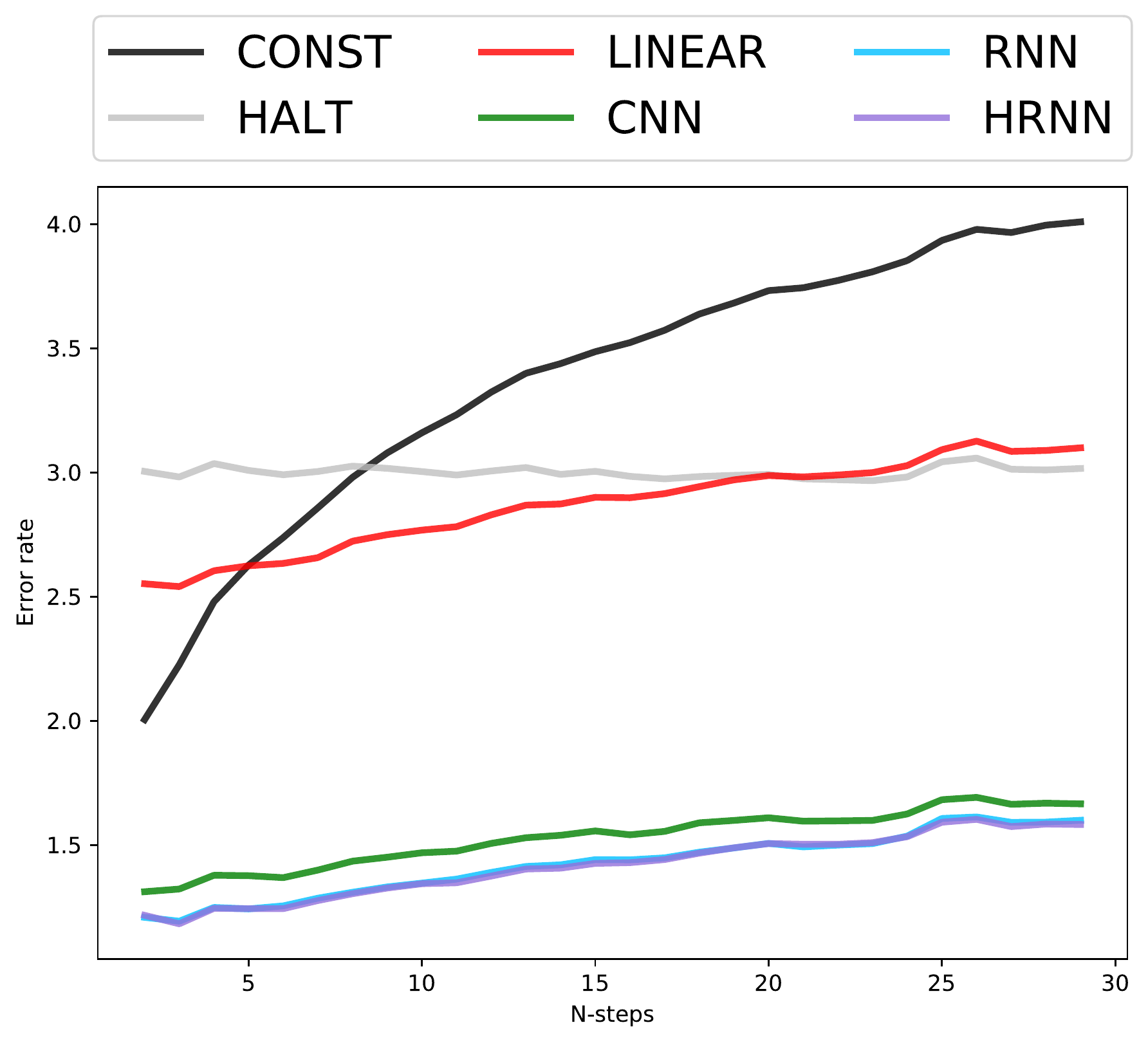}
            \vspace{-0.5cm}
            \subcaption*{Velocity, Male}
        \end{minipage}
        \begin{minipage}{0.24\textwidth}
            \includegraphics[width=\linewidth]{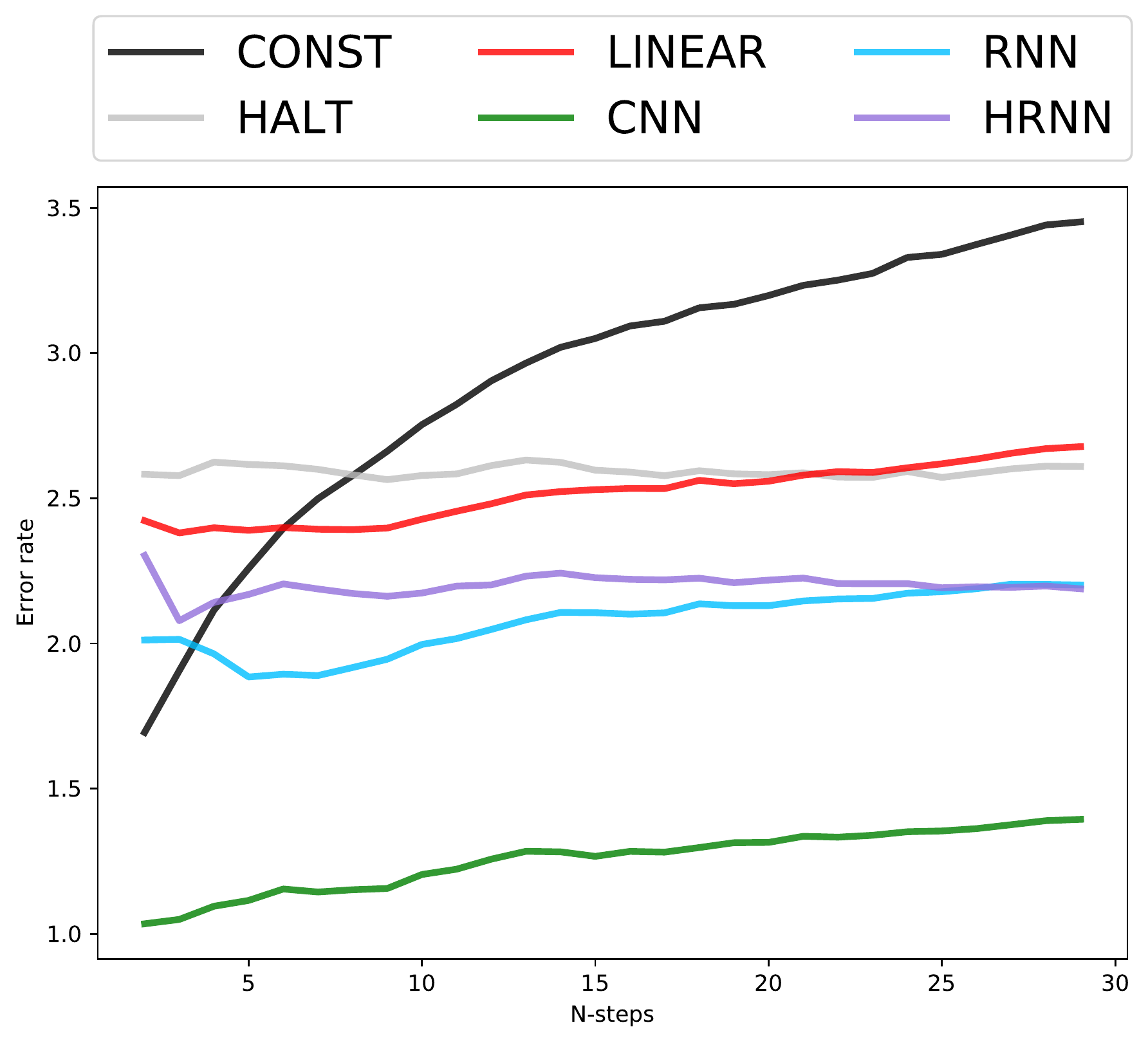}
            \vspace{-0.5cm}
            \subcaption*{Velocity, Female}
        \end{minipage}\\
        \subcaption{R91B01}
    \end{minipage}
    \begin{minipage}{\textwidth}
        \begin{minipage}{0.24\textwidth}
            \includegraphics[width=\linewidth]{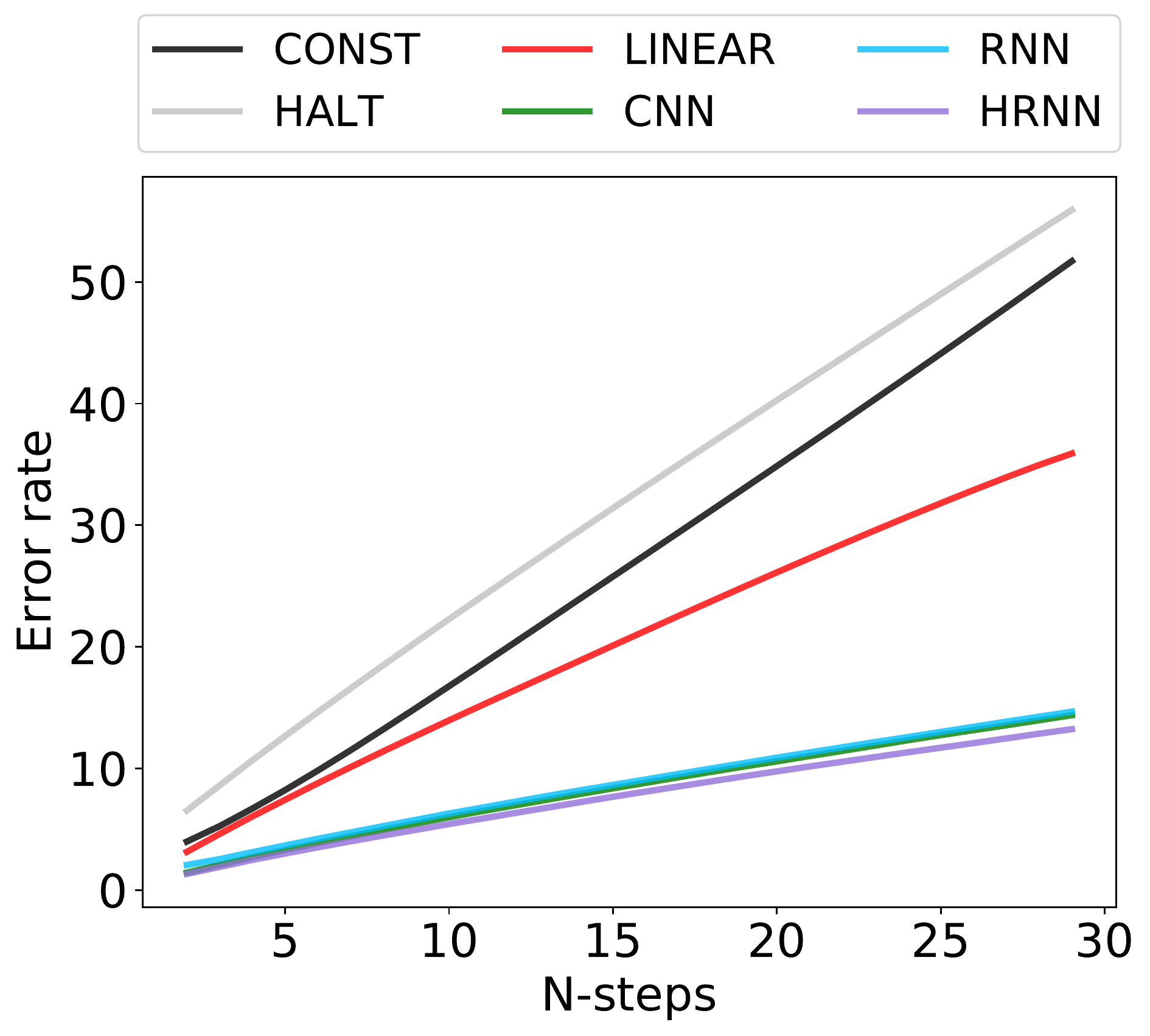}
            \vspace{-0.5cm}
            \subcaption*{Position, Male}
        \end{minipage}
        \begin{minipage}{0.24\textwidth}
            \includegraphics[width=\linewidth]{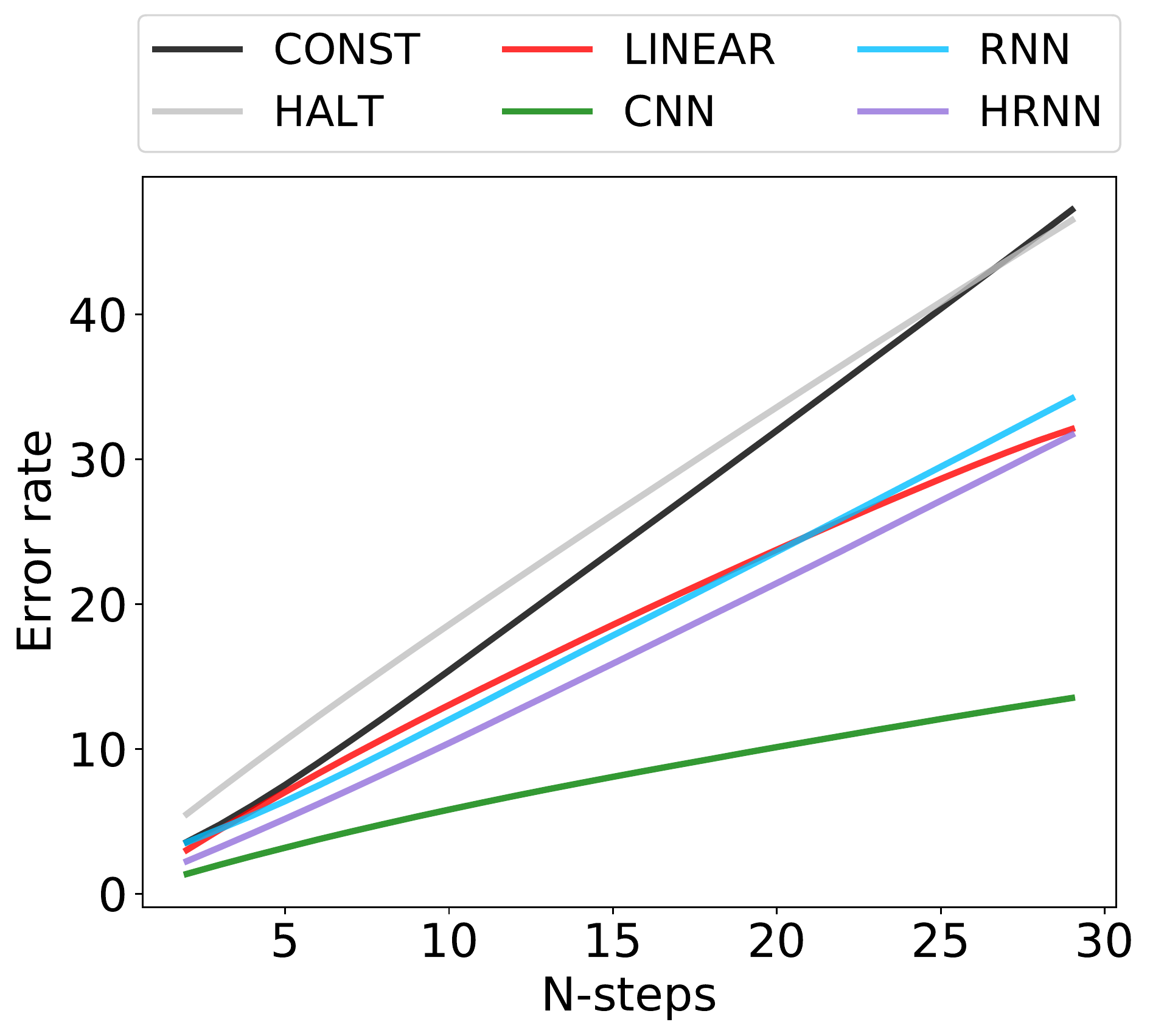}
            \vspace{-0.5cm}
            \subcaption*{Position, Female}
        \end{minipage}
        \begin{minipage}{0.24\textwidth}
            \includegraphics[width=\linewidth]{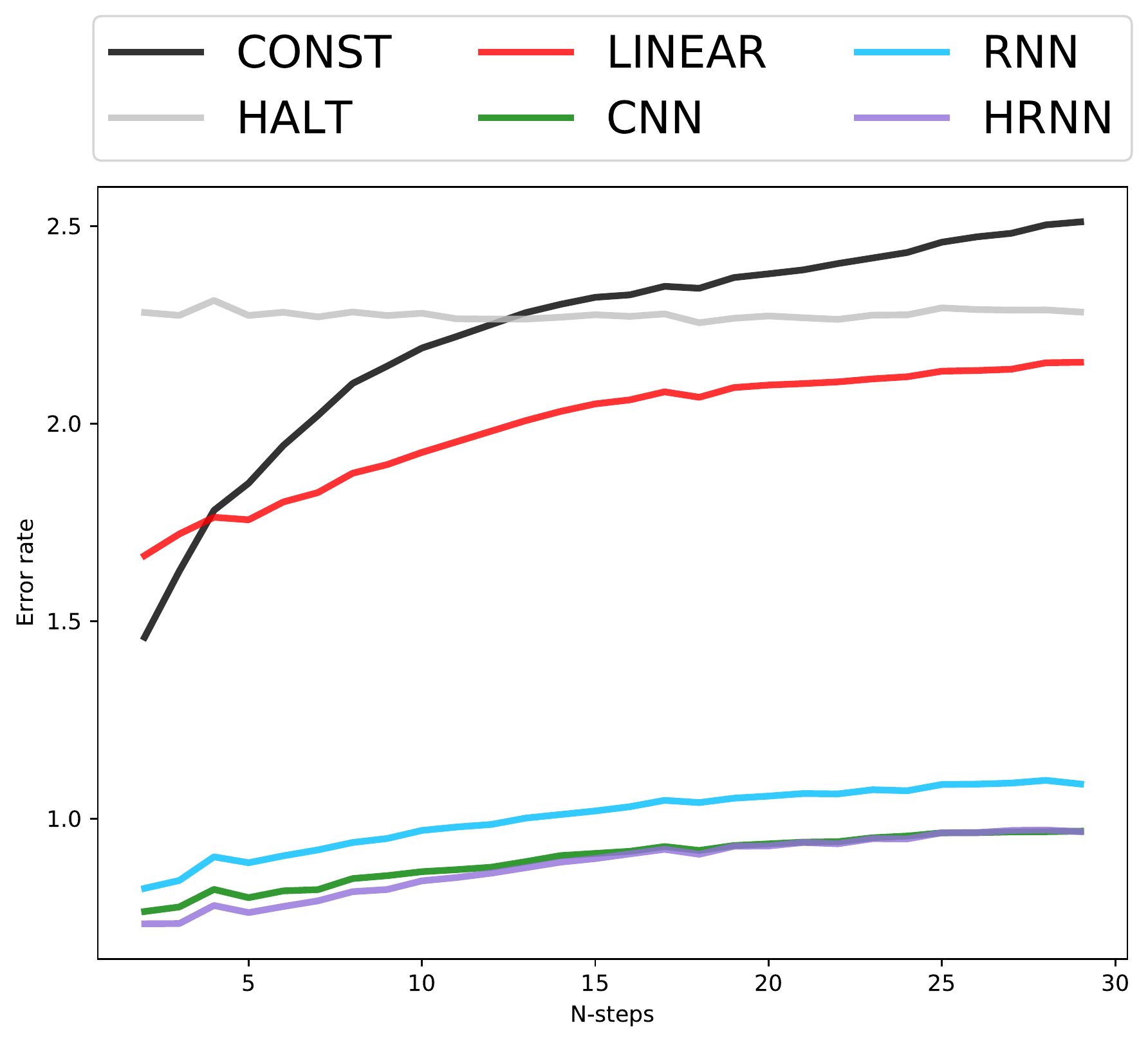}
            \vspace{-0.5cm}
            \subcaption*{Velocity, Male}
        \end{minipage}
        \begin{minipage}{0.24\textwidth}
            \includegraphics[width=\linewidth]{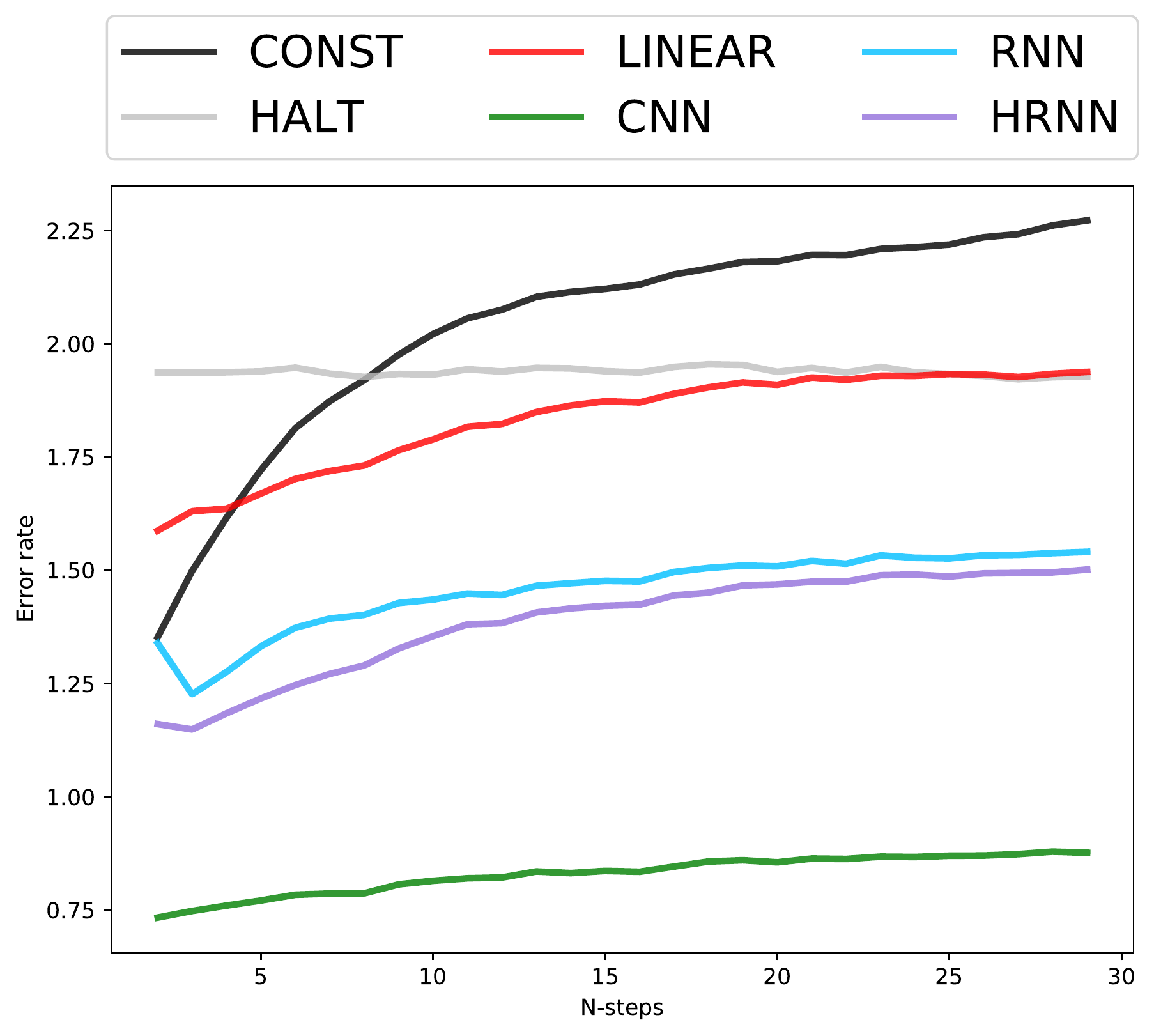}
            \vspace{-0.5cm}
            \subcaption*{Velocity, Female}
        \end{minipage}
        \subcaption{CONTROL}
    \end{minipage}
    \caption{$n$-step prediction position and velocity error rates for $n=1,\cdots,30$}
    \label{fig:nstep_supp}
\end{figure}

\begin{figure}[htp]
    \centering
    \begin{minipage}{0.49\textwidth}
        \centering
        \begin{minipage}{0.49\textwidth}
            \includegraphics[width=\linewidth]{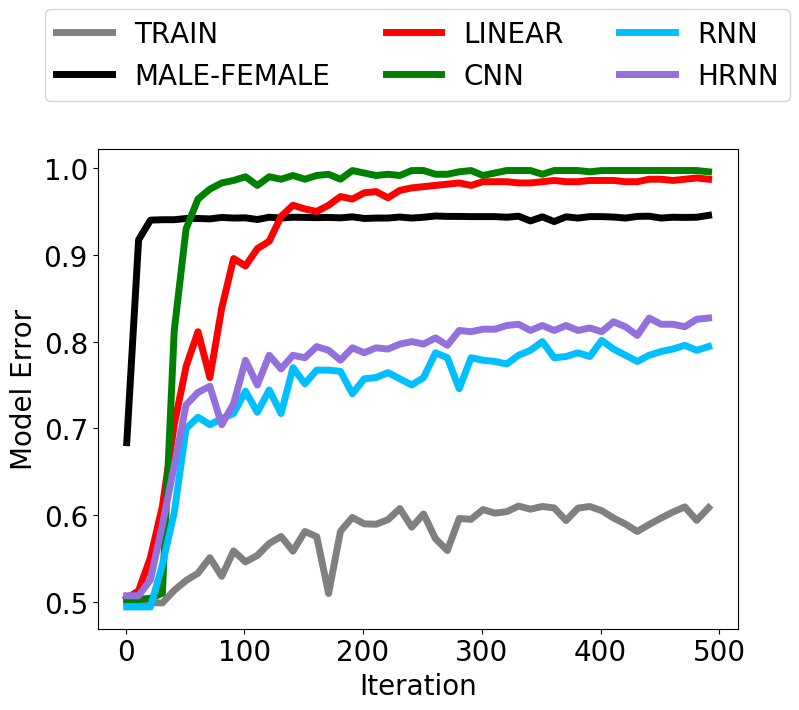}
            \vspace{-0.5cm}
            \subcaption*{Jensen-Shannon}
        \end{minipage}
        \begin{minipage}{0.49\textwidth}
            \includegraphics[width=\linewidth]{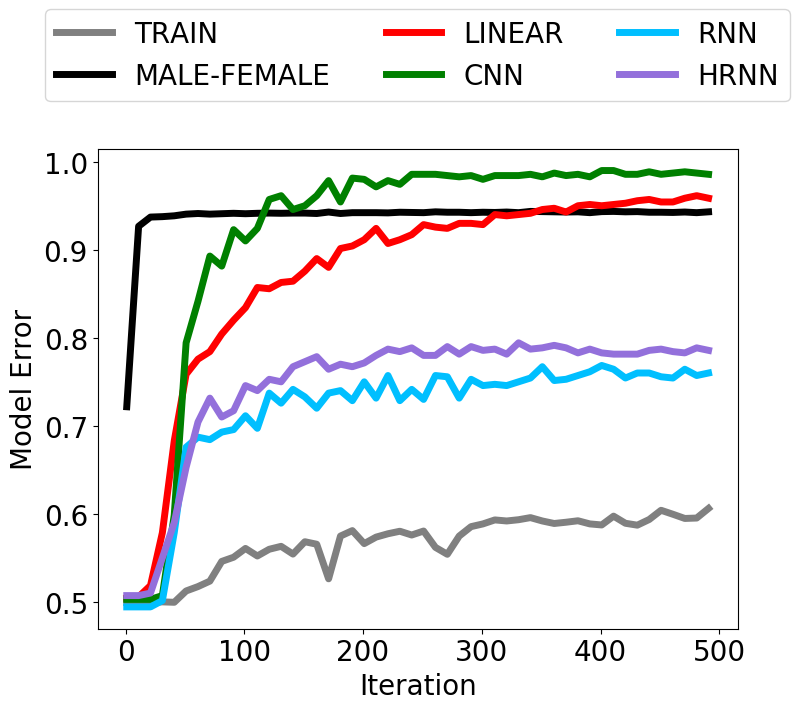}
            \vspace{-0.5cm}
            \subcaption*{Least-Squares}
        \end{minipage}
        \vspace{-0.2cm}
        \subcaption{using biologist's hand engineered features }
        \label{fig:train_fd_eval_supp}
    \end{minipage}
    \begin{minipage}{0.49\textwidth}
        \centering
        \begin{minipage}{0.49\textwidth}
            \includegraphics[width=\linewidth]{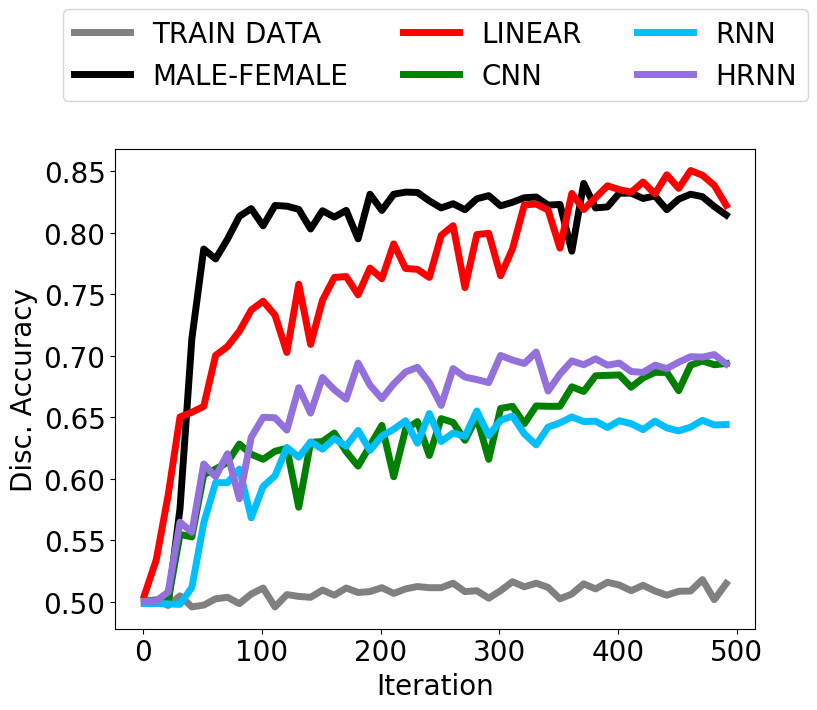}
            \vspace{-0.5cm}
            \subcaption*{Jensen-Shannon}
        \end{minipage}
        \begin{minipage}{0.49\textwidth}
            \includegraphics[width=\linewidth]{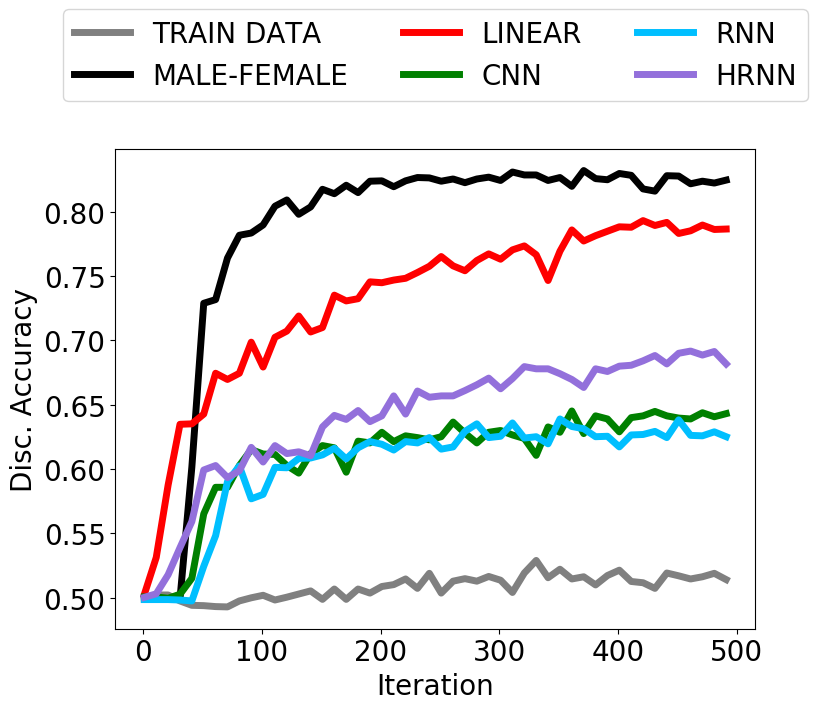}
            \vspace{-0.5cm}
            \subcaption*{Least-Squares}
        \end{minipage}
        \vspace{-0.2cm}
        \subcaption{using raw trajectory and vision-chamber data }
        \label{fig:train_gan_eval_supp}
    \end{minipage}
    \caption{Discriminator Training Curve: Validation accuracy rate over iteration on R71G01 male flies
           for Jensen-Shannon divergence and least-square distance}
    \label{fig:train_disc_supp}
\end{figure}

\begin{figure}[htp]
    \centering
    \begin{minipage}{0.49\textwidth}
        \centering
        \begin{minipage}{0.49\textwidth}
            \includegraphics[width=\linewidth]{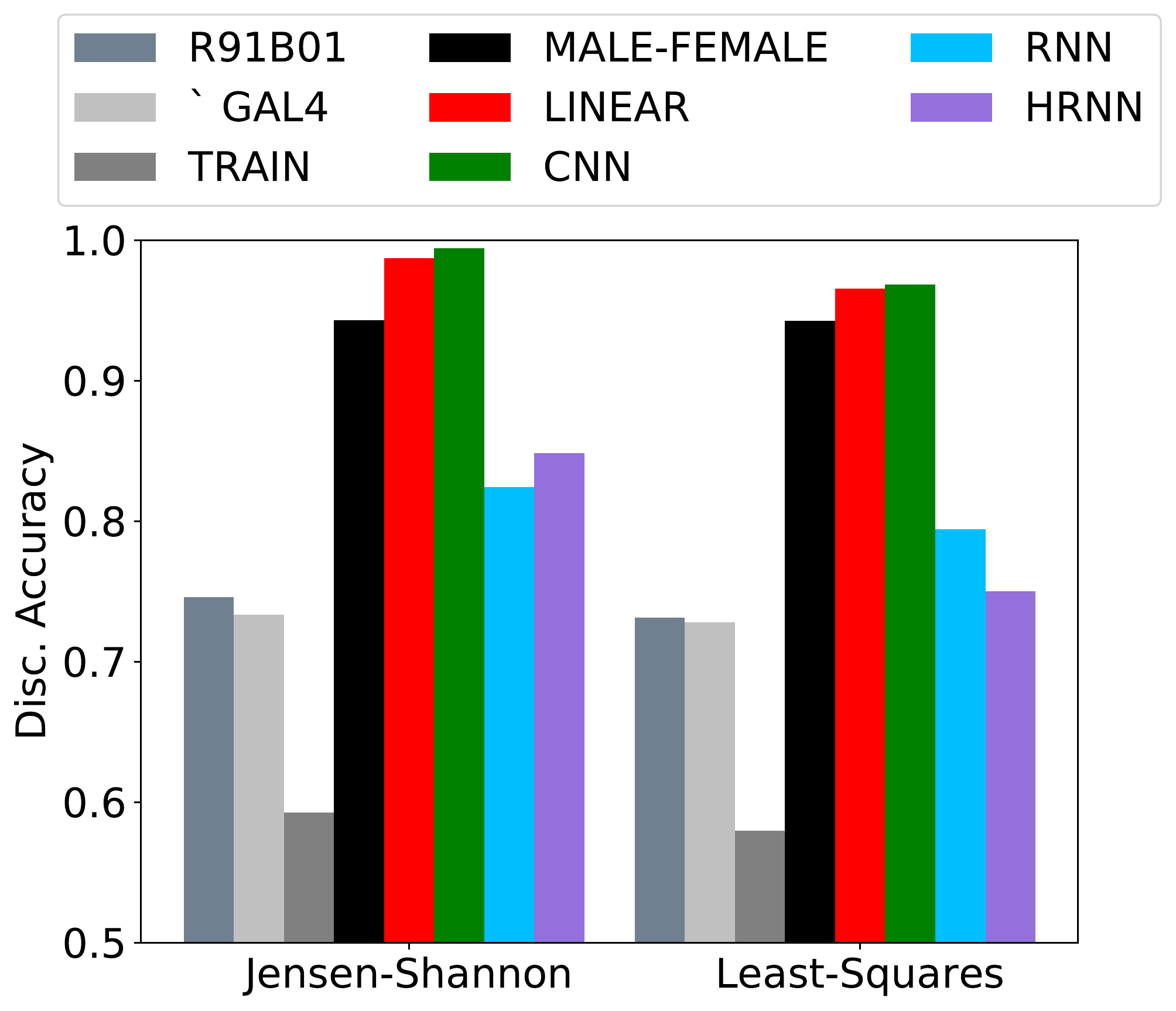}
            \vspace{-0.5cm}
            \subcaption*{Male}
        \end{minipage}
        \begin{minipage}{0.49\textwidth}
            \includegraphics[width=\linewidth]{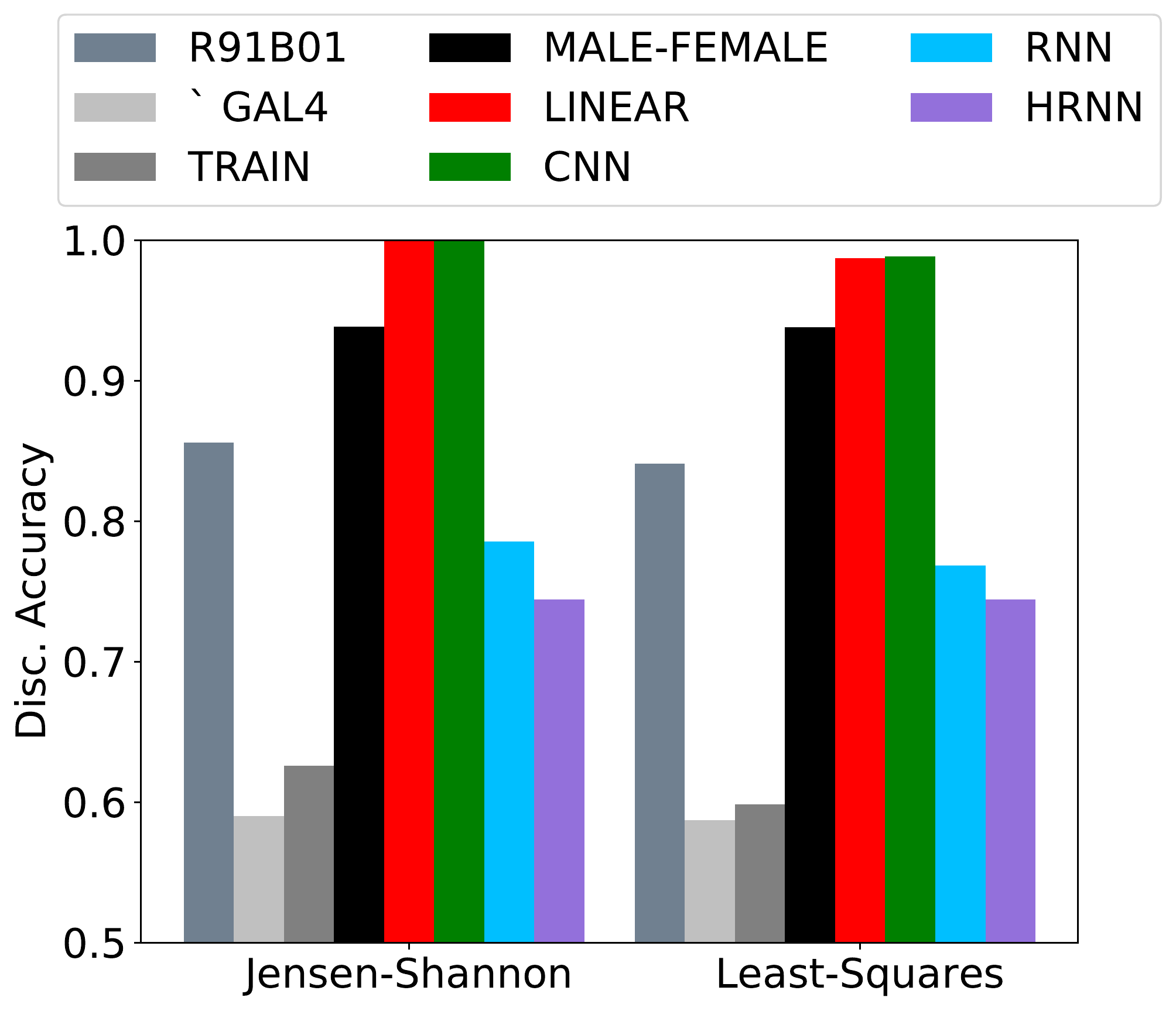}
            \vspace{-0.5cm}
            \subcaption*{Female}
        \end{minipage}
        \vspace{-0.2cm}
        \subcaption{using biologist's hand engineered features }
        \label{fig:fd_eval_supp}
    \end{minipage}
    \begin{minipage}{0.49\textwidth}
        \centering
        \begin{minipage}{0.49\textwidth}
            \includegraphics[width=\linewidth]{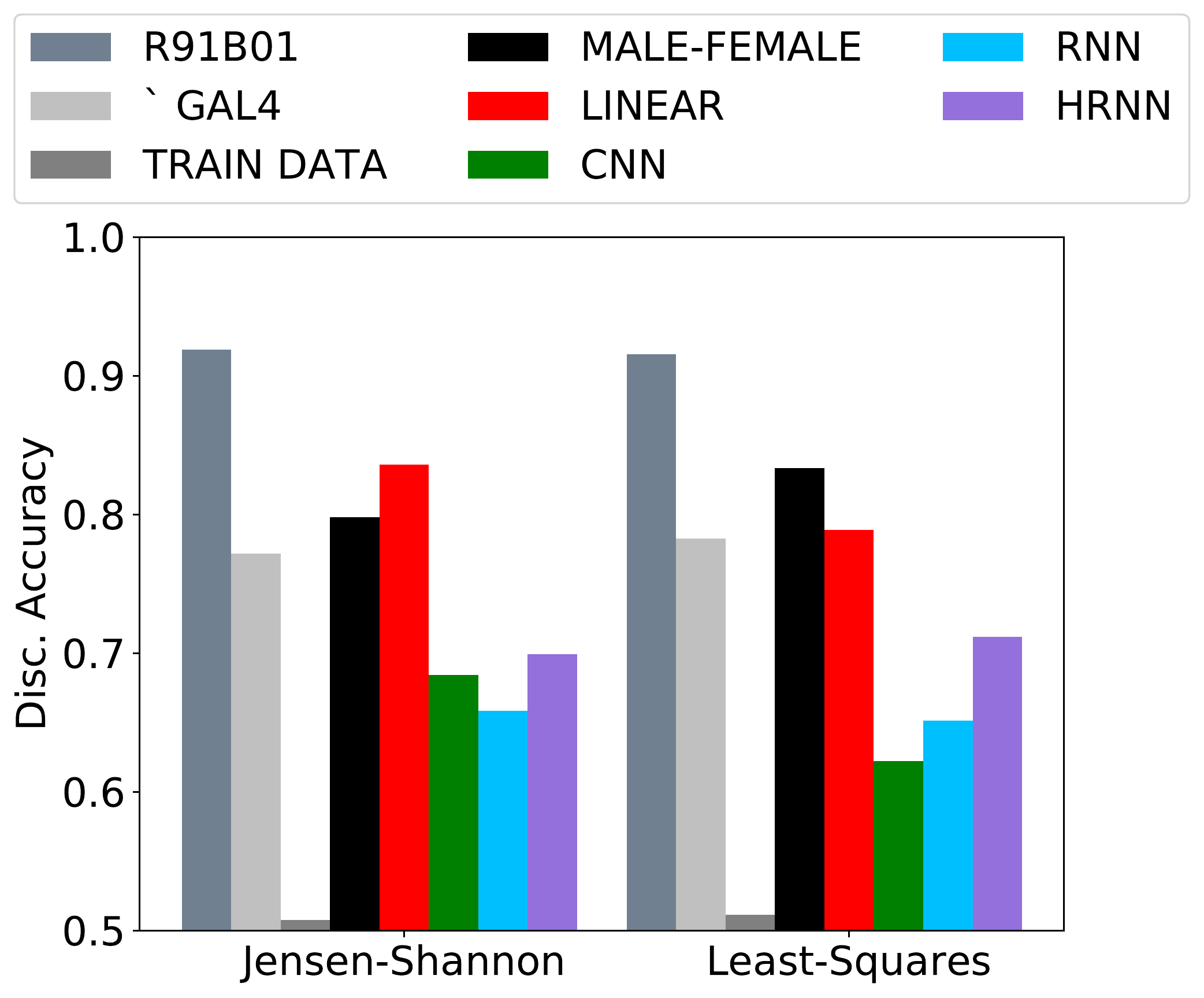}
            \vspace{-0.5cm}
            \subcaption*{Male}
        \end{minipage}
        \begin{minipage}{0.49\textwidth}
            \includegraphics[width=\linewidth]{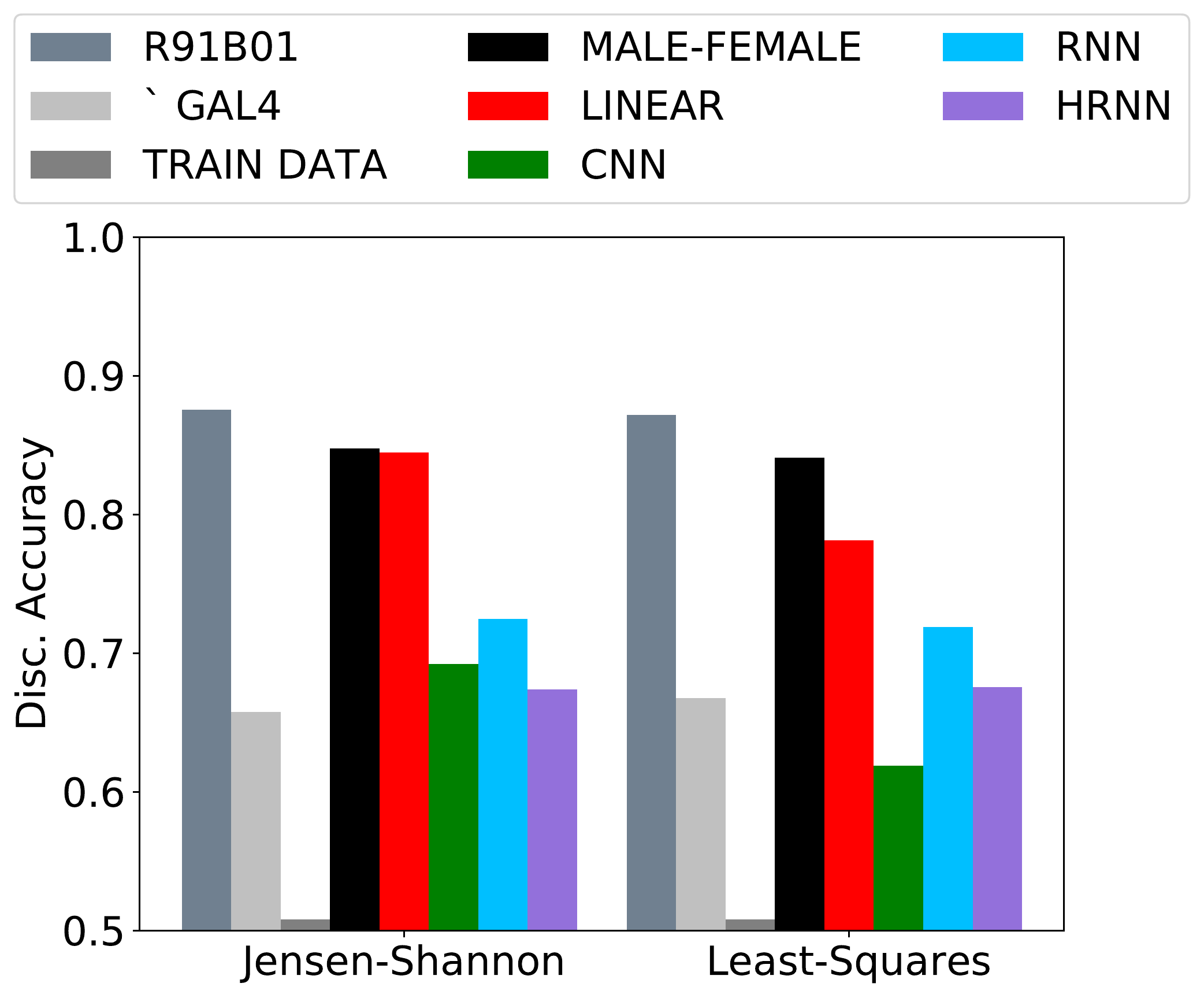}
            \vspace{-0.5cm}
            \subcaption*{Female}
        \end{minipage}
        \vspace{-0.2cm}
        \subcaption{using raw trajectory and vision-chamber data }
        \label{fig:gan_eval_supp}
    \end{minipage}
    \centering
    \begin{minipage}{0.49\textwidth}
        \centering
        \begin{minipage}{0.49\textwidth}
            \includegraphics[width=\linewidth]{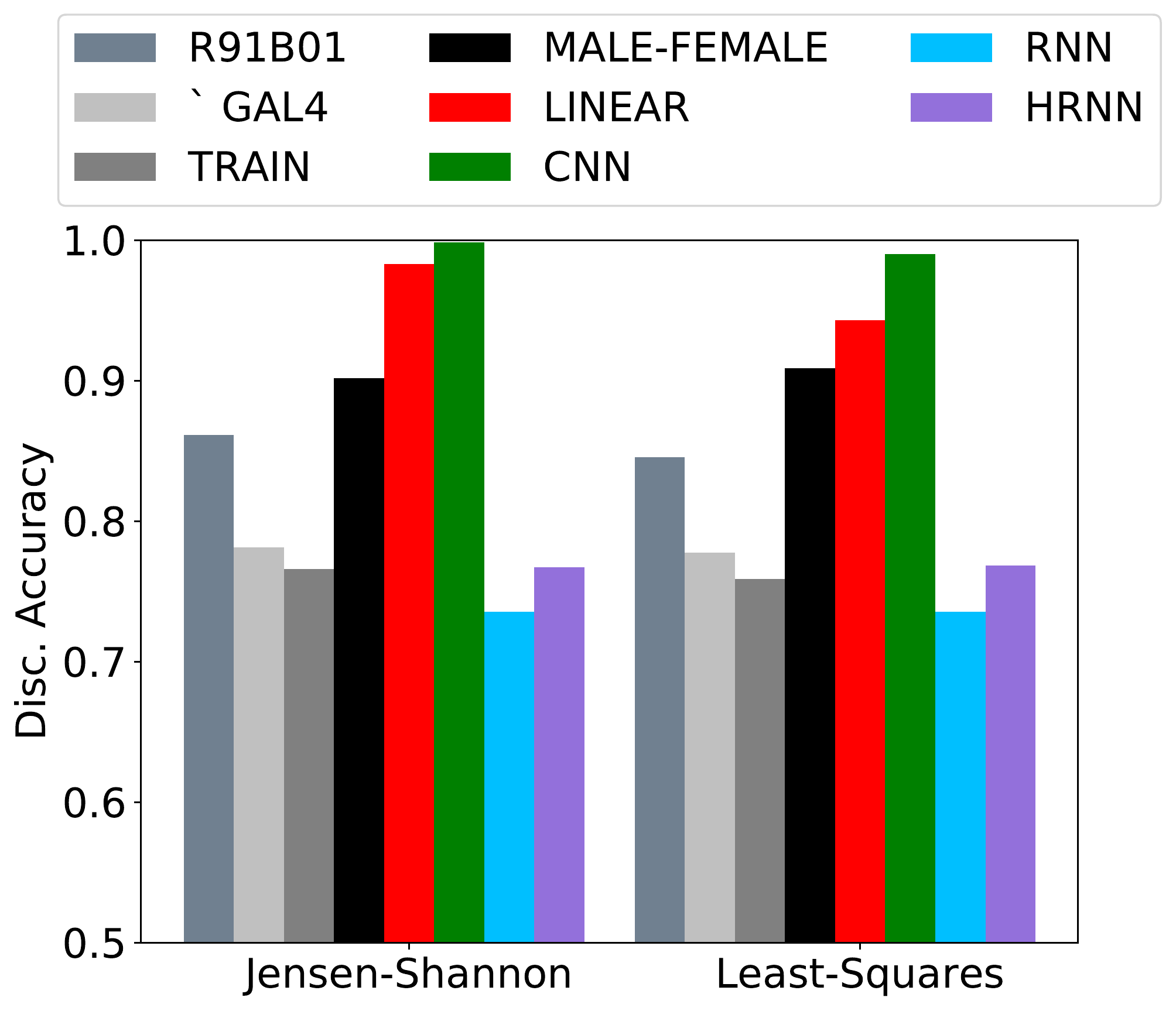}
            \vspace{-0.5cm}
            \subcaption*{Male}
        \end{minipage}
        \begin{minipage}{0.49\textwidth}
            \includegraphics[width=\linewidth]{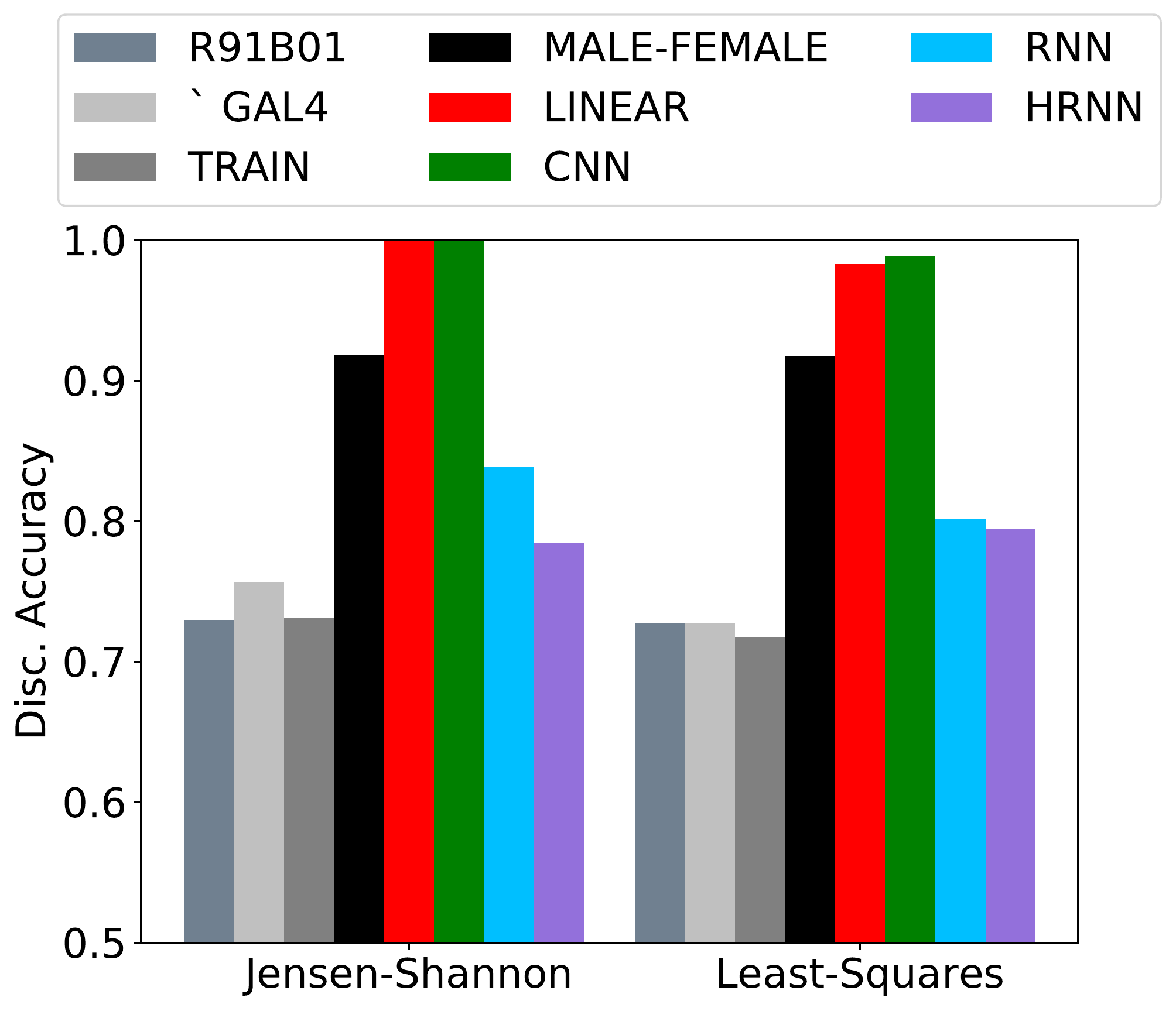}
            \vspace{-0.5cm}
            \subcaption*{Female}
        \end{minipage}
        \subcaption{using biologist's hand engineered features }
        \label{fig:fd_eval_supp1}
    \end{minipage}
    \begin{minipage}{0.49\textwidth}
        \centering
        \begin{minipage}{0.49\textwidth}
            \includegraphics[width=\linewidth]{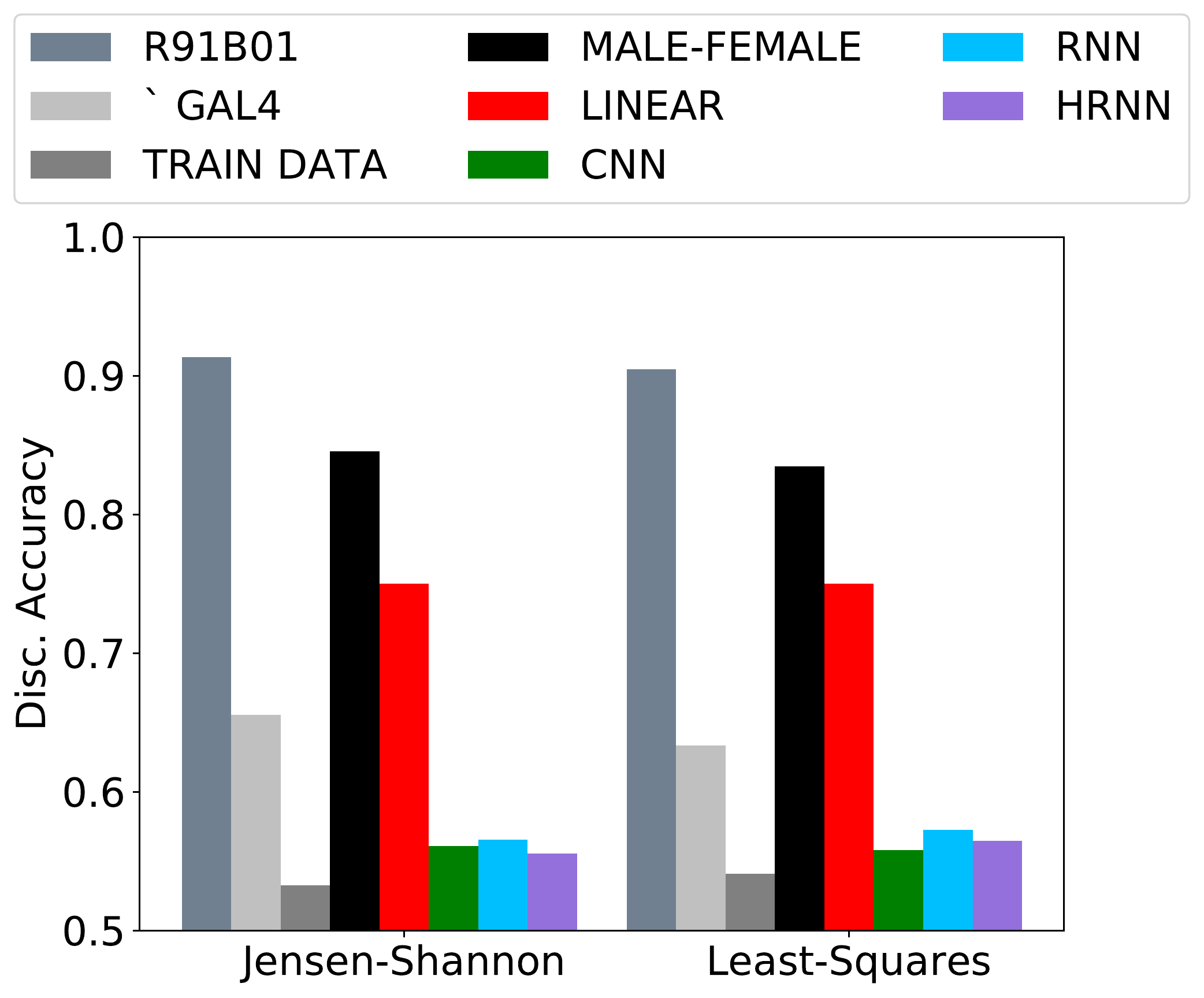}
            \vspace{-0.5cm}
            \subcaption*{Male}
        \end{minipage}
        \begin{minipage}{0.49\textwidth}
            \includegraphics[width=\linewidth]{./figs/ganeval_errorbar__gmr91_male.pdf}
            \vspace{-0.5cm}
            \subcaption*{Female}
        \end{minipage}
        \subcaption{using raw trajectory and vision-chamber data }
        \label{fig:gan_eval_supp1}
    \end{minipage}
    \centering
    \begin{minipage}{0.49\textwidth}
        \centering
        \begin{minipage}{0.49\textwidth}
            \includegraphics[width=\linewidth]{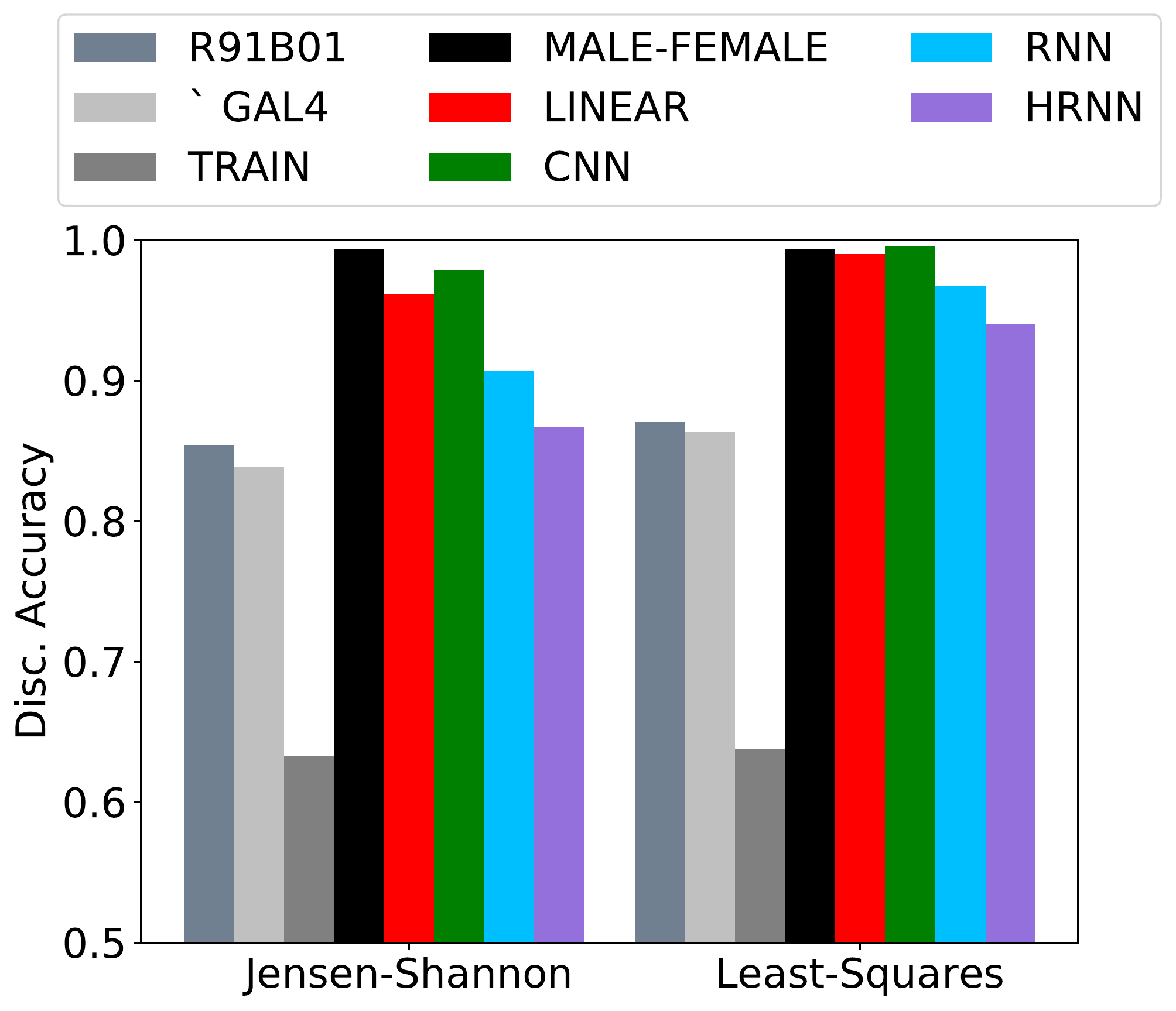}
            \vspace{-0.5cm}
            \subcaption*{Male}
        \end{minipage}
        \begin{minipage}{0.49\textwidth}
            \includegraphics[width=\linewidth]{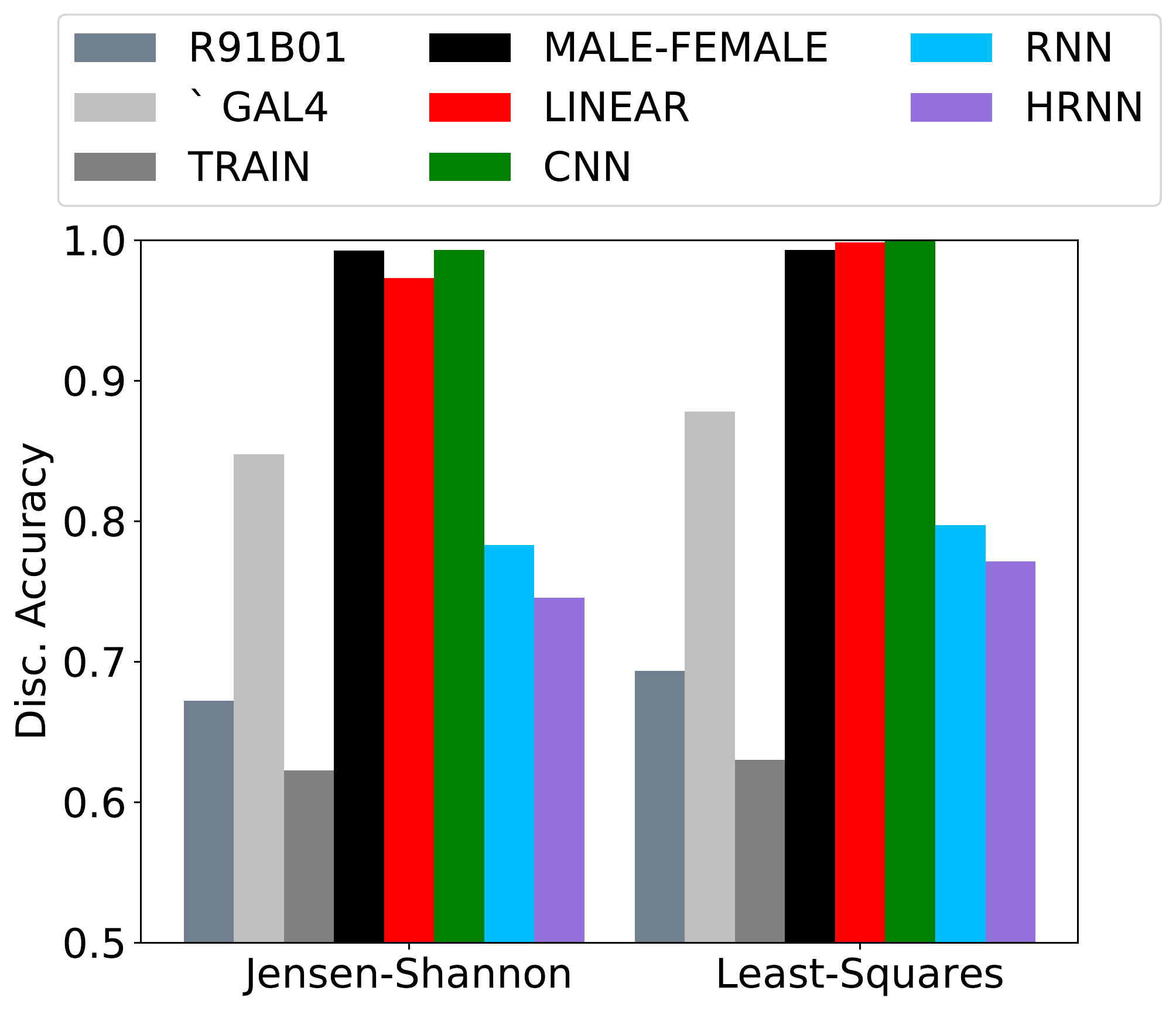}
            \vspace{-0.5cm}
            \subcaption*{Female}
        \end{minipage}
        \subcaption{using biologist's hand engineered features }
        \label{fig:fd_eval_supp2}
    \end{minipage}
    \begin{minipage}{0.49\textwidth}
        \centering
        \begin{minipage}{0.49\textwidth}
            \includegraphics[width=\linewidth]{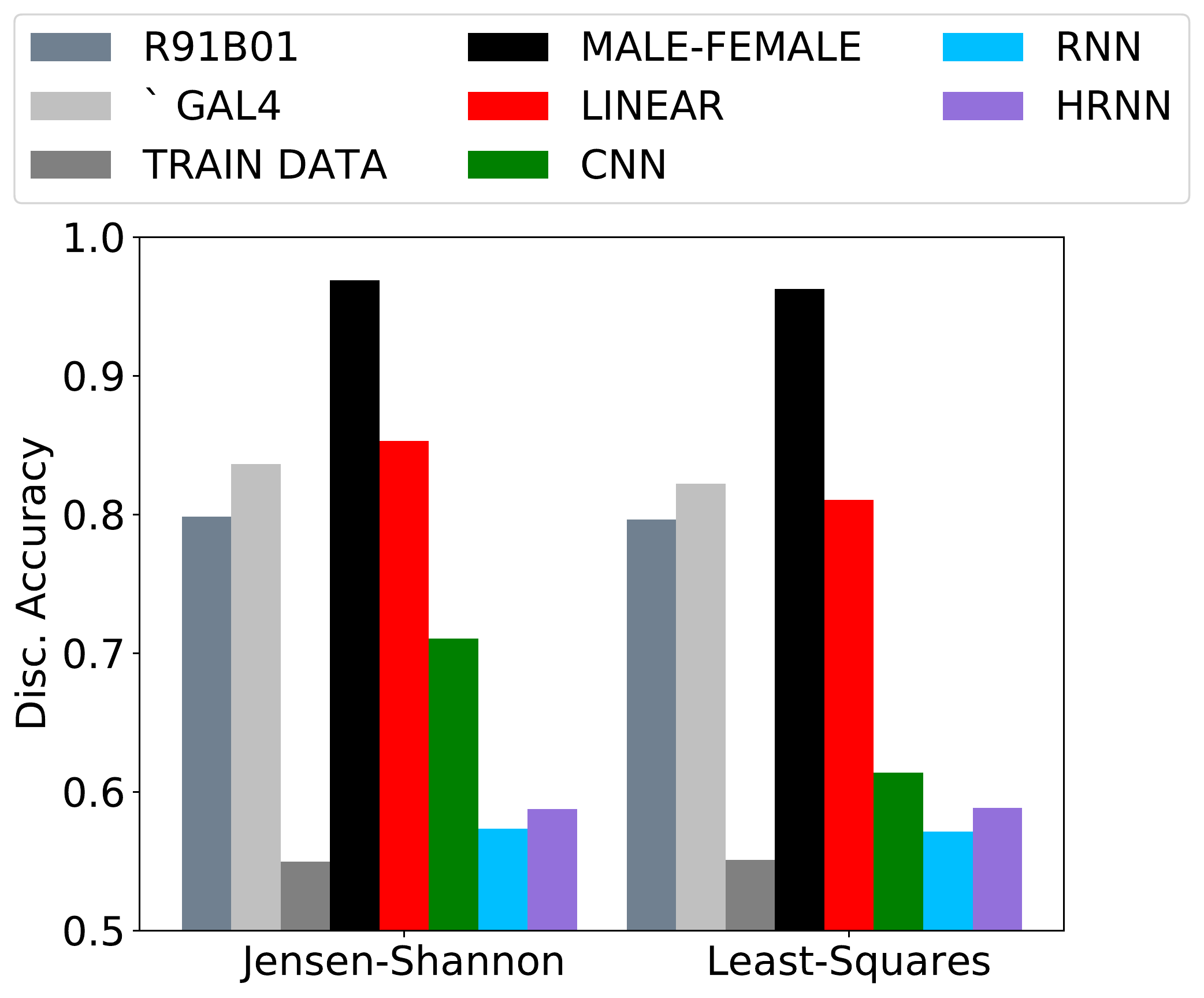}
            \vspace{-0.5cm}
            \subcaption*{Male}
        \end{minipage}
        \begin{minipage}{0.49\textwidth}
            \includegraphics[width=\linewidth]{./figs/ganeval_errorbar__pdb_male.pdf}
            \vspace{-0.5cm}
            \subcaption*{Female}
        \end{minipage}
        \subcaption{using raw trajectory and vision-chamber data }
        \label{fig:gan_eval_supp2}
    \end{minipage}
    \caption{Discriminator Evaluation Accuracy Rate on R71G01, R91B01 and CONTROL.
           The discriminators were trained using Jensen-Shannon divergence and least-square distance}
\end{figure}

\null\newpage
\subsection{S.M. for Long Term Performance Evaluation}
\begin{figure}[htp]
    \centering
    \begin{minipage}{0.99\textwidth}
    \begin{minipage}{0.19\textwidth}
        \includegraphics[width=\linewidth]{./figs/test_hist_velocity_gmr71_male_allmodel.pdf}
        \vspace{-0.5cm}
        \subcaption*{Velocity}
    \end{minipage}
    \begin{minipage}{0.19\textwidth}
        \includegraphics[width=\linewidth]{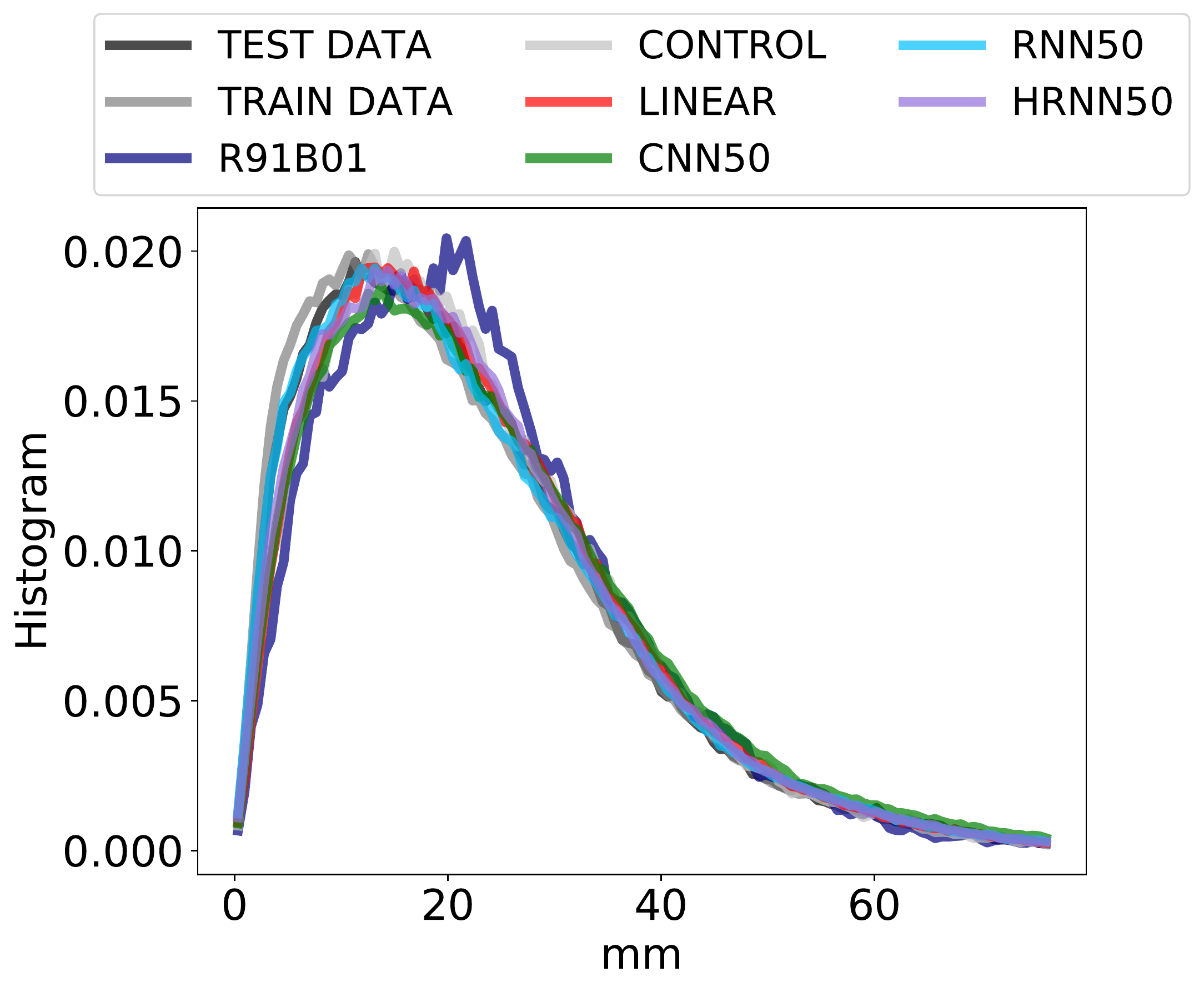}
        \vspace{-0.5cm}
        \subcaption*{Inter Distance}
    \end{minipage}
    \begin{minipage}{0.19\textwidth}
        \includegraphics[width=\linewidth]{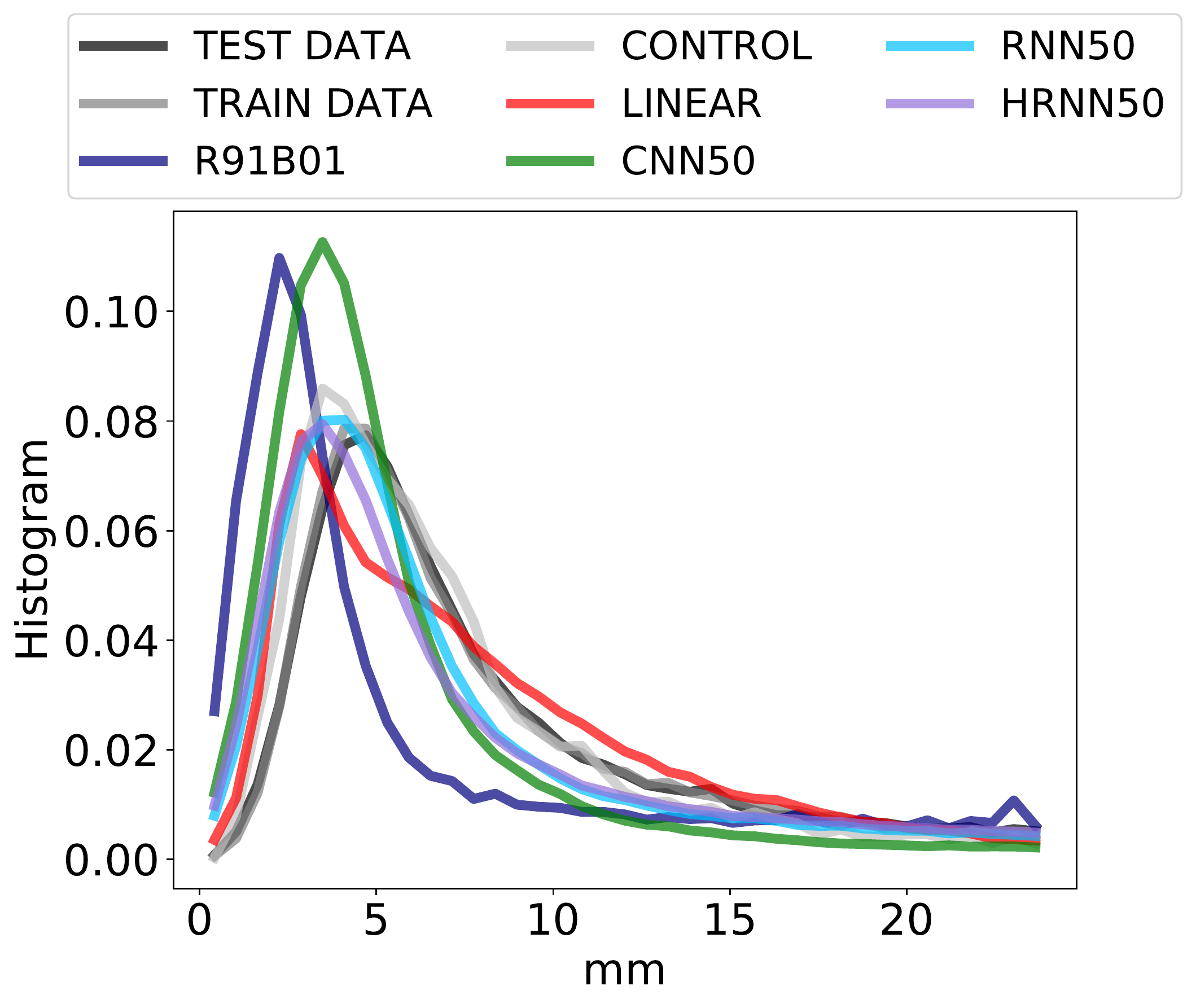}
        \vspace{-0.5cm}
        \subcaption*{Wall Distance}
    \end{minipage}
    \begin{minipage}{0.19\textwidth}
        \includegraphics[width=\linewidth]{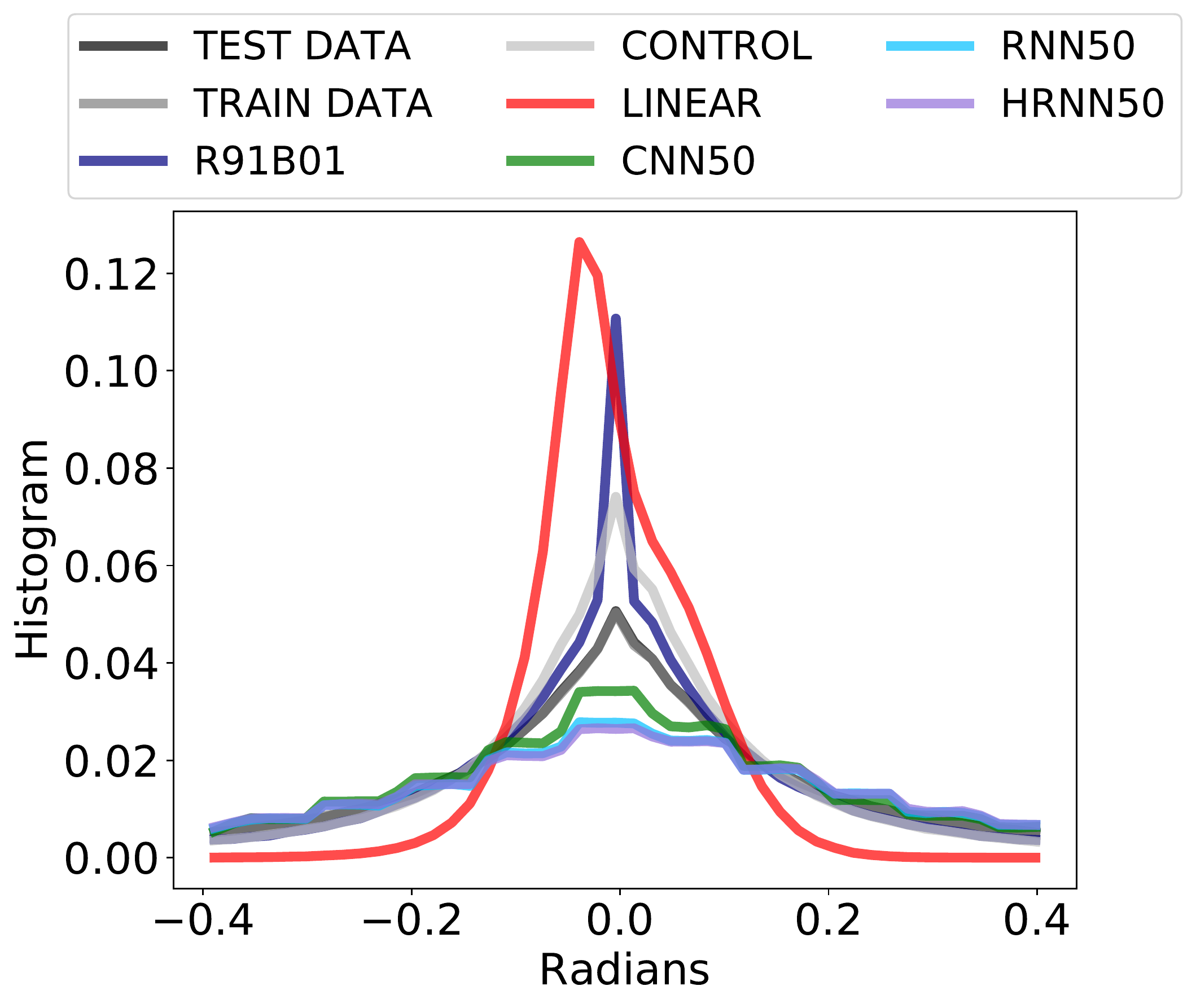}
        \vspace{-0.5cm}
        \subcaption*{Angular Motion}
    \end{minipage}
    \begin{minipage}{0.19\textwidth}
        \includegraphics[width=\linewidth]{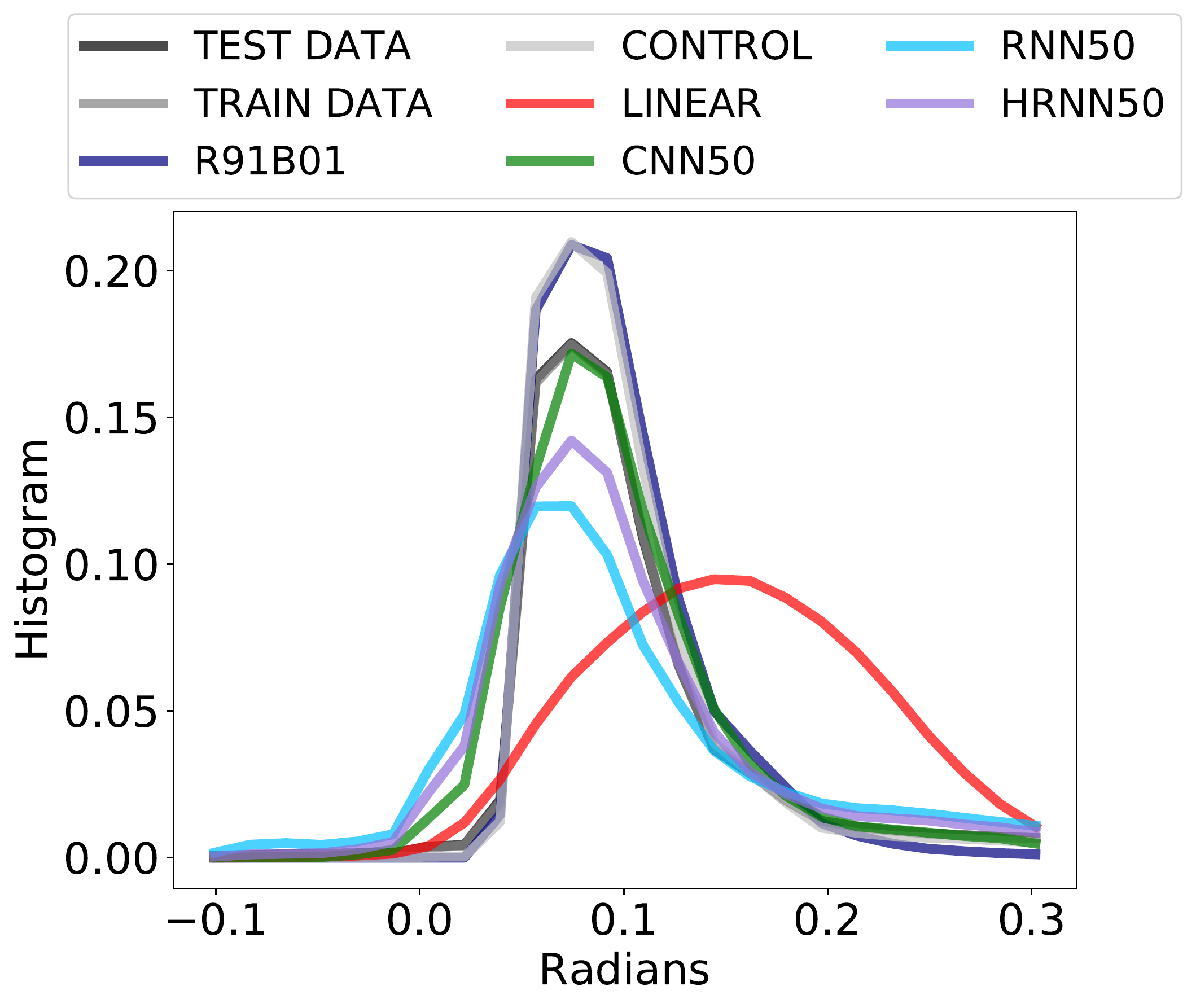}
        \vspace{-0.5cm}
        \subcaption*{Wing Angle}
    \end{minipage}\\
    \vspace{-0.2cm}
    \subcaption*{Male}
    \end{minipage}
    \begin{minipage}{0.99\textwidth}
    \begin{minipage}{0.19\textwidth}
        \includegraphics[width=\linewidth]{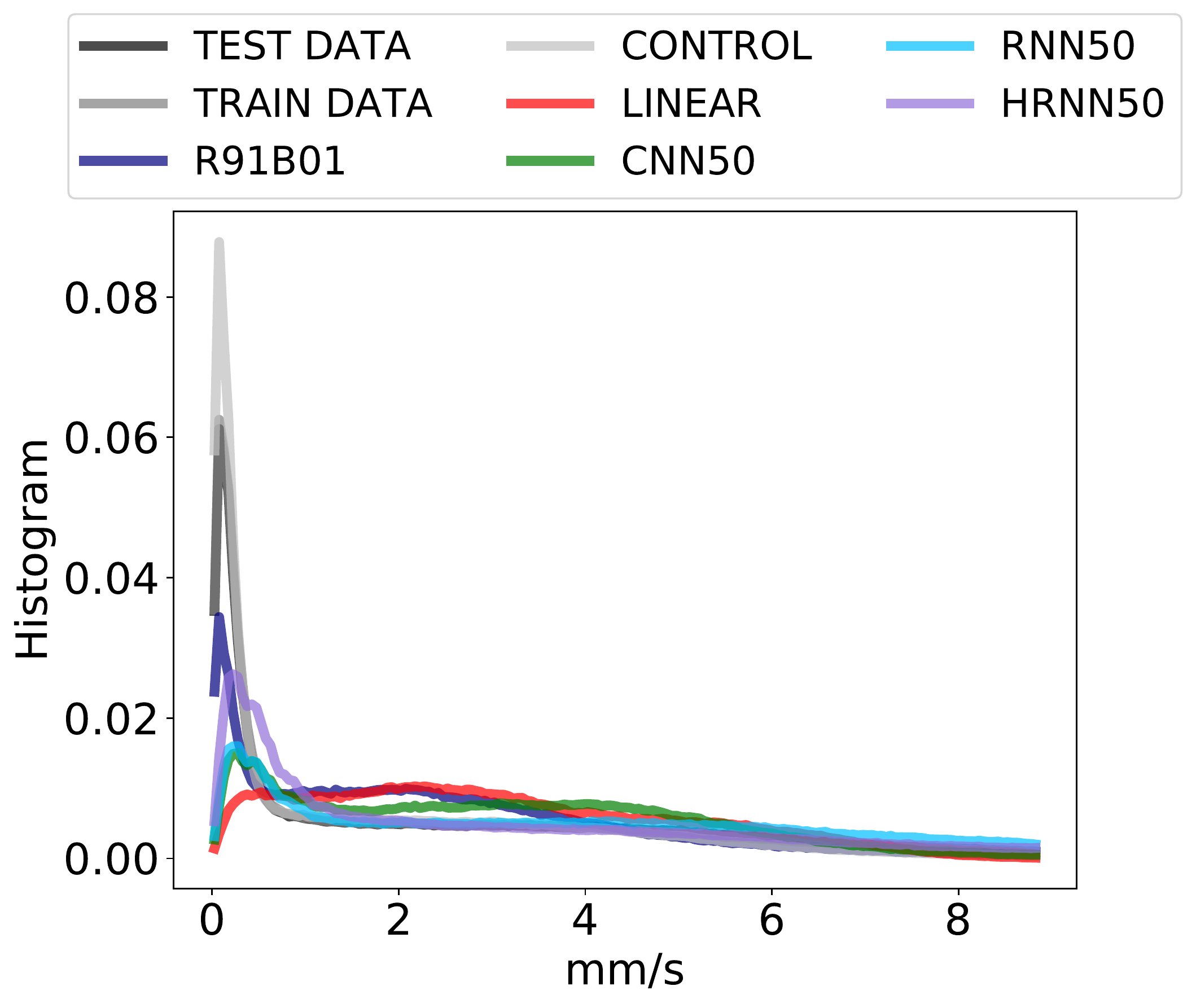}
        \vspace{-0.5cm}
        \subcaption*{Velocity}
    \end{minipage}
    \begin{minipage}{0.19\textwidth}
        \includegraphics[width=\linewidth]{./figs/test_hist_inter_dist_gmr71_fale_allmodel.pdf}
        \vspace{-0.5cm}
        \subcaption*{Inter Distance}
    \end{minipage}
    \begin{minipage}{0.19\textwidth}
        \includegraphics[width=\linewidth]{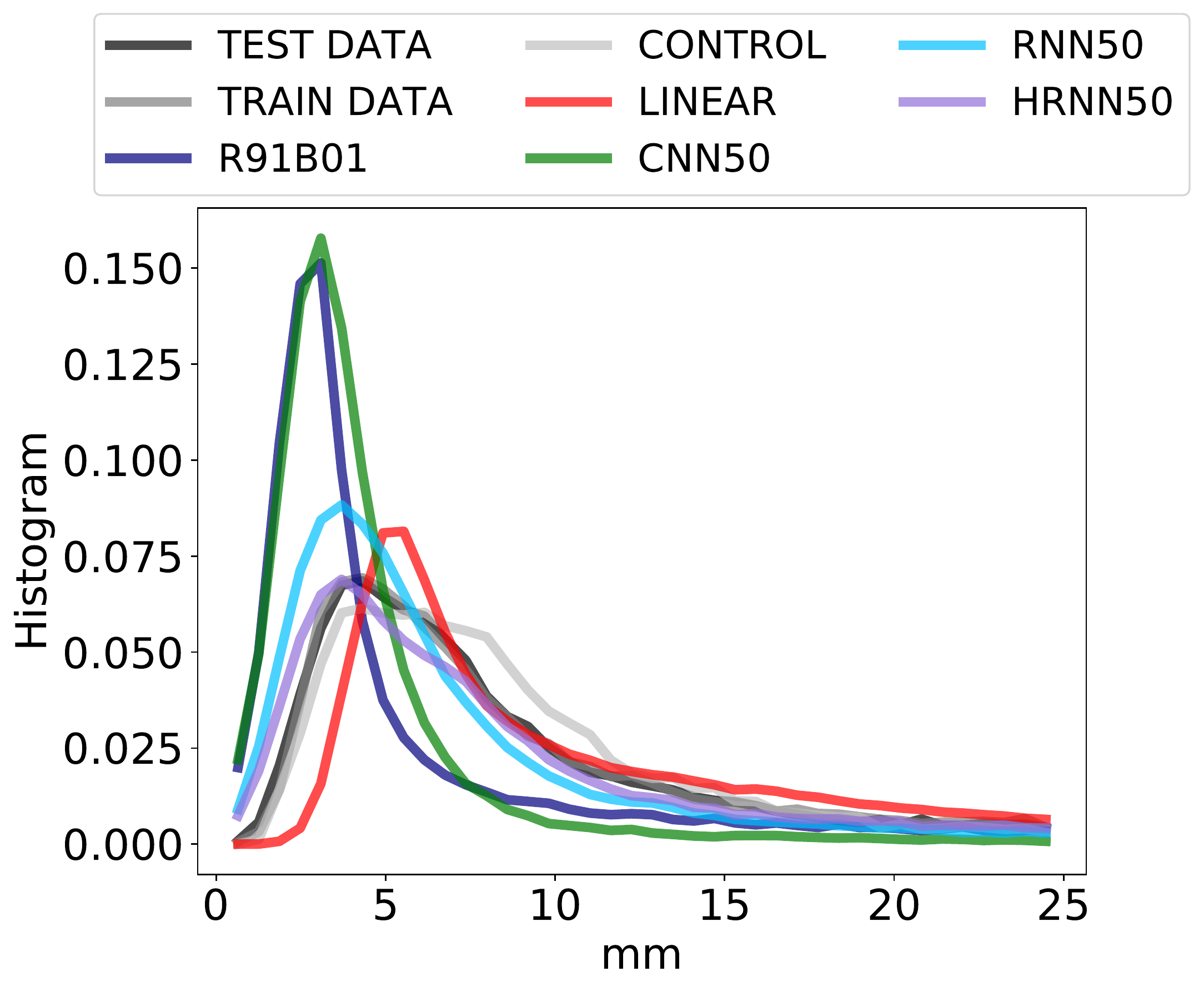}
        \vspace{-0.5cm}
        \subcaption*{Wall Distance}
    \end{minipage}
    \begin{minipage}{0.19\textwidth}
        \includegraphics[width=\linewidth]{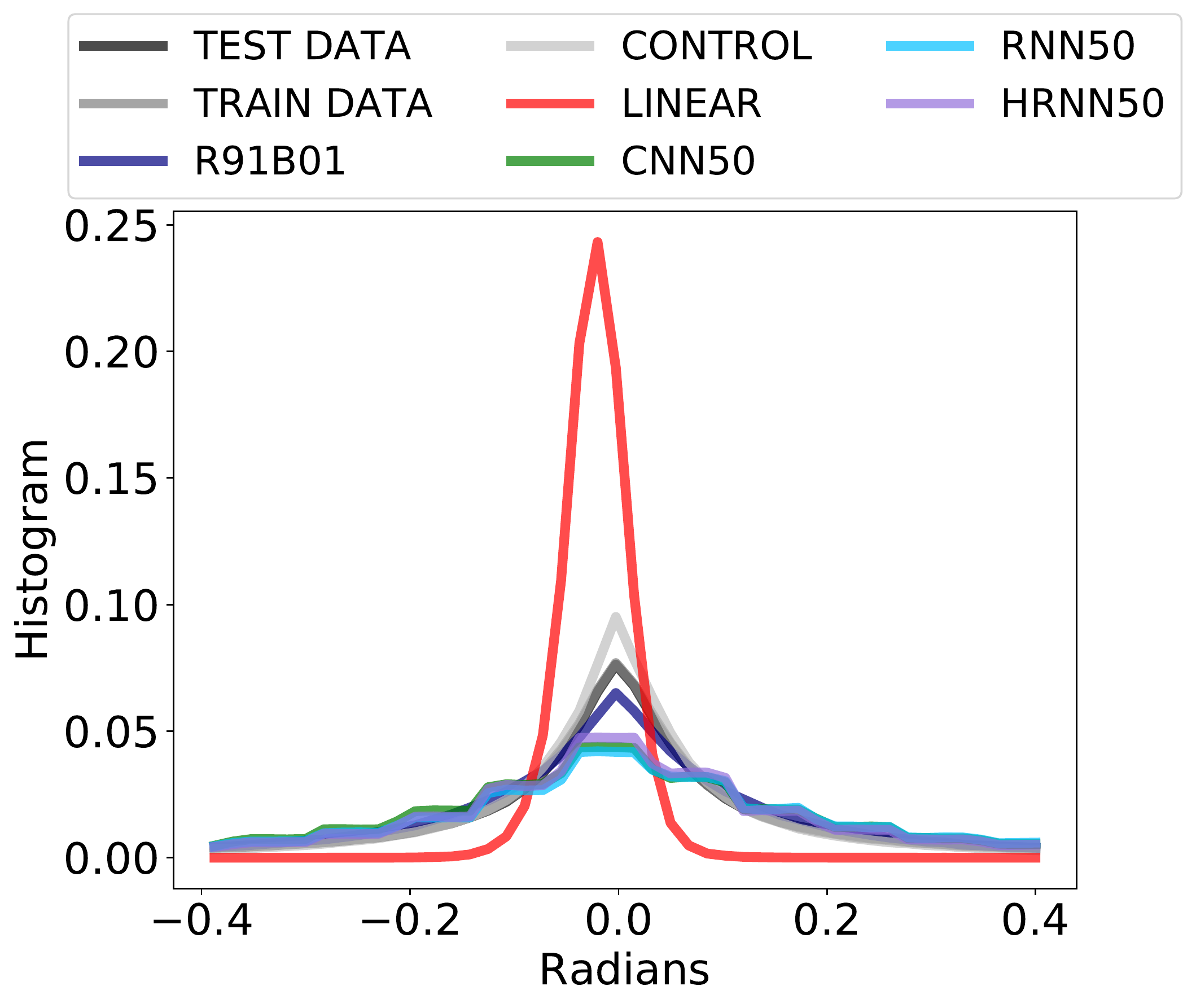}
        \vspace{-0.5cm}
        \subcaption*{Angular Motion}
    \end{minipage}
    \begin{minipage}{0.19\textwidth}
        \includegraphics[width=\linewidth]{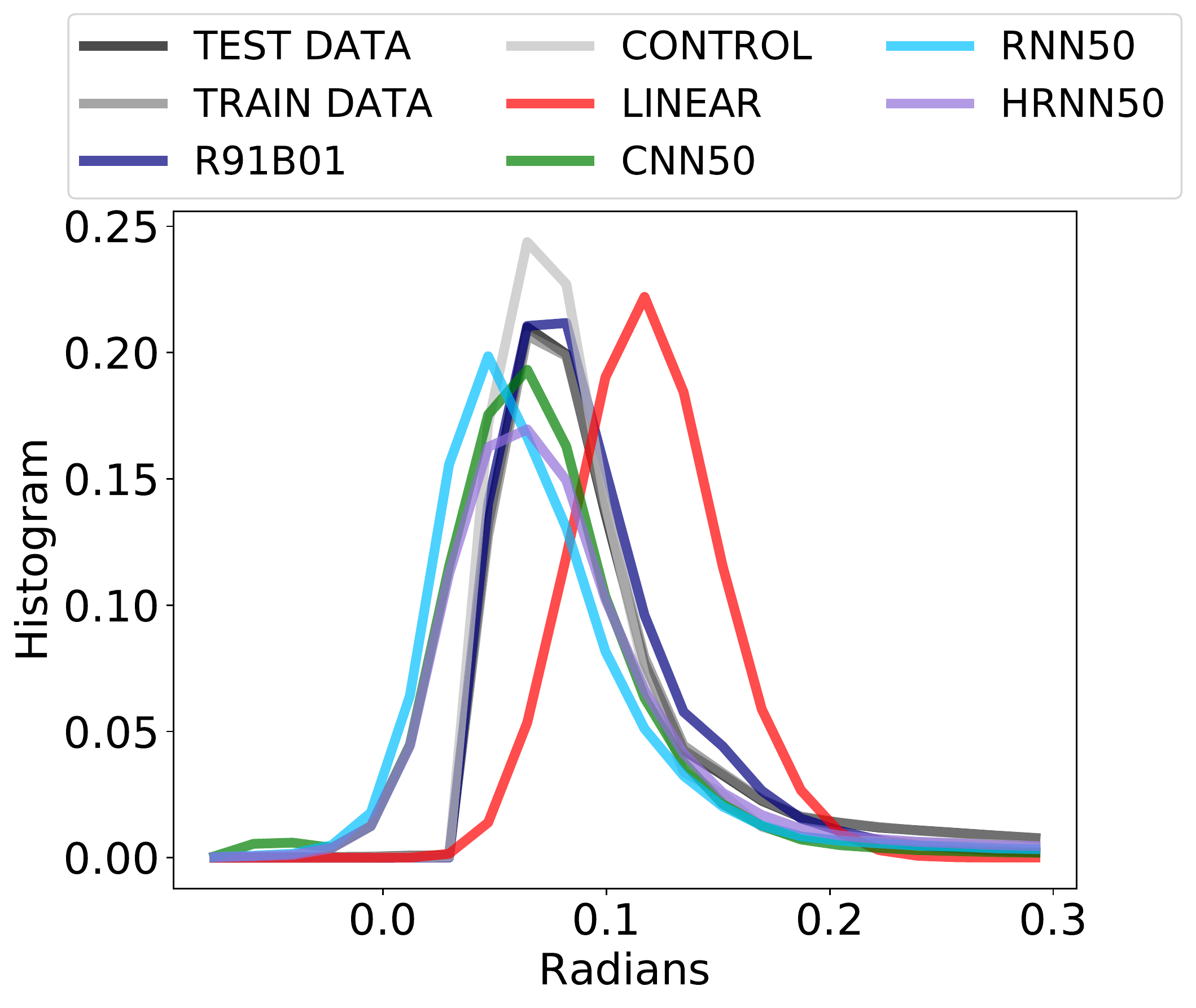}
        \vspace{-0.5cm}
        \subcaption*{Wing Angle}
    \end{minipage}
    \vspace{-0.2cm}
    \subcaption*{Female}
    \end{minipage}
    \vspace{-0.2cm}
    \caption{Histogram R71G01}
    \centering
    \begin{minipage}{0.99\textwidth}
    \begin{minipage}{0.19\textwidth}
        \includegraphics[width=\linewidth]{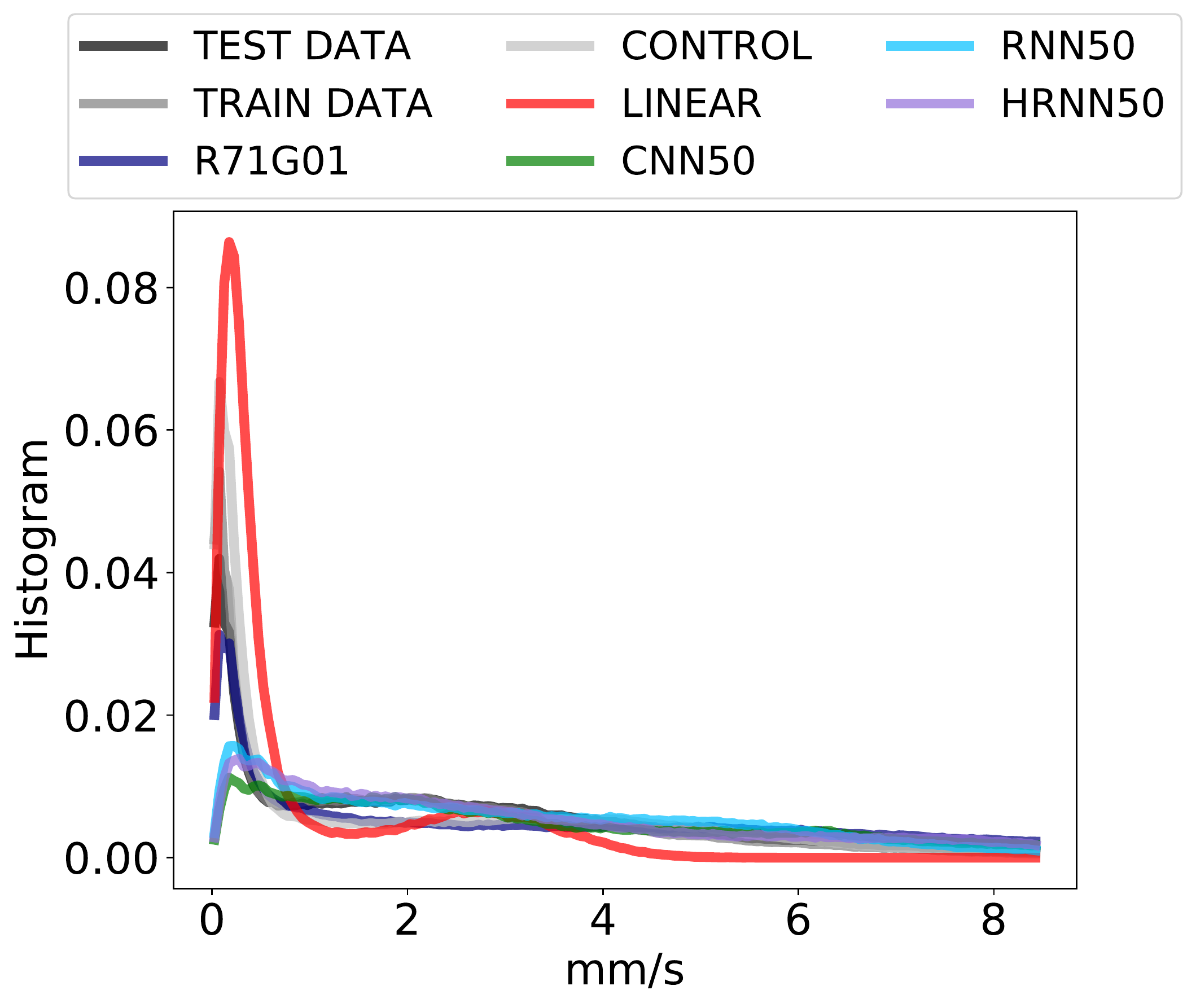}
        \vspace{-0.5cm}
        \subcaption*{Velocity}
    \end{minipage}
    \begin{minipage}{0.19\textwidth}
        \includegraphics[width=\linewidth]{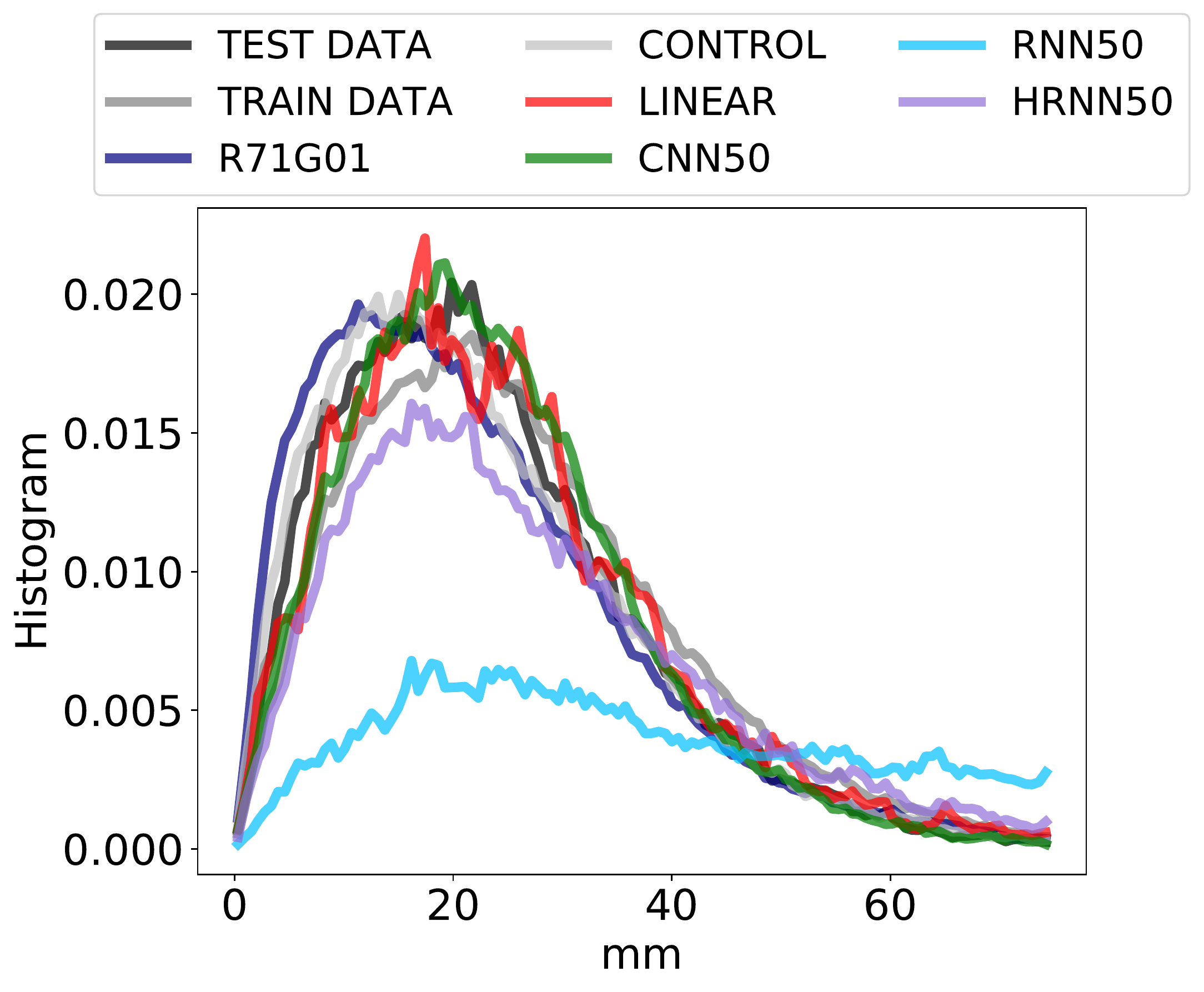}
        \vspace{-0.5cm}
        \subcaption*{Inter Distance}
    \end{minipage}
    \begin{minipage}{0.19\textwidth}
        \includegraphics[width=\linewidth]{./figs/test_hist_wall_distance_gmr91_male_allmodel.pdf}
        \vspace{-0.5cm}
        \subcaption*{Wall Distance}
    \end{minipage}
    \begin{minipage}{0.19\textwidth}
        \includegraphics[width=\linewidth]{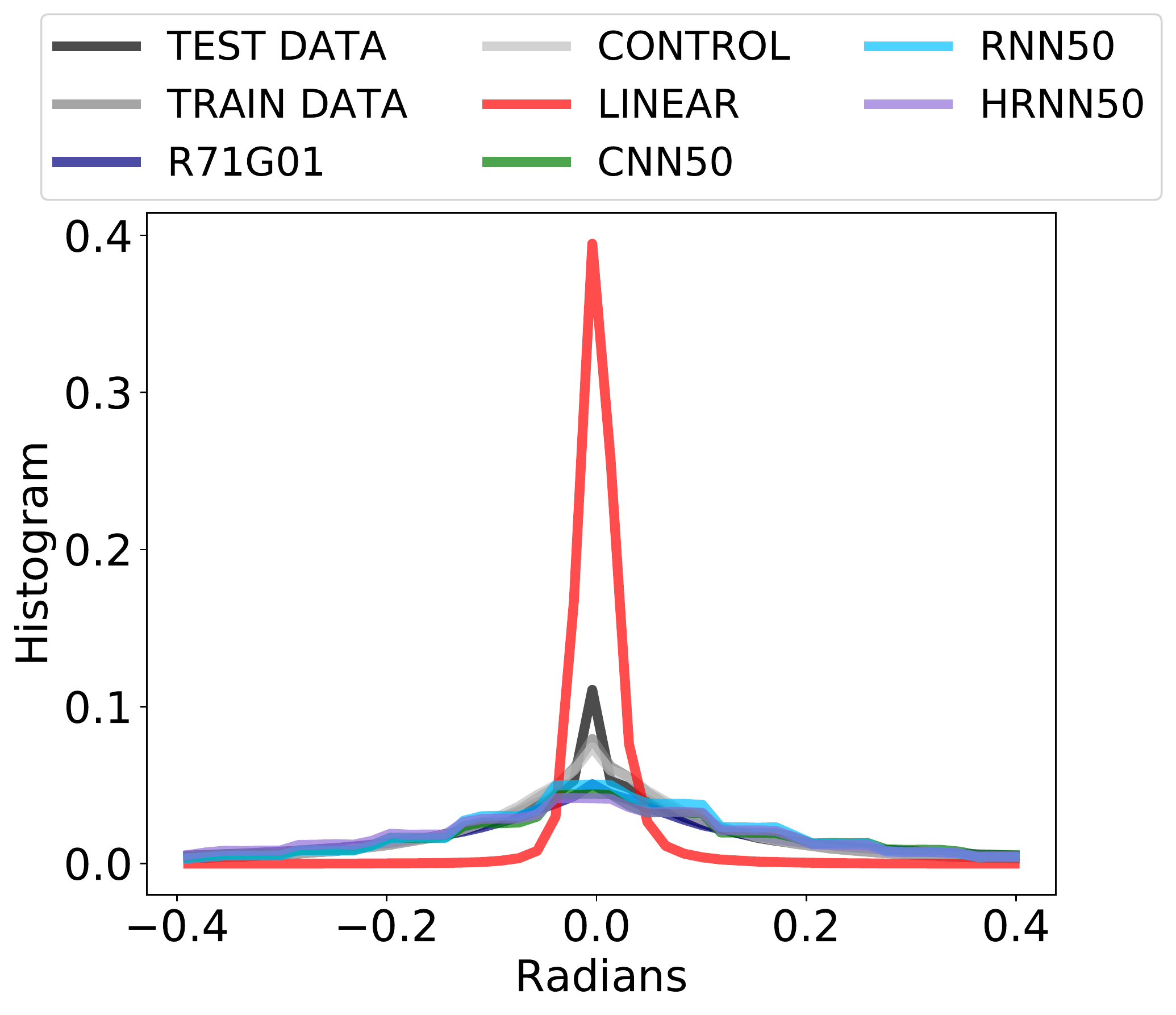}
        \vspace{-0.5cm}
        \subcaption*{Angular Motion}
    \end{minipage}
    \begin{minipage}{0.19\textwidth}
        \includegraphics[width=\linewidth]{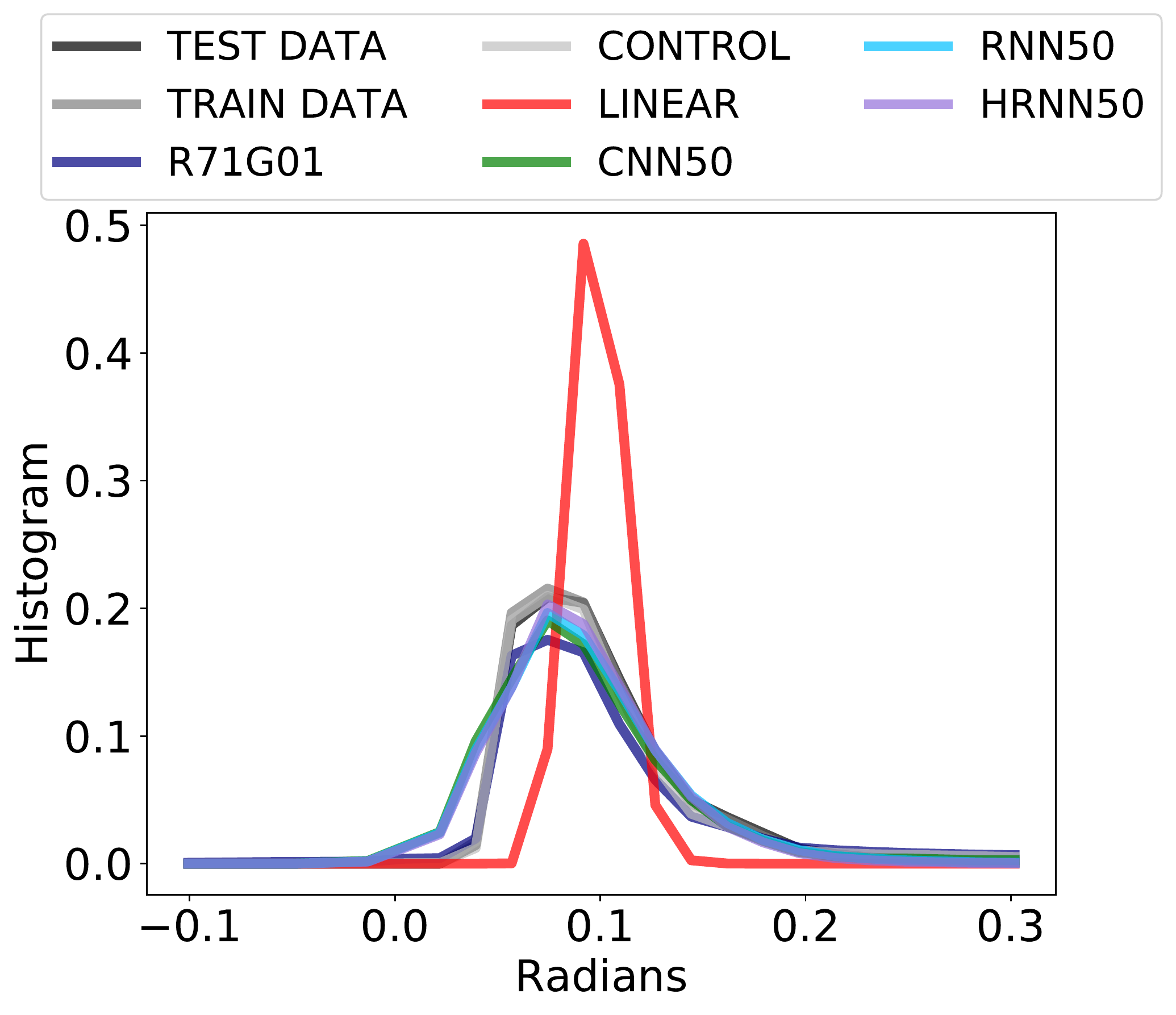}
        \vspace{-0.5cm}
        \subcaption*{Wing Angle}
    \end{minipage}\\
    \vspace{-0.2cm}
    \subcaption*{Male}
    \end{minipage}
    \begin{minipage}{0.99\textwidth}
    \begin{minipage}{0.19\textwidth}
        \includegraphics[width=\linewidth]{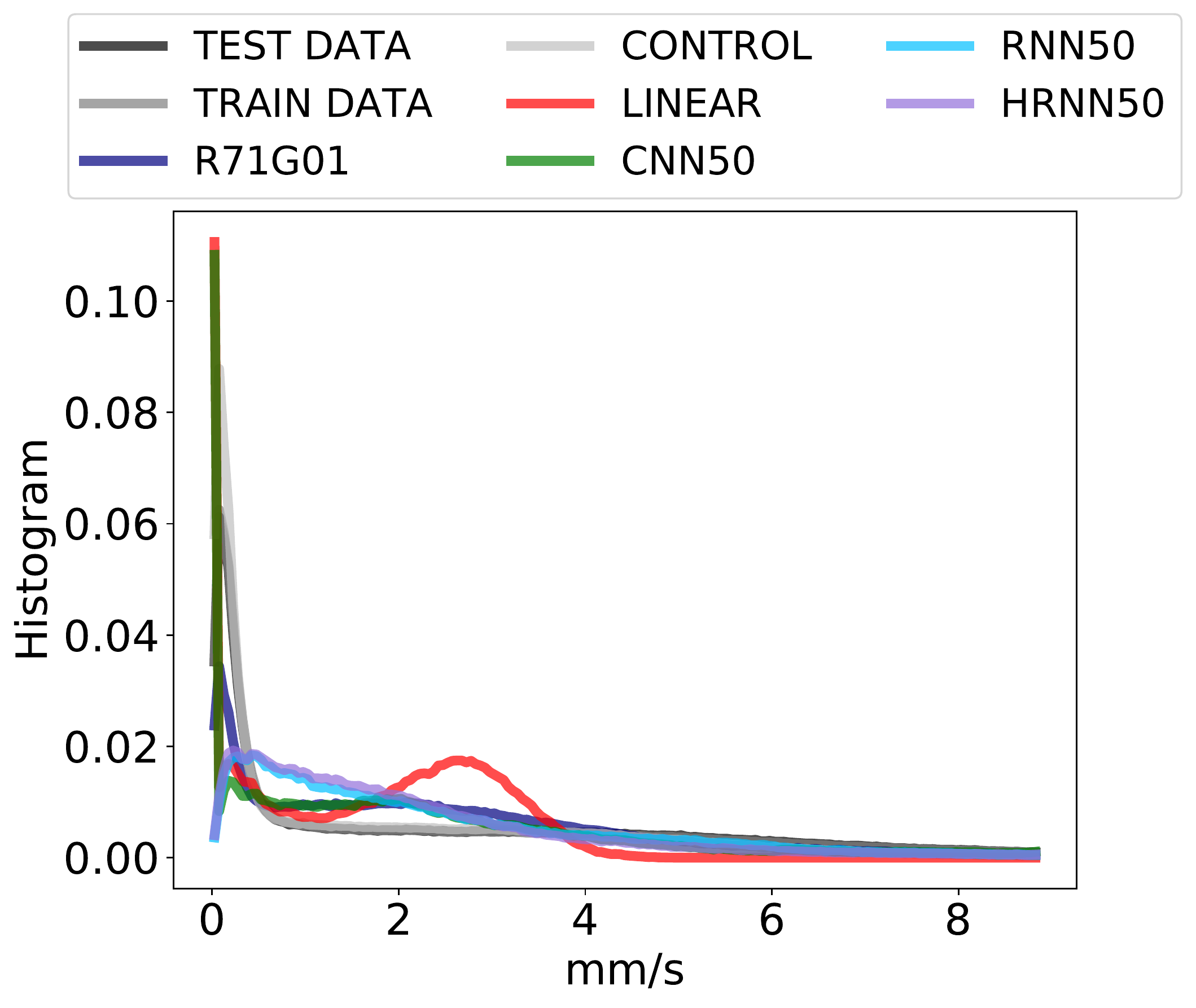}
        \vspace{-0.5cm}
        \subcaption*{Velocity}
    \end{minipage}
    \begin{minipage}{0.19\textwidth}
        \includegraphics[width=\linewidth]{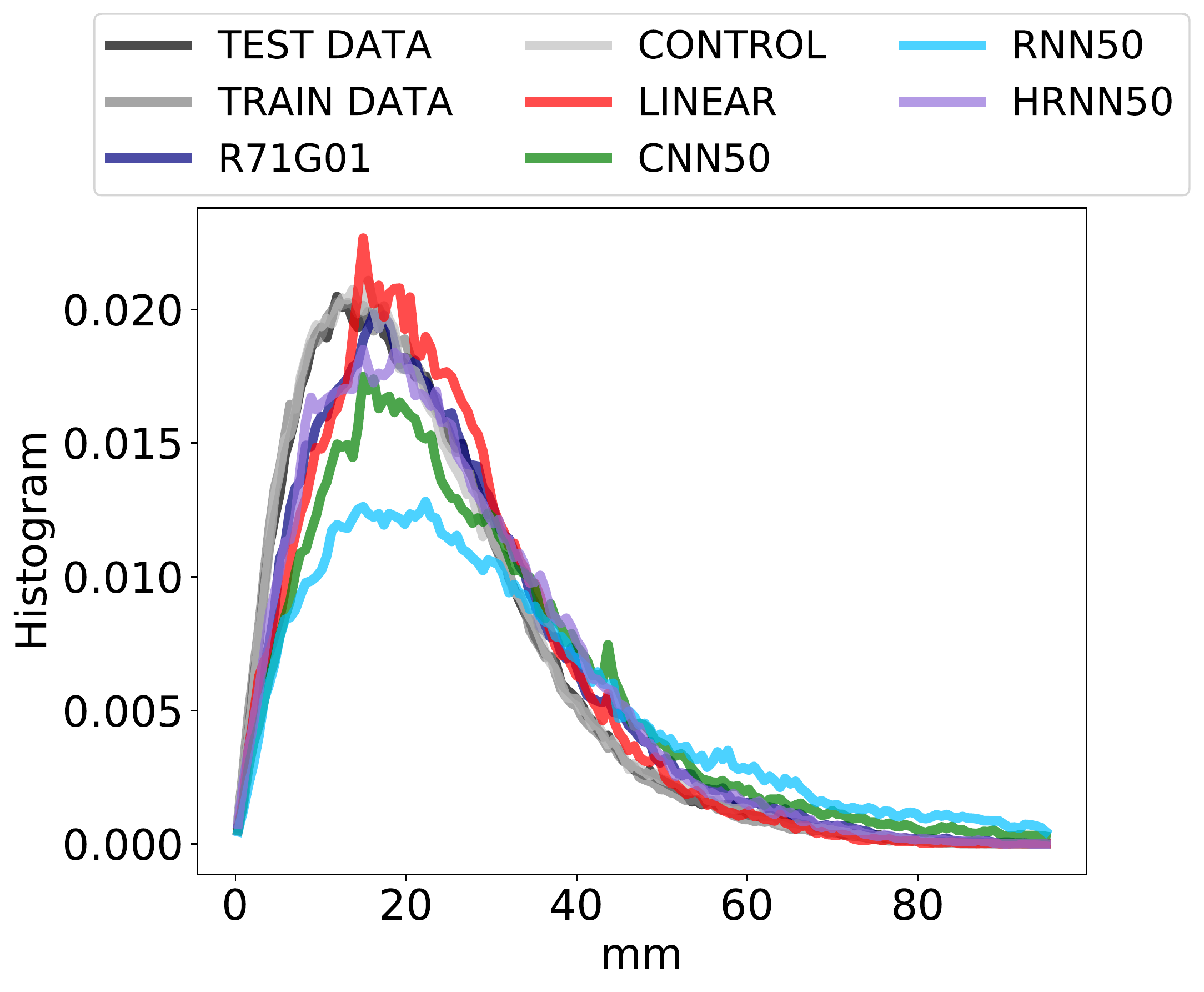}
        \vspace{-0.5cm}
        \subcaption*{Inter Distance}
    \end{minipage}
    \begin{minipage}{0.19\textwidth}
        \includegraphics[width=\linewidth]{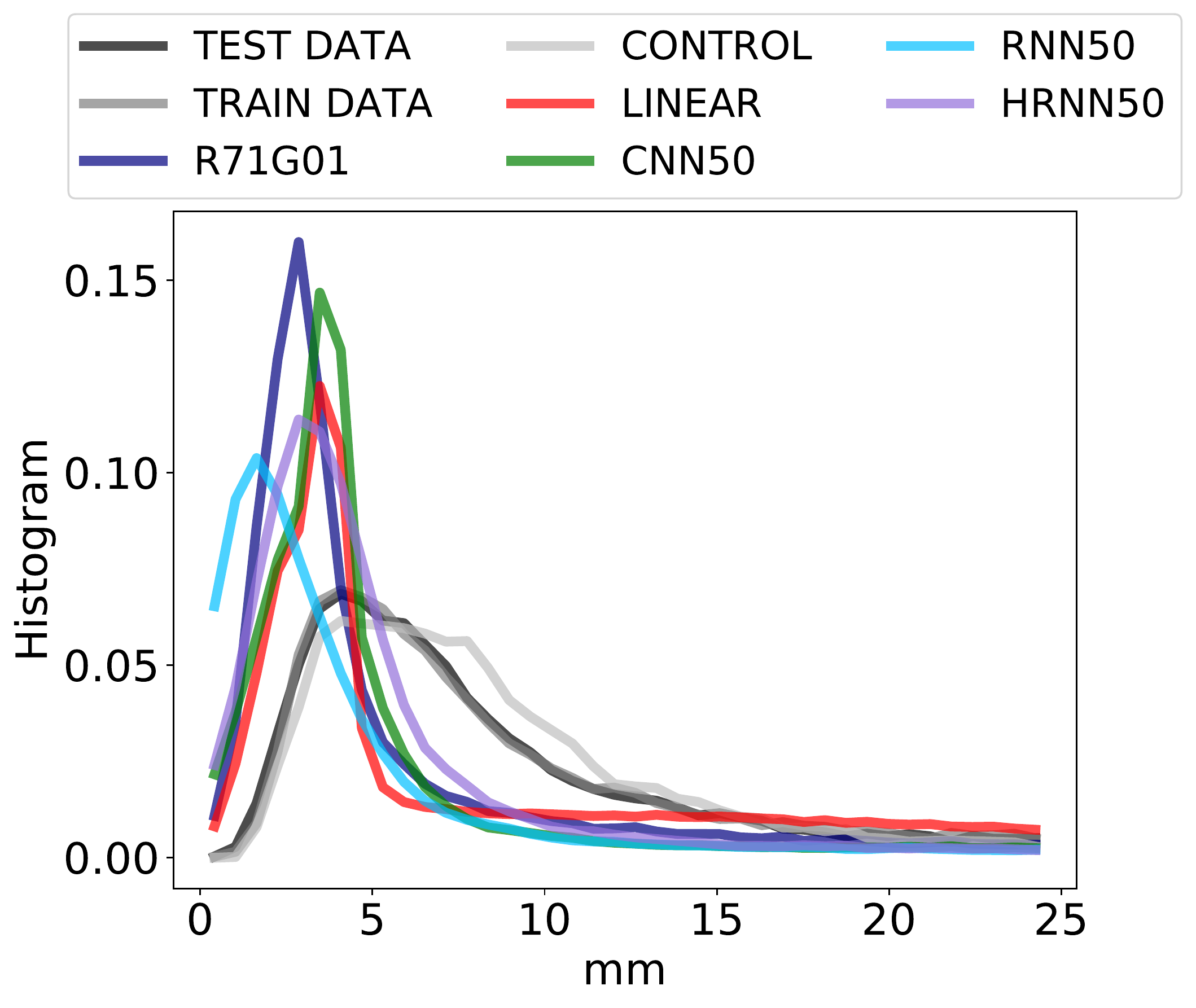}
        \vspace{-0.5cm}
        \subcaption*{Wall Distance}
    \end{minipage}
    \begin{minipage}{0.19\textwidth}
        \includegraphics[width=\linewidth]{./figs/test_hist_delta_theta_gmr91_fale_allmodel.pdf}
        \vspace{-0.5cm}
        \subcaption*{Angular Motion}
    \end{minipage}
    \begin{minipage}{0.19\textwidth}
        \includegraphics[width=\linewidth]{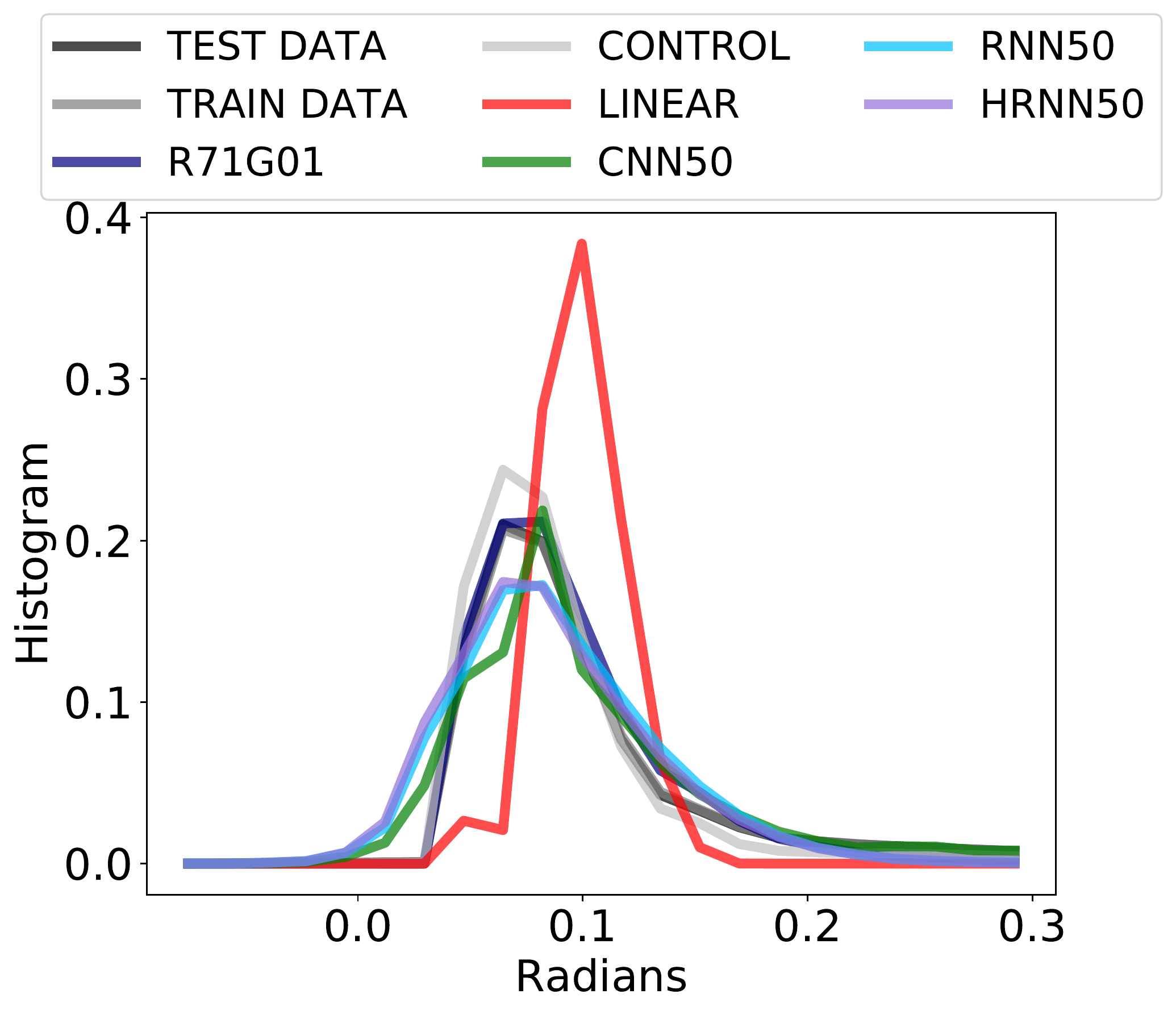}
        \vspace{-0.5cm}
        \subcaption*{Wing Angle}
    \end{minipage}
    \vspace{-0.2cm}
    \subcaption*{Female}
    \end{minipage}
    \vspace{-0.2cm}
    \caption{Histogram R91B01}
    \centering
    \begin{minipage}{0.99\textwidth}
    \begin{minipage}{0.19\textwidth}
        \includegraphics[width=\linewidth]{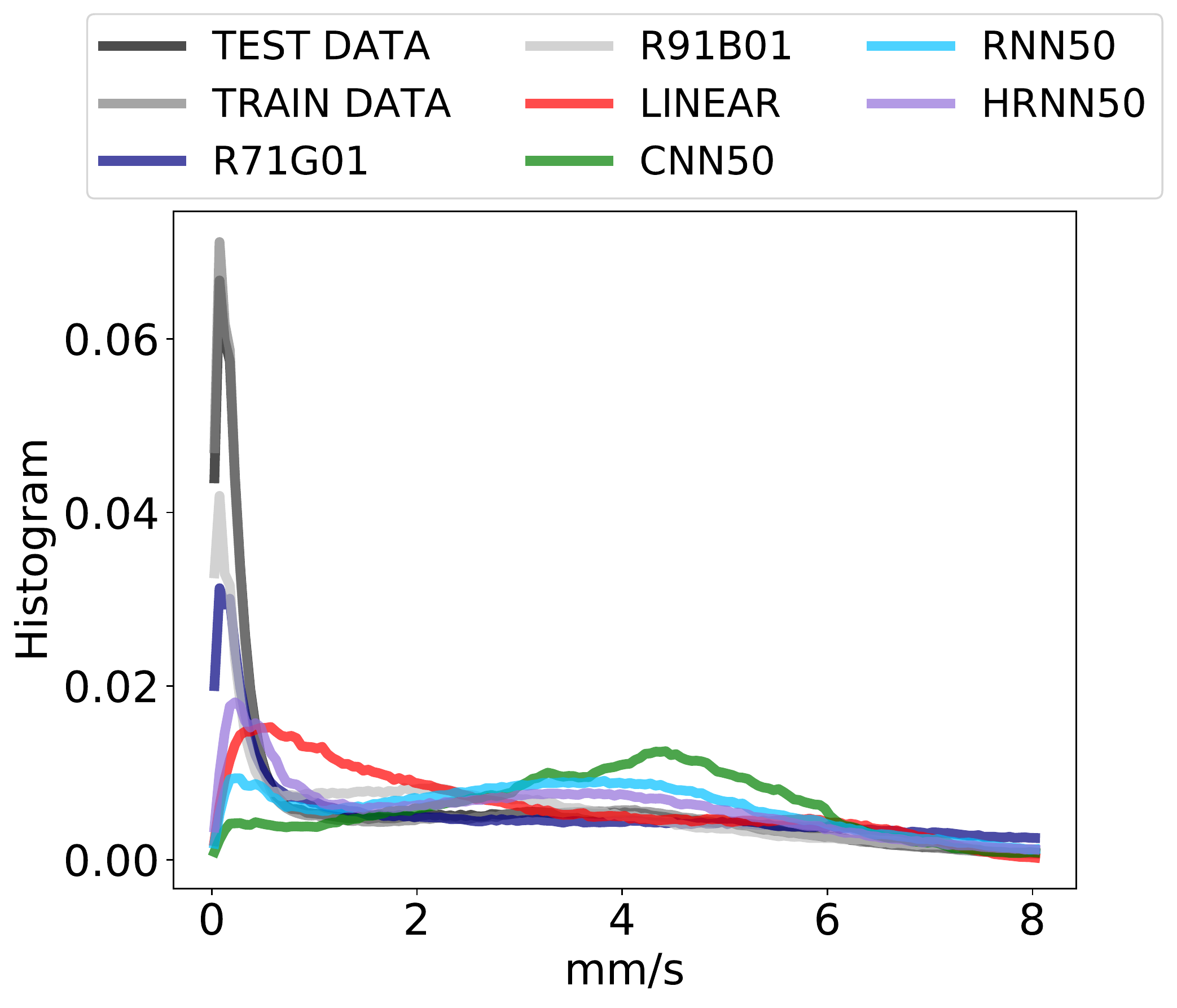}
        \vspace{-0.5cm}
        \subcaption*{Velocity}
    \end{minipage}
    \begin{minipage}{0.19\textwidth}
        \includegraphics[width=\linewidth]{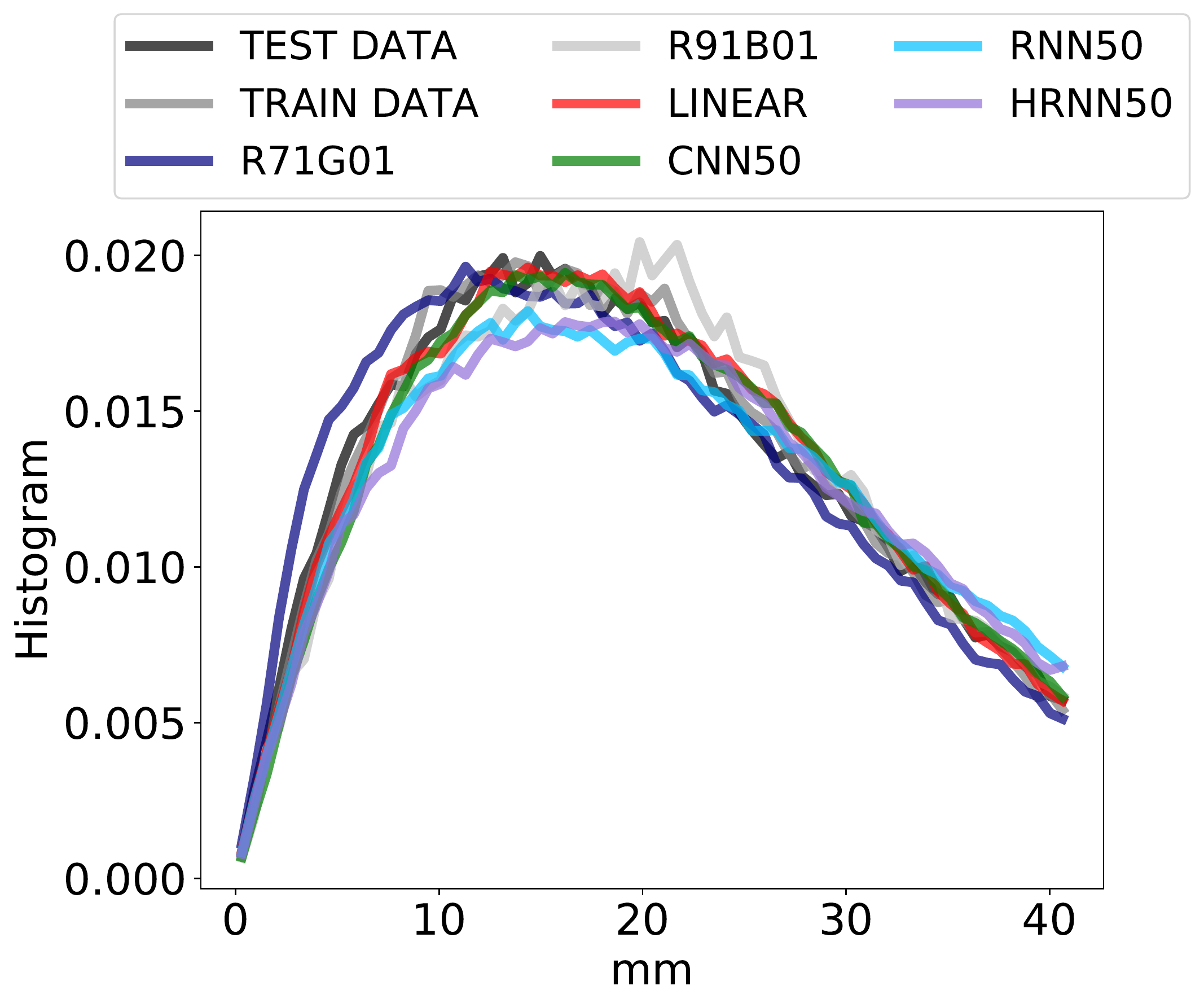}
        \vspace{-0.5cm}
        \subcaption*{Inter Distance}
    \end{minipage}
    \begin{minipage}{0.19\textwidth}
        \includegraphics[width=\linewidth]{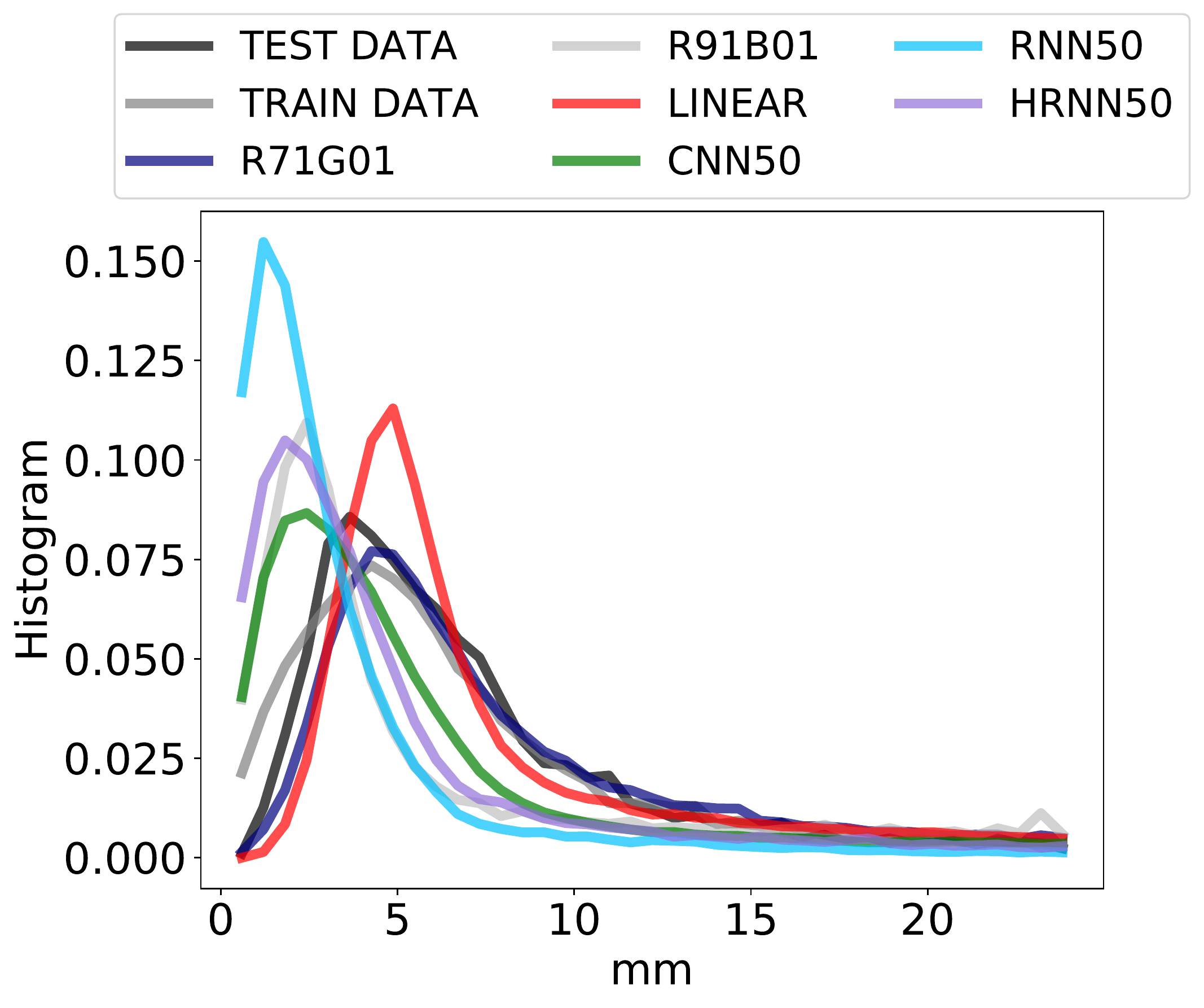}
        \vspace{-0.5cm}
        \subcaption*{Wall Distance}
    \end{minipage}
    \begin{minipage}{0.19\textwidth}
        \includegraphics[width=\linewidth]{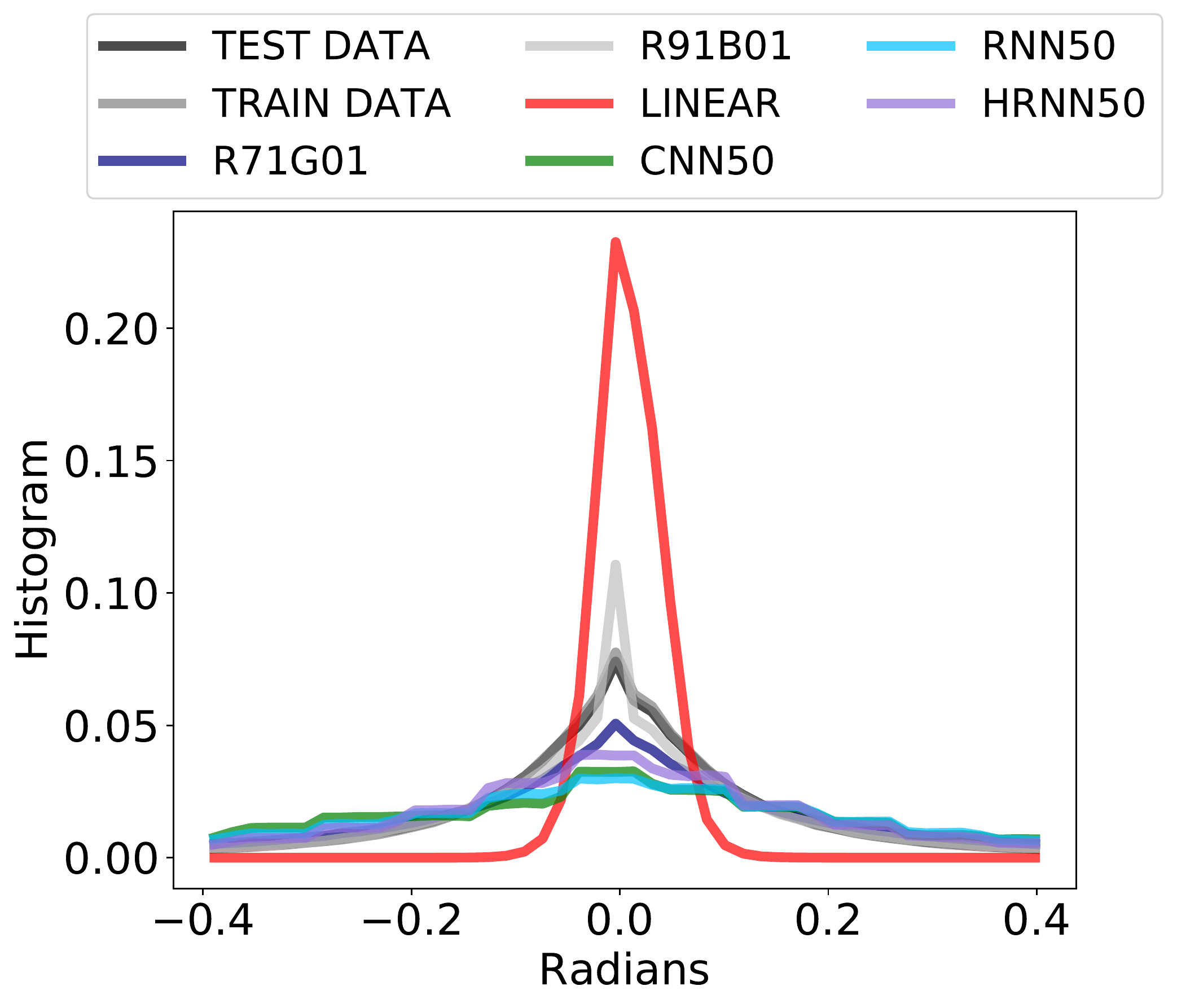}
        \vspace{-0.5cm}
        \subcaption*{Angular Motion}
    \end{minipage}
    \begin{minipage}{0.19\textwidth}
        \includegraphics[width=\linewidth]{./figs/test_hist_wing_angle_pdb_male_allmodel.pdf}
        \vspace{-0.5cm}
        \subcaption*{Wing Angle}
    \end{minipage}
    \vspace{-0.2cm}
    \subcaption*{Male}
    \end{minipage}
    \begin{minipage}{0.99\textwidth}
      \begin{minipage}{0.19\textwidth}
        \includegraphics[width=\linewidth]{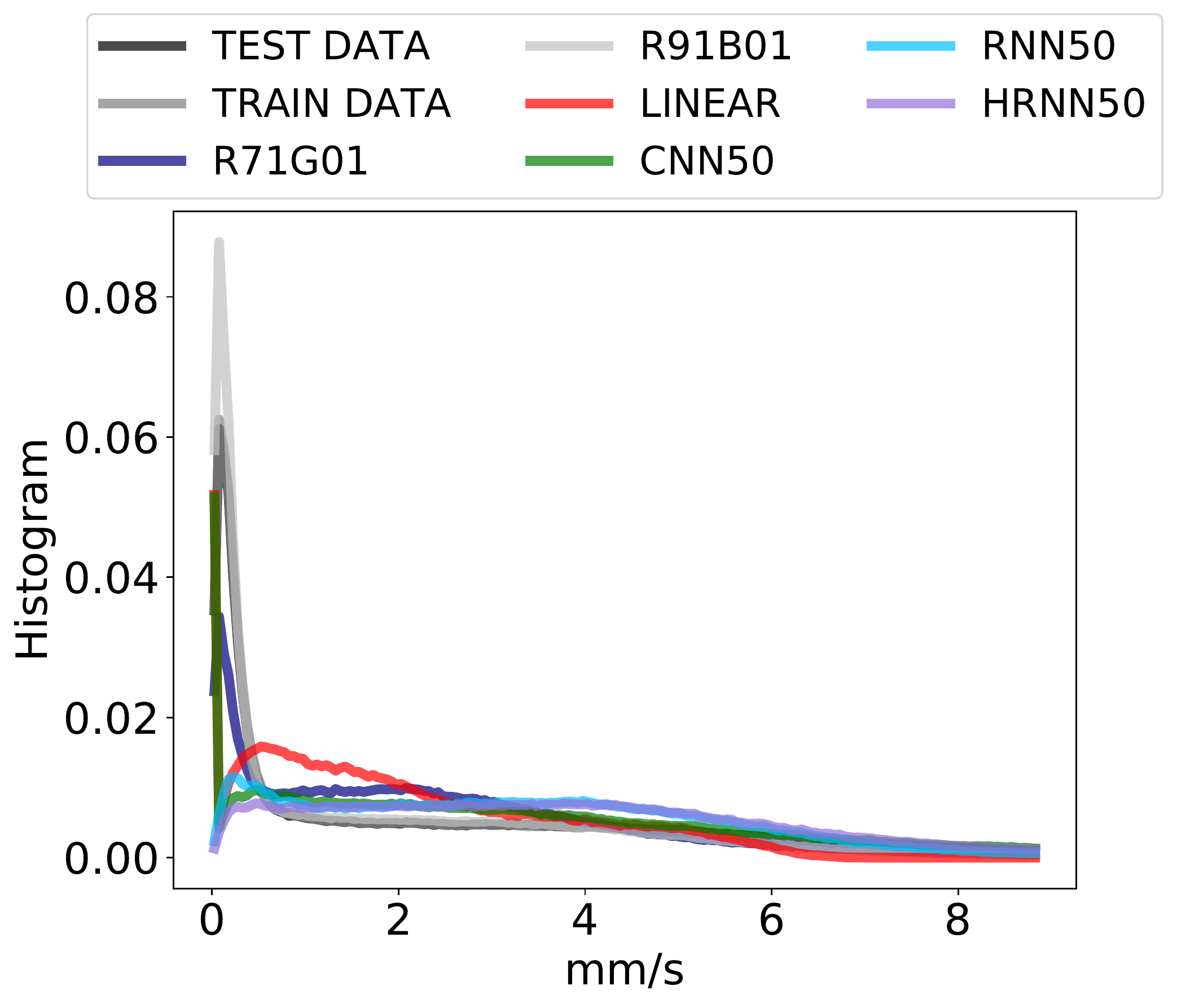}
        \vspace{-0.5cm}
        \subcaption*{Velocity}
    \end{minipage}
    \begin{minipage}{0.19\textwidth}
        \includegraphics[width=\linewidth]{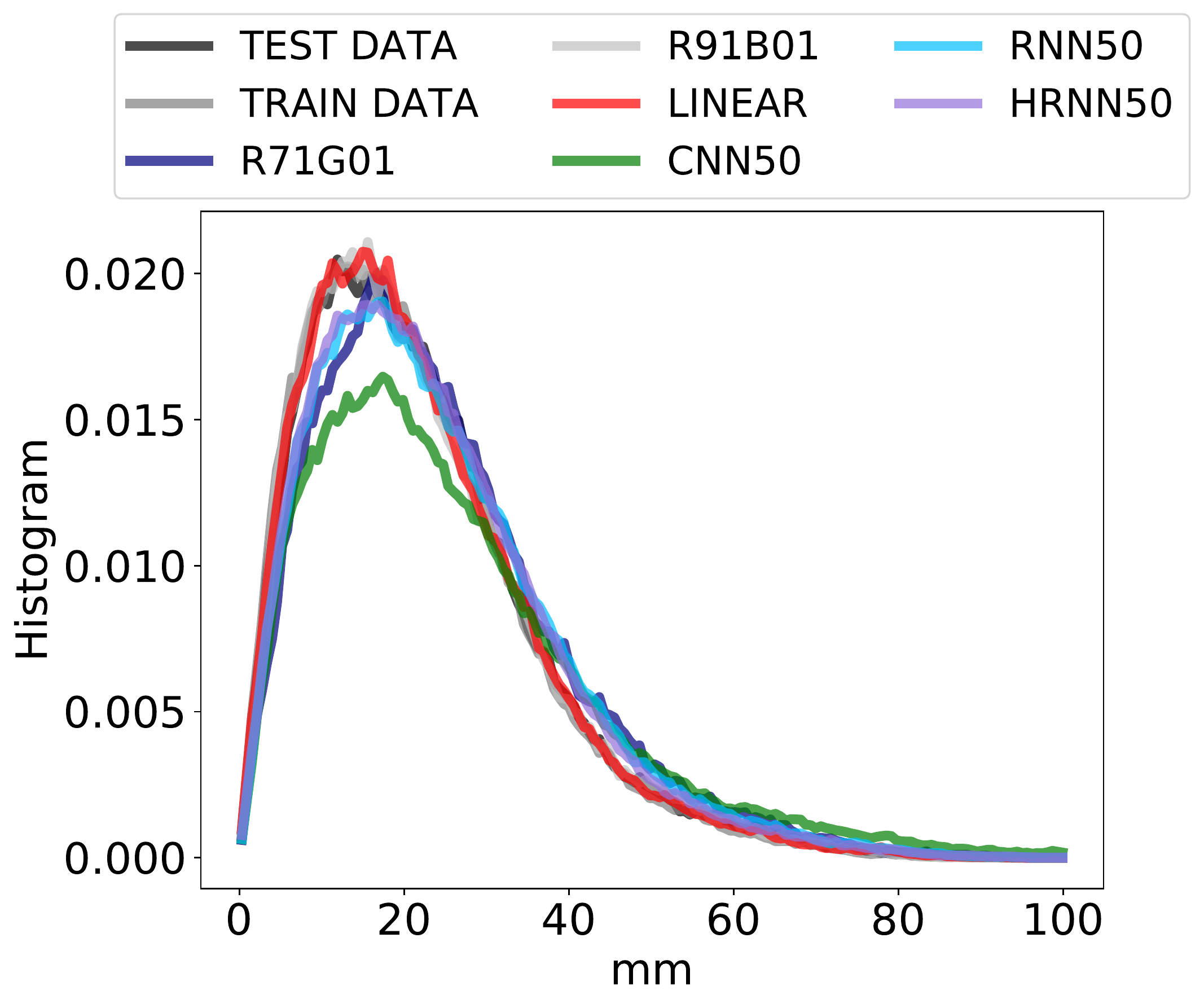}
        \vspace{-0.5cm}
        \subcaption*{Inter Distance}
    \end{minipage}
    \begin{minipage}{0.19\textwidth}
        \includegraphics[width=\linewidth]{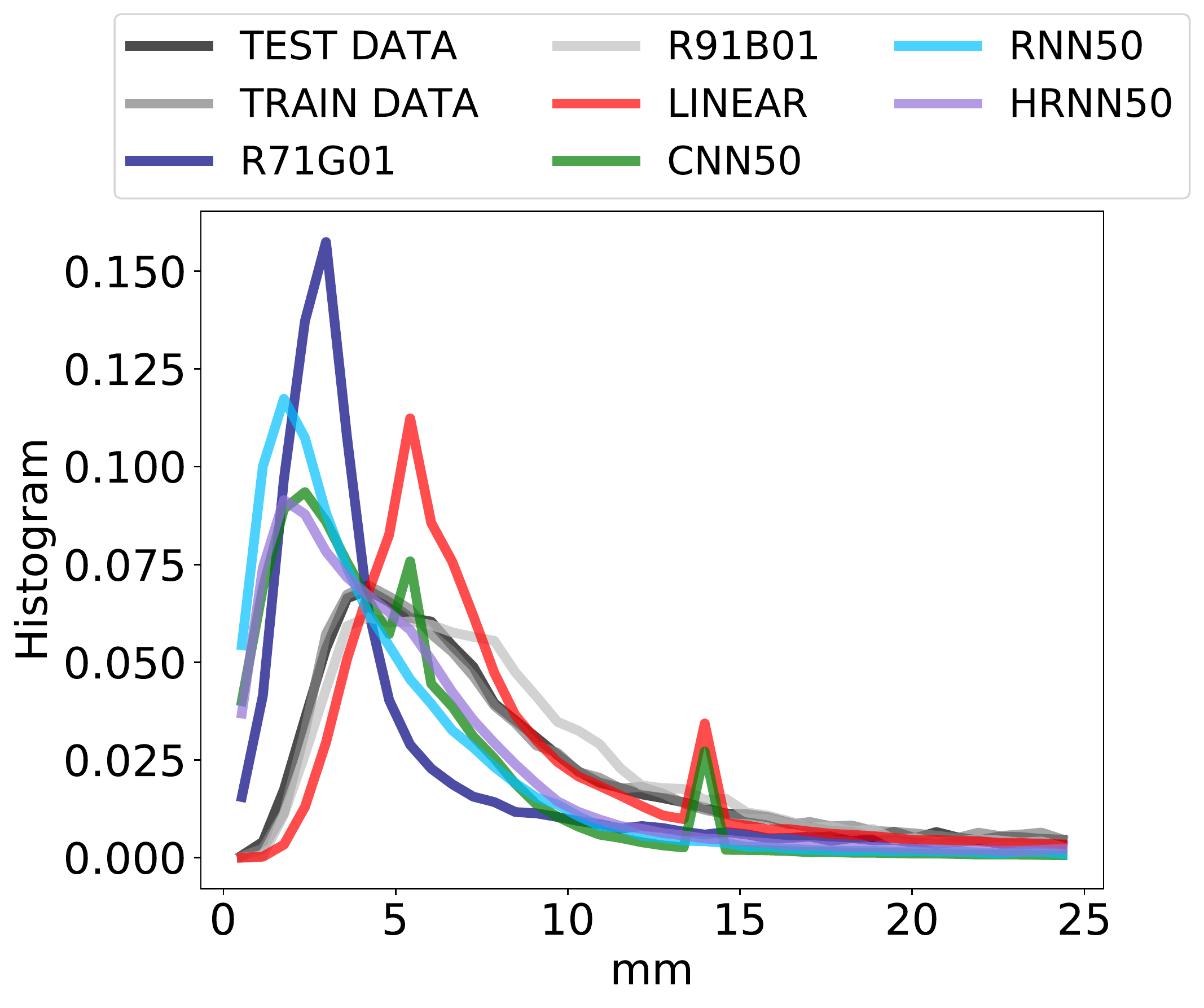}
        \vspace{-0.5cm}
        \subcaption*{Wall Distance}
    \end{minipage}
    \begin{minipage}{0.19\textwidth}
        \includegraphics[width=\linewidth]{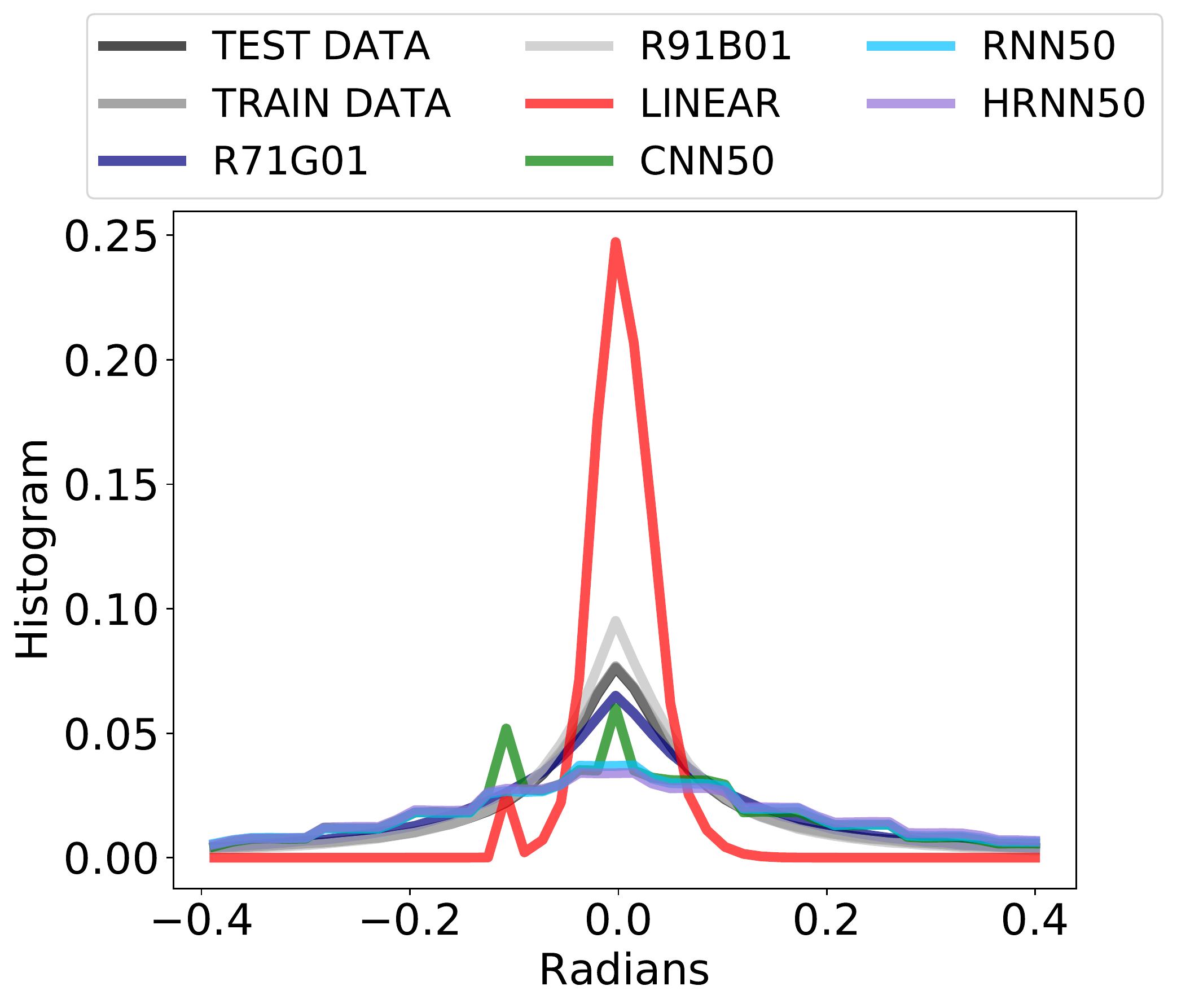}
        \vspace{-0.5cm}
        \subcaption*{Angular Motion}
    \end{minipage}
    \begin{minipage}{0.19\textwidth}
        \includegraphics[width=\linewidth]{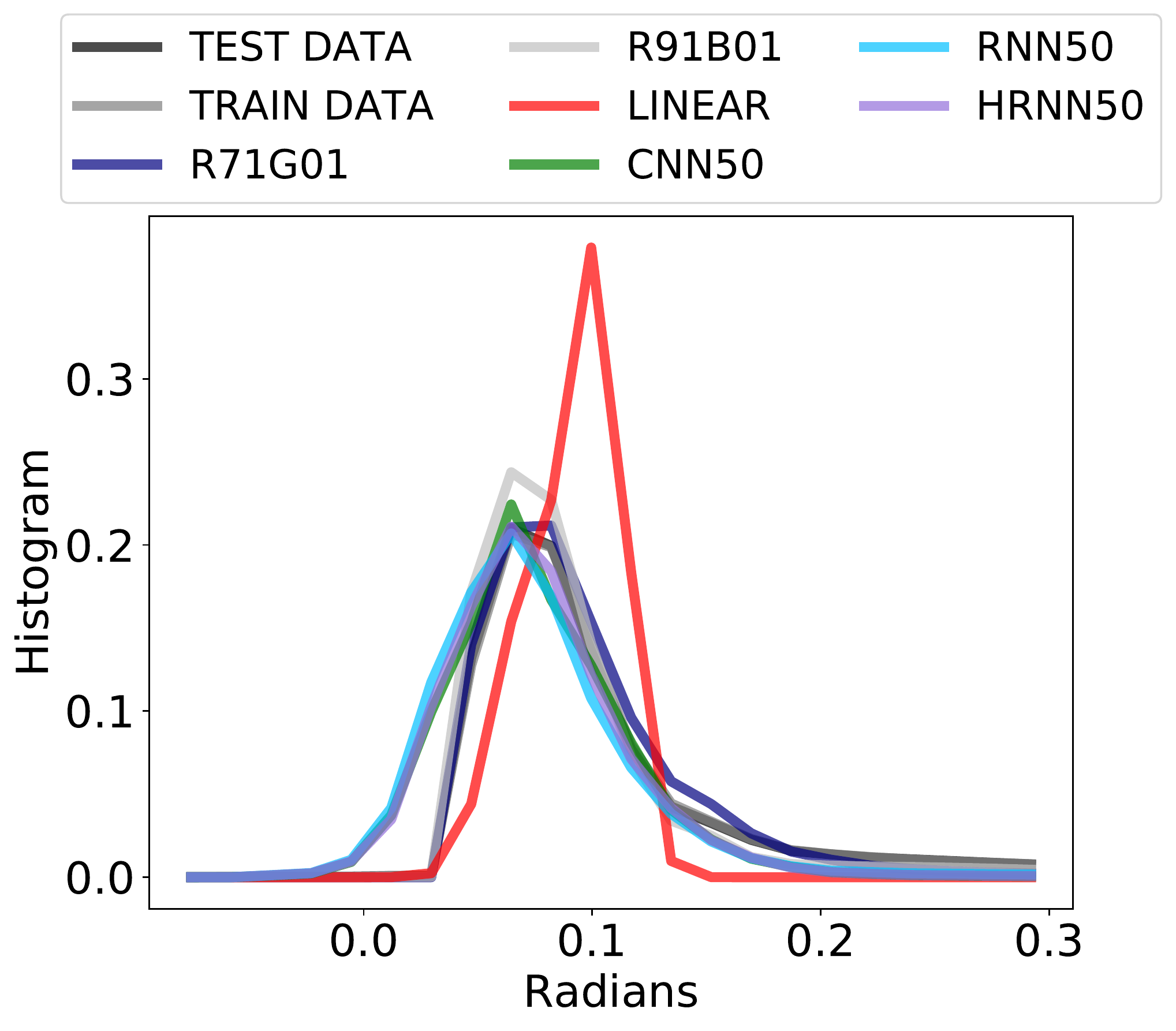}
        \vspace{-0.5cm}
        \subcaption*{Wing Angle}
    \end{minipage}
    \vspace{-0.2cm}
    \subcaption*{Female}
    \end{minipage}
    \vspace{-0.2cm}
    \caption{Histogram CONTROL}
    \label{fig:histogram_feat_supp}
\end{figure}

\begin{table*}[htp]

    \centering
    \caption{  Performance overview under various metrics for R71B01 Female}
    \label{tab:histo_dist_gmr71_female}
    {\tiny
     \begin{tabular}{|l|l|c|ccc|cc|cccc|}\hline
                                &               & TRAIN & R91B01 & CONTROL & MALE & CONST & HALT & LINEAR & CNN  & RNN  & HRNN \\\hline\hline
      NLL      &       & -     & -      & -       &-            & -     & -     & -      & 42182& 40869 & {\bf 40628} \\ \hline
      1-step   &       & -     & -      & -       &  -          & 3.75  & 5.42 & 2.67   & {\bf 2.57} & 4.27 & 4.47\\
      30-step  &       & -     & -      & -       &  -          & 69.1  & 67.2 & 29.2   & {\bf 20.7} & 46.2 & 42.0\\\hline
      \multirow{2}{*}{RvF}      & Eng.          & 59.8  & 84.0   & {\bf 58.7}    & 93.7        & -     & -    & 98.7   & 98.8  & 76.8 & 74.4\\
                                & Raw traj.     & 50.8  & 87.2   & 66.7    & 84.1        & -     & -    & 78.1   & {\bf 61.9} & 71.8 & 67.5\\\hline
      \multirow{4}{*}{HD}       & Vel.          & 0.039 & 0.472  & {\bf 0.216}   & -           & -     & -    & 0.886  & 0.354& 0.464& 0.360\\
                                & Inter. Dist.  & 0.034 & 0.126  & {\bf 0.037}   & -           & -     & -    & 0.080  & 0.078& 0.071& 0.0.095\\
                                & Wall Dist.    & 0.015 & 0.237  & 0.271   & -           & -     & -    & 0.917  & 0.652& 0.455& {\bf 0.185}\\
                                & Ang. Mot.     & 0.023 & {\bf 0.135}  & 0.141   & -           & -     & -    & 0.948  & 0.268& 0.368& 0.244\\\hline
    \end{tabular}
    }\\
    \caption{  Performance overview under various metrics for R91B01}
    \label{tab:histo_dist_gmr91}
  \centering
    \subcaption{ Male}
  \label{tab:histo_dist_gmr91_male}
    {\tiny
  \begin{tabular}{|l|l|c|ccc|cc|cccc|}\hline
                &               & TRAIN & R71G01 & CONTROL & FEMALE & CONST & HALT  & LINEAR & CNN  & RNN  & HRNN \\\hline\hline
      NLL      &       & -     & -      & -       &-            & -     & -     & -      & 57837 & 52589 & 52738 \\ \hline
      1-step   &       & -     & -      & -       &-            & 5.49  & 8.07  & 5.46   & 2.56 & 2.26 & 2.30 \\
      30-step   &       & -     & -      & -       &-           & 80.4 & 71.9  & 56.1   & 28.9 & 25.6 & 25.6\\\hline
      \multirow{2}{*}{RvF}      & Eng.          & 75.9  & 84.5  & 77.7    & 90.9        & -     & -     & 94.2   & 98.9 & 73.5 & 76.8\\
                                & Raw traj.     & 54.1  & 90.4  & 63.4    & 83.5        & -     & -     & 75.0   & 55.8  & 57.3  & 56.5   \\\hline
      Chase                     &  &  -  & -      & -       & -           & -     & -    & 0.014 & 0.185 & 0.020 & 0.022 \\\hline
      \multirow{4}{*}{HD}       & Vel.          & 0.108 & 0.353  & 0.352  & -           & -     & -     & 0.839  & 0.378   & 0.289 & 0.305\\
                                & Inter. Dist.  & 0.159 & 0.136  & 0.084  & -           & -     & -     & 0.159  & 0.109 & 0.097 & 0.368\\
                                & Wall Dist.    & 0.131 & 0.755  & 0.761  & -           & -     & -     & 0.534  & 0.322& 0.474 & 0.496\\
                                & Ang. Mot.     & 0.237 & 0.135  & 0.116  & -           & -     & -     & 1.260  & 0.273 & 0.203 & 0.284\\\hline
  \end{tabular}
    }
  \centering
    \subcaption{Female}
  \label{tab:histo_dist_gmr91_female}
    {\tiny
  \begin{tabular}{|l|l|c|ccc|cc|cccc|}\hline
                                &               & TRAIN & R71G01 & CONTROL & MALE & CONST & HALT & LINEAR & CNN  & RNN  & HRNN \\\hline\hline
      NLL      &       & -     & -      & -       &-            & -     & -     & -      & 45609& 43306 & 43597 \\ \hline
      1-step   &       & -     & -      & -       &  -          & 4.65  & 7.54 & 4.87   & 2.03 & 3.63 & 5.11\\
      30-step  &       & -     & -      & -       &  -          & 69.1  & 61.7 & 46.8   & 23.5 & 48.5 & 55.0\\\hline
      \multirow{2}{*}{RvF}      & Eng.          & 71.7  & 72.8   & 72.7    & 91.7        & -     & -    & 98.2   & 98.8  & 80.1 & 79.4\\
                                & Raw traj.     & 56.1  & 87.0   & 72.7    & 84.2        & -     & -    & 78.0   & 56.6  & 58.8 & 57.4\\\hline
      \multirow{4}{*}{HD}       & Vel.          & 0.165 & 0.472  & 0.216   & -           & -     & -    & 0.580  & 0.339 & 0.243 & 0.348\\
                                & Inter. Dist.  & 0.165 & 0.126  & 0.037   & -           & -     & -    & 0.094  & 0.220 & 0.251 & 0.058\\
                                & Wall Dist.    & 0.366 & 0.761  & 0.205   & -           & -     & -    & 0.452  & 0.468 & 0.547 & 0.331\\
                                & Ang. Mot.     & 0.126 & 0.135  & 0.141   & -           & -     & -    & 1.079  & 0.327 & 0.182 & 0.159\\\hline
  \end{tabular}
    }
    \caption{  Performance overview under various metrics for CONTROL}
    \label{tab:histo_dist_pdb}
  \centering
    \subcaption{ Male}
  \label{tab:histo_dist_pdb_male}
    {\tiny
  \begin{tabular}{|l|l|c|ccc|cc|cccc|}\hline
                &               & TRAIN & R71G01 & R91B01 & FEMALE & CONST & HALT  & LINEAR & CNN  & RNN  & HRNN \\\hline\hline
      NLL      &       & -     & -      & -       &-            & -     & -     & -      & 52236 & 49646 & 49766 \\ \hline
      1-step   &       & -     & -      & -       &-            & 3.99  & 6.53  & 3.23   & 1.43 & 1.24 & 1.33 \\
      30-step   &       & -     & -      & -       &-           & 51.7  & 55.9  & 36.8   & 14.3 & 12.7 & 13.2 \\\hline
      \multirow{2}{*}{RvF}      & Eng.          & 63.7  & 87.0  & 86.4    & 99.3        & -     & -     & 98.9   & 99.6 & 96.7 & 94.0\\
                                & Raw traj.     & 55.1  & 79.6   & 82.2    & 96.3        & -     & -     & 81.0   & 61.3  & 57.1  & 58.9   \\\hline
      Chase                     &  &  -  & -      & -       & -           & -     & -    & 0.014 & 0.185 & 0.020 & 0.022 \\\hline
      \multirow{4}{*}{HD}       & Vel.          & 0.045 & 0.421  & 0.352   & -           & -     & -    & 0.594  & 0.761& 0.510 & 0.477\\
                                & Inter. Dist.  & 0.047 & 0.070  & 0.084   & -           & -     & -     & 0.083  & 0.078 & 0.071 & 0.050\\
                                & Wall Dist.    & 0.044 & 0.211  & 0.761   & -           & -     & -     & 0.440  & 0.477 & 0.455 & 0.320\\
                                & Ang. Mot.     & 0.015 & 0.271  & 0.116   & -           & -     & -     & 0.917  & 0.652 & 0.455 & 0.310\\\hline
  \end{tabular}
    }
  \centering
    \subcaption{Female}
  \label{tab:histo_dist_pdb_female}
    {\tiny
  \begin{tabular}{|l|l|c|ccc|cc|cccc|}\hline
                                &               & TRAIN & R71G01 & R91B01 & MALE & CONST & HALT & LINEAR & CNN  & RNN  & HRNN \\\hline\hline
      NLL      &       & -     & -      & -       &-            & -     & -     & -      & 45017& 43141 & 43173 \\ \hline
      1-step   &       & -     & -      & -       &  -          & 3.57  & 5.50 & 3.03   & 1.37 & 2.67 & 2.25\\
      30-step  &       & -     & -      & -       &  -          & 47.2  & 46.5 & 32.1   & 13.5 & 35.2 & 31.7\\\hline
      \multirow{2}{*}{RvF}      & Eng.          & 63.0  & 69.3  & 87.8  & 99.3        & -     & -    & 99.8   & 100  & 79.7 & 77.1\\
                                & Raw traj.     & 56.3  & 66.9   & 83.7    & 95.4        & -     & -    & 78.1   & 58.5 & 56.3 & 58.1\\\hline
      \multirow{4}{*}{HD}       & Vel.          & 0.039 & 0.472  & 0.216   & -           & -     & -    & 0.886  & 0.354& 0.464& 0.360\\
                                & Inter. Dist.  & 0.044 & 0.126  & 0.037   & -           & -     & -    & 0.440  & 0.131& 0.071& 0.320\\
                                & Wall Dist.    & 0.015 & 0.763  & 0.205   & -           & -     & -    & 0.917  & 0.652& 0.455& 0.185\\
                                & Ang. Mot.     & 0.023 & 0.135  & 0.141   & -           & -     & -    & 0.948  & 0.268& 0.368& 0.244\\\hline
  \end{tabular}
    }
\end{table*}

\begin{figure}[htb]
    \centering
    \begin{minipage}{0.29\textwidth}
        \includegraphics[width=\linewidth]{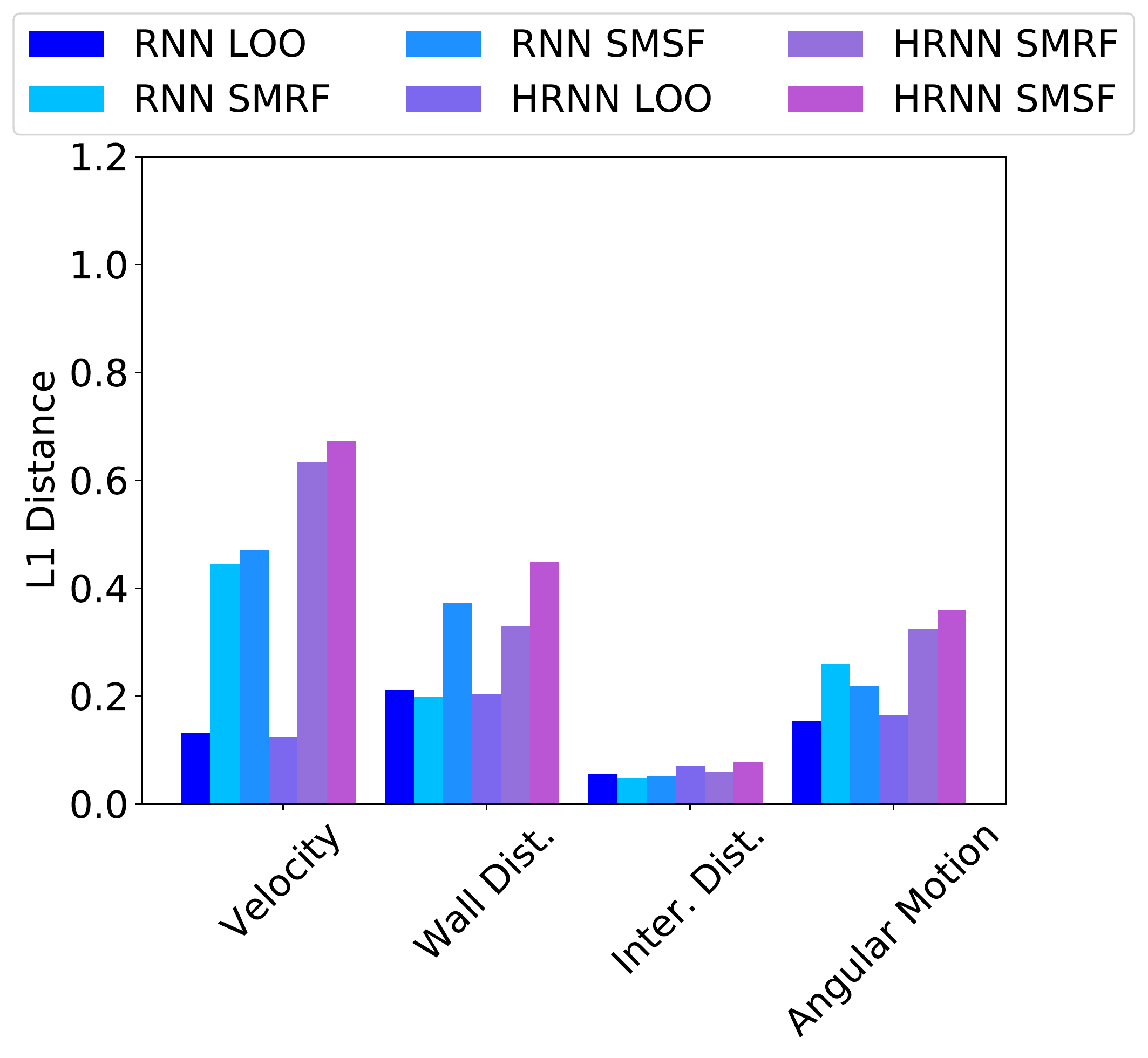}
        \vspace{-0.5cm}
        \subcaption{R71G01, Male}
    \end{minipage}
    \begin{minipage}{0.29\textwidth}
        \includegraphics[width=\linewidth]{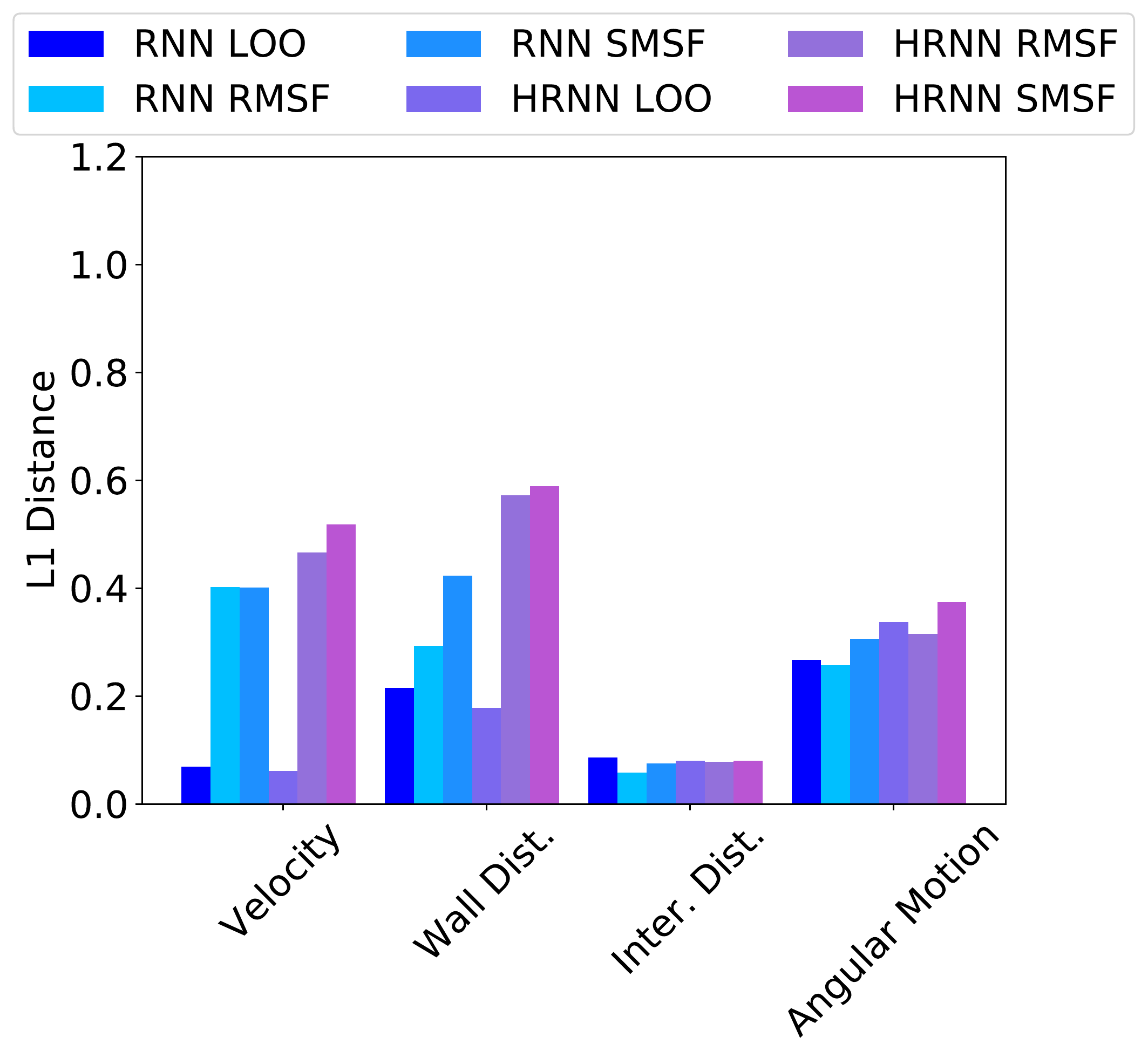}
        \vspace{-0.5cm}
        \subcaption{R71G01,Female}
    \end{minipage}
    \centering
    \begin{minipage}{0.24\textwidth}
        \includegraphics[width=\linewidth]{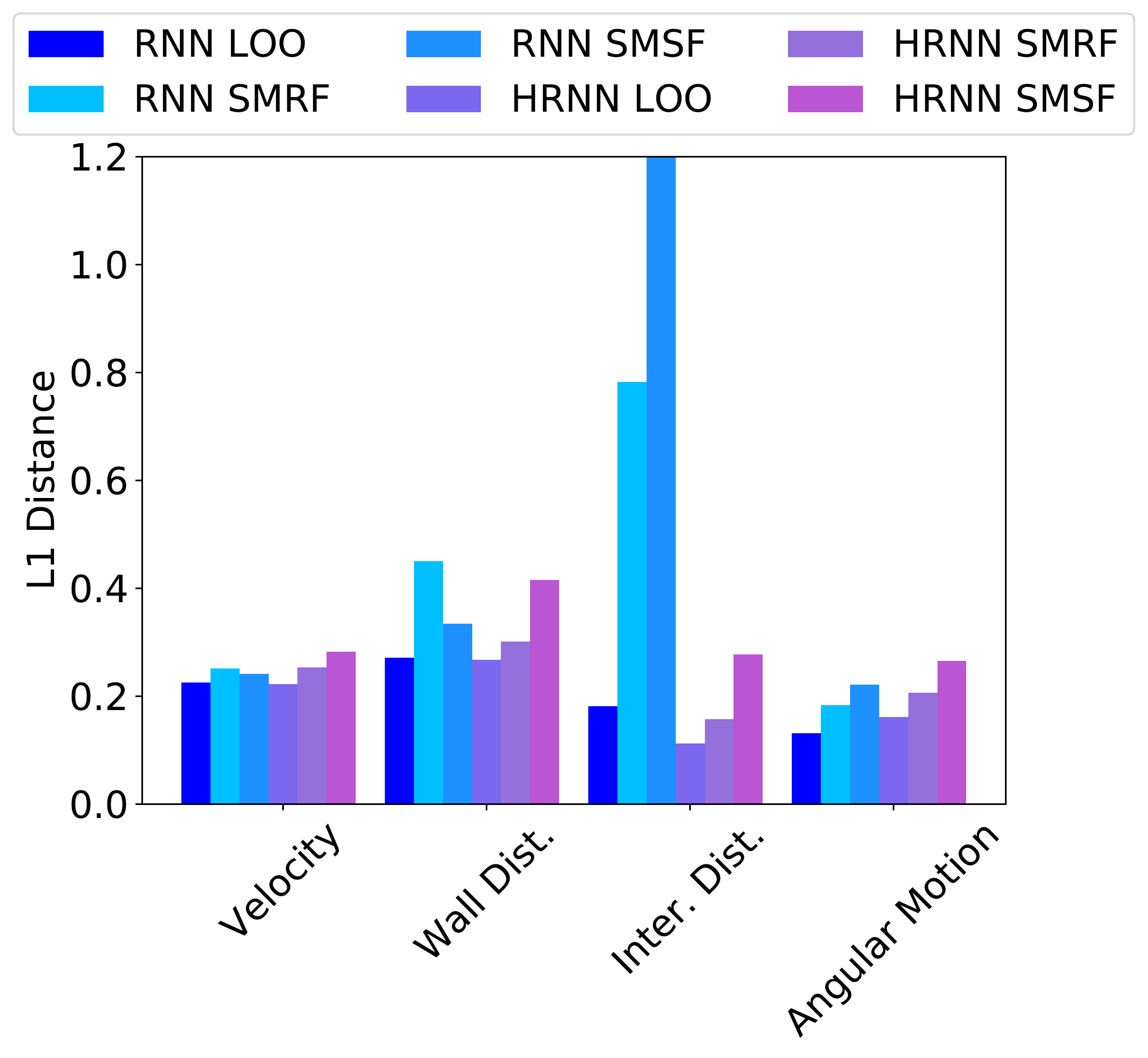}
        \vspace{-0.5cm}
        \subcaption{R91B01, Male}
    \end{minipage}
    \begin{minipage}{0.24\textwidth}
        \includegraphics[width=\linewidth]{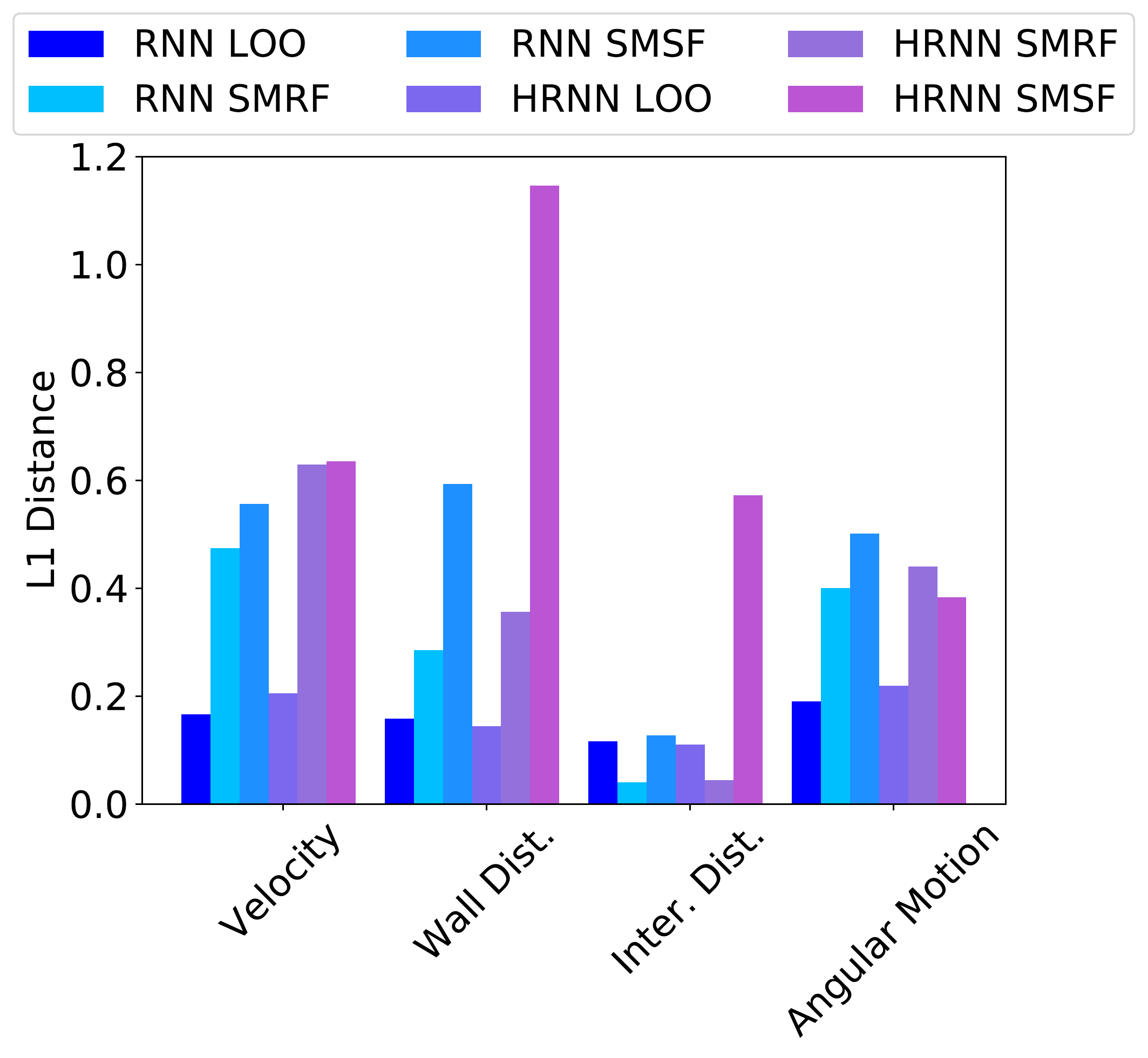}
        \vspace{-0.5cm}
        \subcaption{CONTROL, Male}
    \end{minipage}
    \begin{minipage}{0.24\textwidth}
        \includegraphics[width=\linewidth]{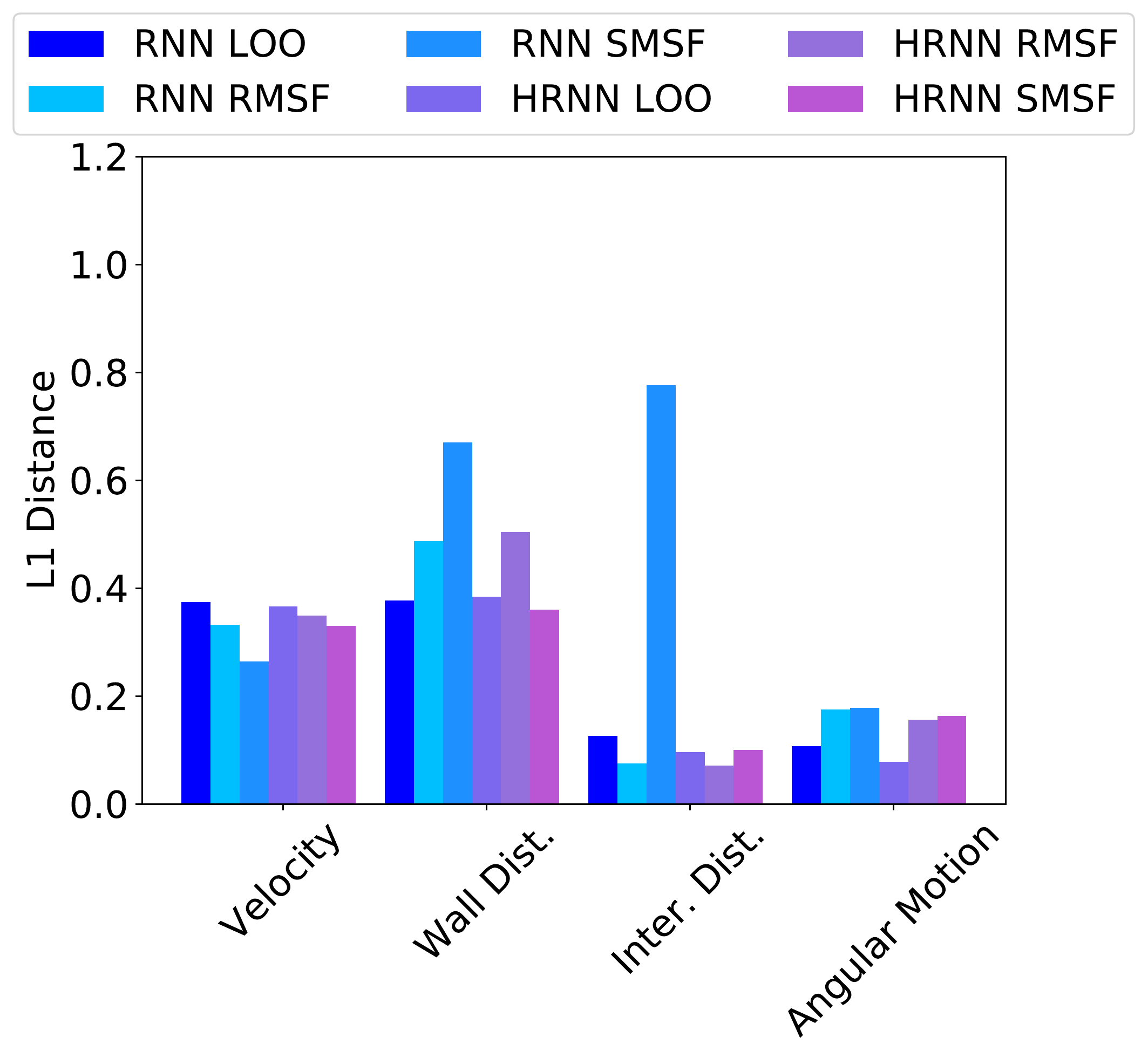}
        \vspace{-0.5cm}
        \subcaption{R91B01, Female}
    \end{minipage}
    \begin{minipage}{0.24\textwidth}
        \includegraphics[width=\linewidth]{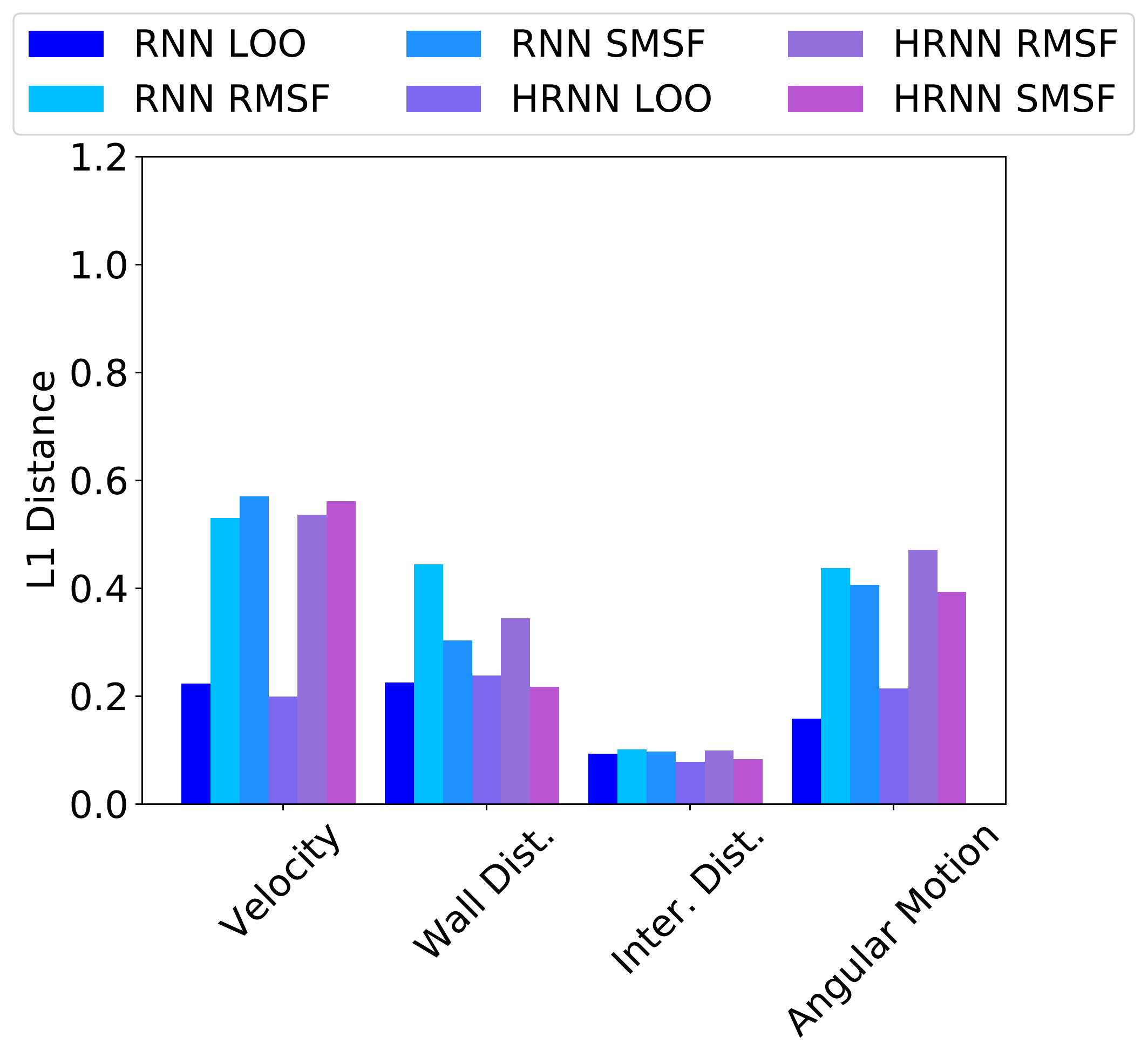}
        \vspace{-0.5cm}
        \subcaption{CONTROL, Female}
    \end{minipage}
    \caption{The distribution distance comparison between simulation
    $\mathcal{S}_{LOO}$, $\mathcal{S}_{\text{RMSF}}$, $\mathcal{S}_{\text{SMRF}}$, and $\mathcal{S}_{\text{SMSF}}$
    for R71G01, R91B01 and CONTROL.}
    \label{fig:simulation}
\end{figure}

\end{appendices}

\end{document}